%% file: arxiv_submitted.tex
\documentclass[10pt,twocolumn,letterpaper]{article}

\usepackage[pagenumbers]{cvpr} 

\input{preamble}
\definecolor{cvprblue}{rgb}{0.21,0.49,0.74}
\usepackage[pagebackref,breaklinks,colorlinks,allcolors=cvprblue]{hyperref}

\usepackage{pifont}    
\usepackage{soul}    
\usepackage{xcolor}  

\newcommand{\edit}[2]{\textcolor{red}{\st{#1}} \textcolor{blue}{#2}}
\usepackage{graphicx}
\definecolor{myGreen}{HTML}{1A9988}
\usepackage{makecell} 

\definecolor{checkmark}{HTML}{068c49}

\def\paperID{7045} 
\def\confName{CVPR}
\def\confYear{2026}

\title{CrossHOI-Bench: A Unified Benchmark for HOI Evaluation across Vision-Language Models and HOI-Specific Methods}

\author{%
  {Qinqian Lei$^1$}\quad
  { Bo Wang$^2$}  \quad
      {Robby T. Tan$^{1,3}$}  \\
  $^1$National University of Singapore \;
   $^2$University of Mississippi  \\
   $^3$ASUS Intelligent Cloud Services\thanks{Robby T. Tan was affiliated with ASUS Intelligent Cloud Services during part of the period in which this work was developed.}\\
  \tt\small {qinqian.lei@u.nus.edu}  \; \tt\small {hawk.rsrch@gmail.com}  \; \tt\small {robby.tan@nus.edu.sg} \\
  \tt\small \url{https://github.com/ChelsieLei/CrossHOI-Bench}
}

\begin{document}
\maketitle

\begin{abstract}
HOI detection has long been dominated by task-specific models, sometimes with early vision-language backbones such as CLIP. With the rise of large generative VLMs, a key question is whether standalone VLMs can perform HOI detection competitively against specialized HOI methods.
Existing benchmarks such as HICO-DET require exact label matching under incomplete annotations, so any unmatched prediction is marked wrong. This unfairly penalizes valid outputs, especially from less constrained VLMs, and makes cross-paradigm comparison unreliable.
To address this limitation, we introduce CrossHOI-Bench, a multiple-choice HOI benchmark with explicit positives and curated negatives, enabling unified and reliable evaluation of both VLMs and HOI-specific models.
We further focus on challenging scenarios, such as multi-person scenes and fine-grained interaction distinctions, which are crucial for revealing real differences between the two paradigms.
Experiments show that large VLMs achieve competitive, sometimes superior, zero-shot performance, yet they struggle with multiple concurrent actions and with correctly assigning interactions to the target person. 
Conversely, HOI-specific methods remain weaker in general HOI reasoning but demonstrate stronger multi-action recognition and more reliable identification of which person performs which action.
These findings expose complementary strengths and weaknesses of VLMs and HOI-specific methods, which existing benchmarks fail to reveal due to incorrect penalization.
\end{abstract}

\section{Introduction}

Human-object interaction (HOI) detection has long been approached with task-specific models~\cite{kim_hotr_2021,chen2021qahoi,Yang_2025_ICCV,wang2024bilateral,li2024human}. 
Existing HOI-specific models have adopted early vision-language models (VLMs) such as CLIP~\cite{radford2021learning} as fixed feature encoders~\cite{yang2024open,mao2024clip4hoi}.
Recently, the rise of large, generative VLMs such as Qwen2.5-VL~\cite{Qwen_2_5_report} and InternVL~\cite{InternVL3_2025} have shown that these models can directly describe complex visual scenes, often including human–object interactions~\cite{GPT_4o_eval_2024,chen2024internvl,Yang_2023_GPT4V_eval,feng2025can,feng2025rewardmap,feng2025efficient}. 
This raises an important question: can such standalone VLMs effectively perform HOI detection, and how do they compare to specialized HOI methods? 
Addressing this question requires a new benchmark that can evaluate both paradigms under a unified protocol.
%
%
%

\begin{figure}[t]
	\centering
	\includegraphics[width=0.98\linewidth]{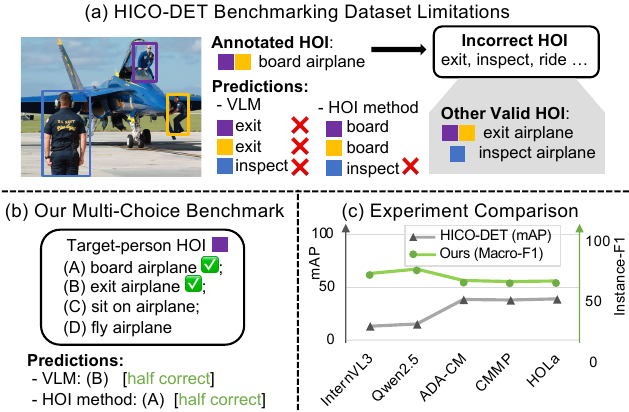}
    \caption{
    (a) Existing HOI benchmarks (e.g., HICO-DET) rely on exact-match evaluation under incomplete annotations, penalizing valid yet unlabeled interactions
    (b) Our multi-choice benchmark accepts multiple correct answers and avoids false negatives and enabling unified evaluation of HOI-specific methods and VLMs.
    (c) Comparison of state-of-the-art (InternVL3~\cite{InternVL3_2025}, Qwen2.5-VL-32B~\cite{Qwen_2_5_report}) and HOI-specific methods (ADA-CM~\cite{lei2023efficient}, CMMP~\cite{lei2024exploring}, HOLa~\cite{lei2025lhola}). Results are shown using Macro-F1 in our benchmark (\textit{Setting 1}) versus mean Average Precision (mAP) in HICO-DET.
        }	
	\label{fig: teaser}
\end{figure}

However, existing HOI benchmarks, such as HICO-DET~\cite{chao2018learning} developed before modern VLMs, do not support such unified evaluation. 
Specifically, they count all unmatched predictions as wrong under incomplete annotations, rejecting plausible alternatives.
Many valid alternatives (e.g., ``boarding'' vs. ``exiting'' an airplane in Fig.~\ref{fig: teaser}(a)) are unlabeled, leading to unfair penalization, especially for VLMs whose outputs are more flexible.
As shown in Fig.~\ref{fig: teaser}(a), 
a person mid-motion with an airplane may reasonably be described as ``boarding'' or ``exiting,'' yet only one is labeled as correct. 
The ambiguity, caused by incomplete visual information, penalizes correct predictions and underestimates model capability, especially for VLMs, whose generative outputs can naturally produce multiple equally valid descriptions.
Moreover, in multi-person or multi-object scenes, only a subset of human-object pairs are labeled, leaving other interactions unannotated, as illustrated in Fig.~\ref{fig: teaser}(a) by the person in the blue box.
%
Together, these two types of incomplete annotation, ambiguity from limited visual information and sparse annotation, lead to systematic underestimation and unfair penalization for both paradigms.
Fig.~\ref{fig: teaser}(c) highlights this impact: while HOI-specific methods remain below 50\% mAP, general-purpose VLMs drop to around 15\% when evaluated on HICO-DET.

A straightforward solution might be to exhaustively annotate every possible interaction, thereby eliminating incompleteness. 
However, this approach is not practical because a single image often lacks the visual evidence needed to determine the ground-truth interaction, especially when the view is limited or the action is mid-motion (e.g., boarding vs. exiting in Fig.~\ref{fig: teaser}), making it inherently ambiguous.
Even setting aside this ambiguity, requiring VLMs to select from a fixed label set remains unsuitable. 
As HOI classes scale to thousands of actions and objects, yielding millions of potential interactions, the label space becomes prohibitively large and makes prompt-based prediction impractical.
Moreover, relying on a fixed label set contradicts the open-ended nature of HOI understanding itself, whose goal is to recognize interactions beyond any predefined set of HOI classes, as well as the generative nature of modern VLMs, which naturally express interactions through free-form language rather than discrete class identifiers.

To overcome these limitations, we introduce a new benchmark, \textbf{CrossHOI-Bench}, a unified benchmark for cross-paradigm HOI evaluation across vision-language models and HOI-specific methods. We reformulate HOI detection as a multiple-answer, multiple-choice task, where each question explicitly defines its annotated positives and curated negatives. 
Unlike traditional benchmarks with a fixed label list, our formulation provides a controlled yet adaptive evaluation space, as each question dynamically presents only the relevant candidate interactions. 
This design preserves clearly defined negatives while allowing flexible interpretation for positives, preventing alternative but correct descriptions from being unfairly penalized.

\begin{figure}[t]
	\centering
	\includegraphics[width=0.99\linewidth]{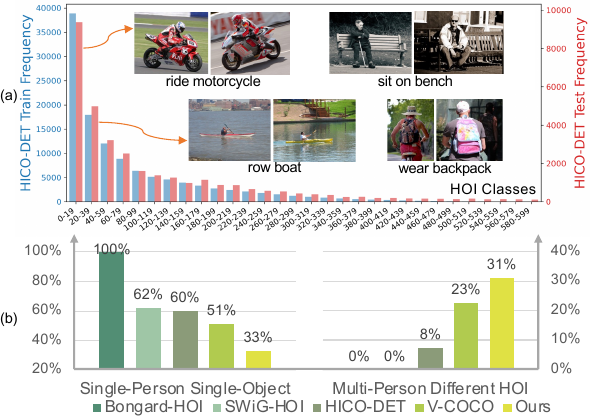}
	\caption{
        (a) HICO-DET test set, showing similar distribution to HICO-DET train set, includes many simple and repetitive scenes for head classes.
        (b) Comparison between existing HOI benchmarks~\cite{bongardhoi,wang2022learning,lin2014microsoft,chao2018learning} and ours. Percentages indicate the proportion of test images that fall into each scenario type (e.g., single-person single-object, multi-person different HOIs).
        %
        }
	
	\label{fig:dataset_compare}
\end{figure}

In addition, existing HOI datasets contain many simple, high-frequency head-class cases, making it difficult to meaningfully assess a model’s true HOI understanding. Fig.~\ref{fig:dataset_compare}(a) illustrates this in HICO-DET through its test-set distribution and several representative head-class examples.
To keep our benchmark both accurate and challenging, we first refine the dataset by excluding overly simple scenarios, such as a single person in a simple background performing a visually obvious action or multiple people engaging in the same interaction without ambiguity. 
From the remaining images, we then apply an automated coarse screening, followed by manual verification that selects the final negative choices and adds any challenging cases missing from the initial screening, producing a benchmark with well-defined negatives and preserved task difficulty for both VLMs and HOI-specific models.
%


Additionally, our benchmark introduces three complementary evaluation settings. 
In some settings, evaluation is focused on the target person or person-object pair HOI detection, ensuring that unannotated interactions involving other individuals do not affect the assessment. 
In other settings, models must identify interactions for all individuals in the scene. To avoid false negatives, we ensure that negative choices exclude any interactions that were previously unannotated.
Together, these settings provide a more comprehensive and controlled evaluation of HOI understanding.
Beyond the main benchmark, we further extend our evaluation to incorporate complementary scenarios.
In particular, we construct additional evaluation sub-benchmarks based on V-COCO~\cite{lin2014microsoft} and SWiG-HOI~\cite{wang2022learning}. 
The V-COCO extension focuses on multi-person scenarios, while the SWiG-HOI extension targets human–human interactions.
These extensions allow us to analyze model behavior across complementary types of interaction scenarios.

Using our benchmarking dataset, we systematically evaluate both standalone VLMs and HOI-specific methods under a unified protocol, enabling direct comparison between the two paradigms. 
Our experiments reveal that large VLMs, even in zero-shot evaluation, sometimes even surpass state-of-the-art HOI-specific methods across most metrics (e.g., achieving +5.18\% Instance-F1). 
Smaller VLMs, however, perform only on par with HOI-specific methods in recognition-only settings, but their performance drops once detection is required.
Furthermore, our analysis highlights the limitations of VLMs: when a human–object pair has several valid HOI labels, they typically predict only one and miss the others, and they frequently misattribute interactions of surrounding people to the target person.

In summary, our contributions are:
\begin{itemize}
    \item 
    We introduce the first benchmarking dataset designed to jointly evaluate standalone general-purpose VLMs and HOI-specific methods. 
    By reformulating HOI detection as a multiple-answer, multiple-choice task with curated negatives, our dataset mitigates the ambiguity and annotation sparsity in existing benchmarks such as HICO-DET. 
    \item 
    We conduct a comprehensive benchmarking study of recent open-source VLMs and HOI-specific methods on our dataset, and observe that large VLMs already achieve state-of-the-art performance on our benchmark without task-specific training. 
	\item 
    Through our analysis, we find that large VLMs demonstrate strong HOI understanding ability, 
   while HOI-specific models show advantages in recognizing multiple co-occurring actions of the same person. Nevertheless, both paradigms still struggle with cross-person attribution, confusing interactions among nearby individuals.
\end{itemize}

\section{Related Work}
\noindent \textbf{HOI Detection Methods }  
HOI detection methods are often divided into two-stage and one-stage approaches.
Two-stage methods first detect humans and objects, and then classify interactions between paired boxes~\cite{zhang2021spatially, zhang2022efficient, park2023viplo,hou2022discovering,wu_posehoi_2024}.  
One-stage methods instead predict $\langle$human, object, verb$\rangle$ triplets directly in an end-to-end manner~\cite{zou2021end,qu2022distillation,li2024neural,Hu_2025_ICCV,wu2024exploring,kim_muren_2023}.  
Despite these advances, existing benchmarks still rely on exact matches with annotated HOIs, implicitly assuming exhaustive labeling of all human–object pairs.  
In practice, annotations are often incomplete.
As a result, plausible but unlabeled interactions are penalized, leading to underestimated performance (e.g., below 50\% mAP on HICO-DET for SOTA HOI methods).  

\noindent \textbf{Existing HOI Benchmarks }  
HICO-DET provides HOI annotations across 600 classes (117 verbs and 80 objects)~\cite{chao2018learning}, with evaluation reported as mAP over all classes.  
V-COCO follows a similar protocol with 29 HOI classes defined on COCO images~\cite{lin2014microsoft}.  
More recently, SWiG-HOI expands this setup to over 5,500 HOI classes for open-vocabulary evaluation~\cite{wang2022learning}.  
Although these benchmarks differ in label space, they all adopt exact-match evaluation, which requires predictions to align strictly with annotated HOI classes.  
This design is problematic because annotations are incomplete, i.e., in ambiguous scenarios and multi-person images, leaving many valid interactions unlabeled.  
Consequently, correct predictions are often penalized.

\noindent \textbf{Vision–Language Models for HOI }  
Large VLMs have recently become the frontier of general-purpose image understanding~\cite{Qwen_2_5_report,InternVL3_2025,liu2024llavanext,GPT_4o_eval_2024,openai_gpt4_2023,Yang_2023_GPT4V_eval}.  
Although not explicitly trained for HOI detection, they exhibit strong open-vocabulary recognition ability, making them natural candidates for HOI understanding.  
However, their performance has not been systematically evaluated in HOI detection.  
Directly applying traditional benchmarks such as HICO-DET or V-COCO with exact-match mAP is unsuitable, as VLMs may generate multiple valid but unlabeled descriptions, causing their performance to be underestimated.

\section{New Benchmark}

\begin{figure*}[t]
	\centering
	\includegraphics[width=0.98\linewidth]{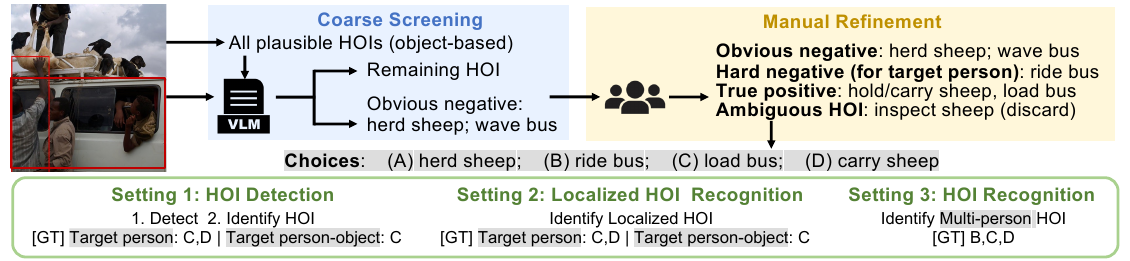}
	\caption{
        Overview of our HOI benchmark construction. Input image undergoes coarse screening and manual refinement to produce a four-choice question, followed by evaluation under three settings.
        }
	
	\label{fig: benchmark_sample}
\end{figure*}

\noindent \textbf{A Unified Benchmarking Dataset}
We build our main benchmark on top of HICO-DET~\cite{chao2018learning}, a widely used dataset for HOI detection~\cite{tamura2021qpic,wu2023end,hou2020visual,yuan2022rlip,Tu_2023_ICCV,ambiguous_hoi,li2022improving,dphoi_2024,lei2024efficient}. 
Our benchmark addresses a critical gap in existing HOI datasets, which cannot support a unified evaluation protocol for comparing VLMs and HOI-specific methods.
Traditional HOI benchmarks are limited by incomplete annotations, caused by lack of information and sparse annotation (more details in supplementary material).
Another issue in existing HOI datasets (i.e., HICO-DET) is the prevalence of simple, repetitive head-class scenes. As shown in Fig.~\ref{fig:dataset_compare}(a), these scenes commonly appear among the test images.
%
In addition, the original train and test splits of HICO-DET share almost identical distributions. We quantify this overlap by computing the KL divergence of the two distributions, which is only 0.088. Such similarity encourages HOI-specific models to bias toward the head classes, whereas our test set reduces this distribution overlap (Sec.~\ref{sec:benchmark_construct}).

Our benchmark is carefully curated to mitigate these issues. 
We reformulate HOI detection as a multiple-choice task with curated hard negatives to prevent incorrect penalization from incomplete annotations, and we include a dedicated detection-required setting to evaluate full HOI detection alongside additional settings for broader HOI understanding.
By removing overly simple head-class scenes (See Fig.~\ref{fig:dataset_compare}(b)), our test distribution becomes less tied to HICO-DET’s training priors, where the original train and test splits share nearly identical, heavily head-dominated distributions, thus exposing model biases that existing benchmarks cannot reveal.
As a result, although our dataset is smaller than existing HOI datasets, it provides a focused testbed for analyzing the relative strengths and weaknesses of VLMs and HOI-specific models, especially on challenging scenarios such as subtle interaction distinctions, multi-person scenes, and human–human interactions.


\subsection{Dataset Construction}
\label{sec:benchmark_construct}
\noindent \textbf{Task Reformulation }
To address the limitations of existing HOI benchmarks, we reformulate HOI detection as a multiple-answer, multiple-choice task. 
For a human-object pair in an image, we construct a question with four candidate options. 
Since a person may simultaneously engage in multiple valid interactions (e.g., \emph{hold knife} and \emph{cut with knife}), a question can include more than one positive answer. 
In our benchmark, positive choices are taken from HICO-DET ground-truth annotations, with unannotated yet visually clear interactions added manually as additional positives. Curated negatives exclude plausible but unlabeled interactions.
This process reduces the likelihood of penalizing valid predictions and provides a unified evaluation protocol for both VLMs and HOI-specific methods.

\noindent \textbf{Coarse Screening}  
This stage serves only as a coarse, automatically generated screening step to obtain negative candidates before manual refinement.
For a human-object pair in an image, we first build a candidate pool by collecting all plausible actions for the object and removing its ground-truth labels. 
To reduce false negatives caused by incomplete annotations, we apply a multi-stage VLM-based pipeline. 
Because a single VLM often struggles with negative choice judgments, we combine multiple VLMs in sequence to improve robustness.
GPT-4.1~\cite{openai_gpt4_2023} initially separates candidates into semantically consistent or inconsistent with the image, and we retain only the inconsistent ones.  
We then evaluate each remaining candidate using Qwen2.5-VL-32B~\cite{Qwen_2_5_report} and GPT-4o~\cite{openai_gpt4o_2024}.
A candidate is kept only when both models agree that it is an incorrect HOI compared with the positive choices, reducing false negatives.
%

\noindent \textbf{Manual Refinement}
Although the coarse screening provides candidate negatives, it often removes hard cases, making the benchmark not sufficiently challenging.
Moreover, the original HOI test dataset contains a large proportion of simple cases, such as single-person single-object interactions in simple backgrounds or multiple people performing the same action unambiguously. 
To create a more challenging benchmark, we introduce a manual refinement stage, performing manual checks on every question (i.e., every image in the benchmark).
This manual step corrects errors introduced during coarse screening by re-adding valid interactions that were mistakenly filtered out and recovering missed hard cases.
We first remove these overly simple scenes from our dataset.
From the remaining images, we strengthen the benchmark in two ways. 
First, we include harder positives by allowing multiple plausible actions to be correct (e.g., both \emph{boarding} and \emph{exiting} as shown in Fig.~\ref{fig: teaser}). 
Second, we add hard negatives, including interactions of surrounding people that differ from those of the target person (e.g., \emph{riding a bus} in Fig.~\ref{fig: benchmark_sample}), and fine-grained distinctions between visually similar actions (e.g., \emph{holding} vs. \emph{hugging a person}). 
This refinement process increases the difficulty of the benchmark while maintaining the validity of negative choices, ensuring that the benchmark captures the complexity of real-world HOI understanding.

\noindent \textbf{Redistributed Test Set}
As illustrated in Fig.~\ref{fig:dataset_compare}(a), the training and test distributions of HICO-DET are highly similar, which risks inflating performance by allowing a model to exploit dataset priors rather than demonstrating genuine HOI understanding~\cite{Agrawal_2018_prior_vqa}. 
After removing overly simple and repetitive scenes during manual refinement, the remaining test images naturally form a distribution that is more distinct from the training set.
The KL divergence between the HICO-DET training and test splits is only 0.088, whereas the divergence between the training split and our redistributed test set increases to 0.629. 

\noindent \textbf{Dataset Summary}
In total, our benchmarks include 3,773 multiple-choice questions with one main benchmark and two complementary ones. Our main benchmark contains 1,274 images from the HICO-DET test set, involving 600 HOI classes.
We curate a challenging, less head-dominated benchmark by removing overly simple cases and redistributing class frequencies, while keeping coverage across 600 classes.
We keep the main benchmark HICO-DET-based rather than merging multiple HOI datasets to support evaluation for both HOI-specific methods and VLMs, since the former cannot be consistently evaluated across datasets with different predefined HOI classes. 
All three evaluation settings (\textit{Setting 1–3}, introduced in Sec.~\ref{sec: benchmarking_eval}) are based on the same number of images.
Each question follows a fixed four-choice format with randomized option order, allowing multiple correct answers. Randomization prevents positional bias, as large language models (LLMs) are known to be sensitive to option ordering~\cite{pezeshkpour-hruschka-2024-large, zheng2024large}.

Beyond the main benchmark, we construct two complementary benchmarks based on V-COCO~\cite{lin2014microsoft} and SWiG-HOI~\cite{wang2022learning} with the same construction and refinement process, focusing on multi-person and human–human interaction scenarios. 
The V-COCO-based sub-benchmark emphasizes multi-person interactions, where multiple individuals in close proximity perform distinct actions within the same scene. 
It contains 647 questions, derived from 499 images, covering 323 HOI classes.  
The SWiG-HOI-based sub-benchmark focuses on human–human interactions, where both participants are humans and recognition often relies on contextual cues rather than object affordance, such as ``encourage person'' and ``complain person''. 
It includes 1,852 questions from 1,851 images, with 210 unique HOI classes in total. 
These sub-benchmarks are provided as complementary to the main benchmark, where they evaluate models in different scenarios and assess out-of-distribution generalization for models finetuned on HOI datasets.
For completeness, we also provide the fine-tuning data constructed from HICO-DET with details in the supplementary material, but the focus in our paper is zero-shot VLM evaluation and unified comparison with HOI-specific methods.

\begin{table*}[t]
\centering
\resizebox{1.9\columnwidth}{!}{
\begin{tabular}{l|c|c|c|c|c|c}
\toprule
{Method}
& Macro-F1 ($\%$)   
& Instance-F1 ($\%$)
& Micro-F1 ($\%$)
& EM ($\%$)
& Avg. Prec. ($\%$)
& Avg. Rec. ($\%$) \\
\midrule
\multicolumn{7}{l}{\textit{HOI-specific methods}} \\
\midrule
ADA-CM~\cite{lei2023efficient}  &  43.02 & \textbf{47.76} & \textbf{61.69} & 19.15 & 76.25 & 51.80 \\
CMMP~\cite{lei2024exploring}   &  43.06 & 46.62 & 60.85 & 18.84 & 75.06 & 51.16 \\
LAIN~\cite{KimJC_CVPR_2025}  &  41.28 & 45.64 & 59.09 & 19.31 & 73.42 & 49.44  \\
HOLa~\cite{lei2025lhola}   &  43.61 & 47.12 & 61.29 & 19.78 & 74.31 & \textbf{52.15} \\
CMD-SE~\cite{Lei_2024_CVPR}  & \textbf{47.49} & 44.66 & 58.71 & \textbf{20.33} & \textbf{78.33} & 46.96 \\
\midrule
\multicolumn{7}{l}{\textit{VLM zero-shot evaluation}} \\
\midrule
InternVL2.5-38B~\cite{chen2024internvl} & 22.44 & 19.97 & 29.05 & 9.18 & 85.18 & 17.51 \\
InternVL3-38B~\cite{InternVL3_2025}  & 38.04 & 38.68 & 48.08 & 20.33 & \textbf{84.72} & 33.56  \\
{Qwen3-VL-30B-Instruct}~\cite{Qwen_2_5_report} & 29.10 & 0.25 & 0.51 & 0.08 & 100.0 & 0.26 \\
Qwen2.5-VL-32B~\cite{Qwen_2_5_report}  & \textbf{50.71} & \textbf{52.94} & \textbf{61.41} & \textbf{26.06} & 75.03 & \textbf{51.97} \\
\midrule
LLaVA-OV-7B~\cite{llavaov_tmlr_2025} &  -&-&-&-&-&-  \\
InternVL3-8B~\cite{InternVL3_2025}   &  6.58 & 4.96  & 8.51 & 2.04 & \textbf{76.09} & 4.51 \\
Qwen2-VL-7B~\cite{Qwen2-VL}  & -& -& -& -&- & -    \\
Qwen2.5-VL-7B~\cite{Qwen_2_5_report}   & \textbf{29.73} & \textbf{30.53} & \textbf{37.49} & \textbf{14.29} & 75.92 & \textbf{24.89} \\
\bottomrule
\end{tabular}
}
\caption{\textit{Setting~1} experiment results comparison. Results are grouped by VLM and HOI-specific methods. Best performance within each group is highlighted in \textbf{bold}. ``Avg. Prec.'' and ``Avg. Rec.'' denote the precision and recall averaged across the test set, respectively.}
\label{tab:new_setting1_grouped}
\end{table*}

\subsection{Benchmark Evaluation}
\label{sec: benchmarking_eval}
\noindent \textbf{Evaluation Settings }
We design three evaluation settings that capture different aspects of HOI evaluation, all under the same multiple-choice format (see Fig.~\ref{fig: benchmark_sample}, with additional examples in supplementary materials).
\textbf{Setting 1} evaluates the full HOI detection pipeline.
A model must first perform object detection and then recognize the interactions for the detected person overlapping with the ground-truth target (IoU $\ge$ 0.5), where the ground-truth box is used only for evaluation. This setting evaluates the core HOI detection capability, combining detection and interaction recognition, which is compatible with existing HOI evaluation.
\textbf{Setting 2} is a diagnostic extension of \textit{Setting 1} that requires localized HOI recognition: the bounding box of the target person is provided, and the model must predict that individual’s interactions.
This removes detection error and isolate recognition.
Many HOI methods cannot be evaluated with box input due to architectural constraints (end-to-end detection pipelines or reliance on detector embeddings). However, \textit{Setting 2} is not intended to claim a paradigm gap, which is assessed in \textit{Setting 1}, but for analyzing the localization bottleneck of VLMs.
Such a setting is useful in downstream applications when person localization is provided by upstream modules, such as tracking, egocentric perception, video action recognition, or embodied AI~\cite{gu2018ava,sun2018actor,feichtenhofer2019slowfast,grauman2022ego4d,pmlr-v205-li23a,lei2024few,sun2019deep,cheng20203d,cheng2021graph}.
\textbf{Setting 3} evaluates image-level HOI recognition across multiple persons.
This setting focuses on identifying correct HOI labels across multiple individuals.
This setting isolates global HOI understanding from localization, which is useful for downstream tasks that require holistic interaction reasoning, such as fine-grained social interaction detection~\cite{kim2025partaware} or multi-person interaction understanding~\cite{wei2024nonverbal}.

By default, both \textit{Setting 1} and \textit{Setting 2} adopt a \textit{human-centered} formulation, where only human bounding boxes are provided as targets, and all interactions involving the human are treated as ground truth.
The interacted objects are implicitly inferred from the predicted interactions.
For completeness, we also include a \textit{human–object-centric} variant, where human and object bounding boxes are jointly provided as target pairs, enabling complementary evaluation when explicit object detection is required. Discussions for human–object-centric results are included in supplementary material.

\noindent \textbf{Evaluation Metrics }
To evaluate multiple-answer, multiple-choice questions, 
we adopt set-based metrics that directly compare predicted and ground-truth label sets. 
These metrics are widely used in multi-label classification and question answering~\cite{emnlp_RajpurkarZLL16, ICML_WuZ_2017}. 
Specifically, we report Instance-F1, Macro-F1, Micro-F1, and Exact Match Accuracy (EM), which offer a comprehensive evaluation.
Macro-F1 balances across classes, Instance-F1 captures per-question performance, Micro-F1 measures overall aggregate performance, and EM reflects exact prediction correctness (details in supplementary material).
%
%
While F1 captures the precision–recall balance globally, it does not reveal whether performance variations stem more from precision or recall. Thus, we also report average precision and recall across all questions to examine them separately.

\section{Experiments}

\subsection{Experiment Setup}
\noindent \textbf{Baselines }
We evaluate two groups of baselines on our benchmark: general-purpose VLMs and HOI-specific methods.
Recent large VLMs represent the frontier of general-purpose image understanding.  
Although not explicitly trained for HOI detection, they exhibit strong open-vocabulary grounding as well as visual and spatial reasoning abilities, making them natural candidates for HOI evaluation.  
Our VLM baselines include Qwen2/2.5/3-VL~\cite{Qwen_2_5_report}, InternVL2.5/3~\cite{chen2024internvl, wang2024mpo, InternVL3_2025}, and LLaVA-OV~\cite{llavaov_tmlr_2025}.  
For HOI-specific methods, we include ADA-CM~\cite{lei2023efficient}, CMMP~\cite{lei2024exploring}, LAIN~\cite{KimJC_CVPR_2025}, HOLa~\cite{lei2025lhola}, and CMD-SE~\cite{Lei_2024_CVPR}, which report competitive results on HICO-DET~\cite{chao2018learning} and SWiG-HOI~\cite{wang2022learning}.  
%
%
Our experiments focus on zero-shot evaluation of VLMs. We also report supervised fine-tuning results for Qwen2.5-VL-7B in the supplementary material for completeness.

\noindent \textbf{Implementation Details }
For general-purpose VLMs, we provide each question as a prompt together with explicit answer-format instructions. Model outputs are parsed accordingly and evaluated against the ground truth.  
For HOI-specific methods, we follow a top-$k$ matching strategy. Specifically, we take the top-5 predictions for each question and check whether any match the provided choices, consistent with the standard top-5 evaluation used in ImageNet~\cite{deng2009imagenet}. We adopt top-5 rather than top-1 to account for multiple correct answers per question, while top-10 would be unnecessarily permissive.  
%
More implementation details are included in supplementary materials.

\subsection{Experiment Results}

\begin{table*}[t]
\centering
\resizebox{1.9\columnwidth}{!}{
\begin{tabular}{l|c|c|c|c|c|c}
\toprule
{Method}
& Macro-F1 ($\%$)   
& Instance-F1 ($\%$)
& Micro-F1 ($\%$)
& EM ($\%$)
& Avg. Prec. ($\%$)
& Avg. Rec. ($\%$) \\
\midrule
\multicolumn{7}{l}{\textit{VLM zero-shot evaluation}} \\
\midrule
InternVL2.5-38B~\cite{chen2024internvl} & 48.43 & 46.43 & 51.56 & 20.64 & 77.52 & 38.63 \\
InternVL3-38B~\cite{InternVL3_2025}  &  58.94 & 67.41 & 67.81 & \textbf{35.64} & \textbf{81.90} & 57.85 \\
{Qwen3-VL-30B-Instruct}~\cite{Qwen_2_5_report} & 56.43 & 64.00 & 64.59 & 32.57 & 79.89 & 54.21 \\
Qwen2.5-VL-32B~\cite{Qwen_2_5_report}  & \textbf{62.90} & \textbf{69.52} & \textbf{70.69} & 35.01 & 75.30 & \textbf{66.61} \\
\midrule
LLaVA-OV-7B~\cite{llavaov_tmlr_2025} & 47.76 & 56.53 & 54.80 & 25.12 & \textbf{77.43} & 42.40 \\
InternVL3-8B~\cite{InternVL3_2025}   & \textbf{49.88} & 52.35 & 55.54 & 23.86 & 74.41 & 44.31  \\
Qwen2-VL-7B~\cite{Qwen2-VL} & 46.90 & 53.93 & 53.61 & 23.23 & 76.84 & 41.16  \\
Qwen2.5-VL-7B~\cite{Qwen_2_5_report} & 48.93 & \textbf{57.25} & \textbf{57.53} & \textbf{25.98} & 74.49 & \textbf{46.87} \\
\bottomrule
\end{tabular}
}
\caption{\textit{Setting~2} experiment results comparison. Best performance within each group is highlighted in \textbf{bold}. ``Avg. Prec.'' and ``Avg. Rec.'' denote the precision and recall averaged across the test set, respectively.}
\label{tab:new_setting2_grouped}
\end{table*}


\begin{table*}[t]
\centering
\resizebox{1.9\columnwidth}{!}{
\begin{tabular}{l|c|c|c|c|c|c}
\toprule
{Method}
& Macro-F1 ($\%$)   
& Instance-F1 ($\%$)
& Micro-F1 ($\%$)
& EM ($\%$)
& Avg. Prec. ($\%$)
& Avg. Rec. ($\%$) \\
\midrule
\multicolumn{7}{l}{\textit{HOI-specific methods}} \\
\midrule
ADA-CM~\cite{lei2023efficient}  & 45.89 & 56.23 & \textbf{67.49} & 11.85 & \textbf{83.16} & \textbf{56.78} \\
CMMP~\cite{lei2024exploring}   &  46.07 & 55.42 & 67.16 & 10.83 & 82.54 & 56.61 \\
LAIN~\cite{KimJC_CVPR_2025}   &  44.27 & 53.87 & 65.06 & 10.52 & 80.61 & 54.54 \\
HOLa~\cite{lei2025lhola}   &  46.37 & 55.91 & 67.06 & 11.54 & 81.88 & \textbf{56.78} \\
CMD-SE~\cite{Lei_2024_CVPR} &  \textbf{46.51} & \textbf{57.25} & 66.49 & \textbf{14.13} & 83.05 & 55.44 \\
\midrule
\multicolumn{7}{l}{\textit{VLM zero-shot evaluation}} \\
\midrule
InternVL2.5-38B~\cite{chen2024internvl} & 51.96 & 51.81 & 55.31 & 17.27 & 84.47 & 41.11 \\
InternVL3-38B~\cite{InternVL3_2025}  & 58.23 & 63.17 & 63.28 & 22.68 & \textbf{87.06} & 49.71 \\
{Qwen3-VL-30B-Instruct}~\cite{Qwen_2_5_report} & 59.14 & 66.71 & 66.98 & 23.31 & 86.68 & 54.57 \\
Qwen2.5-VL-32B~\cite{Qwen_2_5_report}  & \textbf{63.16} & \textbf{67.19} & \textbf{68.82} & \textbf{23.16} & 77.83 & \textbf{61.68} \\
\midrule
LLaVA-OV-7B~\cite{llavaov_tmlr_2025} & 46.47 & 54.05 & 52.18 & 12.95 & \textbf{85.28} & 37.59 \\
InternVL3-8B~\cite{InternVL3_2025}   & \textbf{55.52} & \textbf{61.17} & \textbf{61.57} & \textbf{20.41} & 83.51 & \textbf{48.76}  \\
Qwen2-VL-7B~\cite{Qwen2-VL}  & 41.92 & 36.91 & 41.64 & 7.14 & 82.75 & 27.82 \\
Qwen2.5-VL-7B~\cite{Qwen_2_5_report} & 51.29 & 56.04 & 56.03 & 14.84 & 80.15 & 43.06   \\
\bottomrule
\end{tabular}
}
 \caption{\textit{Setting~3} experiment results comparison. Results are reported for VLMs and HOI-specific methods. Best performance within each group is highlighted in \textbf{bold}. ``Avg. Prec.'' and ``Avg. Rec.'' denote the precision and recall averaged across the test set, respectively.}
\label{tab:new_setting3_grouped}
\end{table*}

As shown in Table~\ref{tab:new_setting1_grouped},~\ref{tab:new_setting2_grouped},~\ref{tab:new_setting3_grouped}, the results across all three settings show that large VLMs (e.g., Qwen2.5-VL-32B, InternVL3-38B) achieve the best overall performance, often surpassing HOI-specific models by a clear margin in Instance-F1, Macro-F1, and EM accuracy. 
%
%
In Table~\ref{tab:new_setting1_grouped}, no external object detector is used. VLMs predict boxes and HOI answers themselves, so performance drops compared to Table~\ref{tab:new_setting2_grouped} reflect the standalone VLM’s limits. Detector-assisted results and discussions are provided in the supplementary materials (Table 5), suggesting localization is the main bottleneck. 
%
HOI-specific models maintain competitive performance in \textit{Setting 1}, likely due to their structured detection pipelines, showing decent localization despite lower recognition accuracy. 
Abnormal results (e.g., low Instance-F1 but 100\%
Avg. Prec.) in Table~\ref{tab:new_setting1_grouped} are caused by near-zero detection recall. Some VLMs output few valid person/object boxes, missing most GT HOIs and driving F1 down. With valid box, they can still be correct. Avg. Prec. is computed over the tiny predicted set, thus reaching 100\%, showing precision alone is uninformative under near-zero recall, not a metric issue.  Missing entries (``–'') indicate that LLaVA-OV-7B and Qwen2-VL-7B produced no valid boxes, so \textit{Setting-1} metrics cannot be computed.
%
%
In Table~\ref{tab:new_setting3_grouped}, EM accuracy of VLMs decreases by 5-10 in \textit{Setting 3} compared with that in \textit{Setting 2}, indicating the difficulty in accurately identifying multi-person interactions. 
%
%

\input{figures/VLM_problems_main}
\vspace{0.3cm}
\noindent \textbf{Findings: Analysis of Failure Cases}
We further analyzed failure cases of VLMs, and compared with the HOI-specific models.
These cases are grouped into four main categories: incomplete multi-action recognition, cross-person HOI misattribution, HOI similarity confusion and hallucinated HOI inference. 
We focus on analysis for the first two categories, with additional details provided in supplementary material.

\noindent \textbf{Incomplete multi-action recognition}
The failure occurs when the VLM predicts only one of multiple valid actions for the same human–object pair.
As shown in Fig.~\ref{fig:VLM_problem_main}(a), the person is jumping and holding a surfboard, but the model recognizes only {jumping} while omitting {holding}. 
This suggests that the VLM tends to capture the most salient action while ignoring concurrent or secondary interactions. 
This can be further validated by comparing the average precision and recall in Table~\ref{tab:new_setting1_grouped}-\ref{tab:new_setting3_grouped}, 
where VLMs consistently exhibit high precision but notably lower recall, especially small VLMs whose recall is about 30 percentage points lower than their precision, indicating that they often predict only the primary actions while missing co-occurring ones. 
However, HOI-specific models are less affected by this issue, largely because their training data include numerous co-occurring action annotations.
Their lower average recall compared to precision arises from a different cause: in complex scenes, HOI-specific models often produce low-confidence predictions, thus missing the positive choices.

\noindent  \textbf{Cross-person HOI misattribution}
VLMs often misattribute the interaction of a surrounding person to the target person.
When multiple people appear in close proximity performing different actions, the global attention mechanism likely mixes the features of nearby people and assigns the wrong action to the target person.
This limitation is evident when comparing \textit{Setting~2} (Table~\ref{tab:new_setting2_grouped}) and \textit{Setting~3} (Table~\ref{tab:new_setting3_grouped}), which use the same images but differ in scope: \textit{Setting~2} isolates the target individual, whereas \textit{Setting~3} requires recognizing interactions for all people.  
Across models, we observe clear performance drops when moving from \textit{Setting~3} to \textit{Setting~2} (e.g., average precision decreases by 9.1\% for InternVL3-8B, 5.7\% for Qwen2.5-VL-7B, 5.2\% for InternVL3-38B, and 2.5\% for Qwen2.5-VL-32B). Error analysis confirms that 20–25\% of mispredictions come from attributing interactions of surrounding people to the target person (e.g., 21.6\% for InternVL3-8B, 24.6\% for Qwen2.5-VL-7B, 22.3\% for InternVL2.5-38B, 24.2\% for InternVL3-38B, and 22.0\% for Qwen2.5-VL-32B).  
These results highlight a persistent weakness of VLMs in disentangling individual-level interactions in multi-person scenarios. Additional qualitative examples are provided in supplementary material.
As shown in Fig.~\ref{fig:VLM_problem_main}(b), the model incorrectly predicts throwing sports ball for the defending player whose ground-truth interaction is blocking sports ball.
HOI-specific models alleviate such confusion by preserving spatial topology, but overlapping human features are still entangled, causing cross-person misattribution.



\begin{figure}[t]
	\centering
	\includegraphics[width=\linewidth]{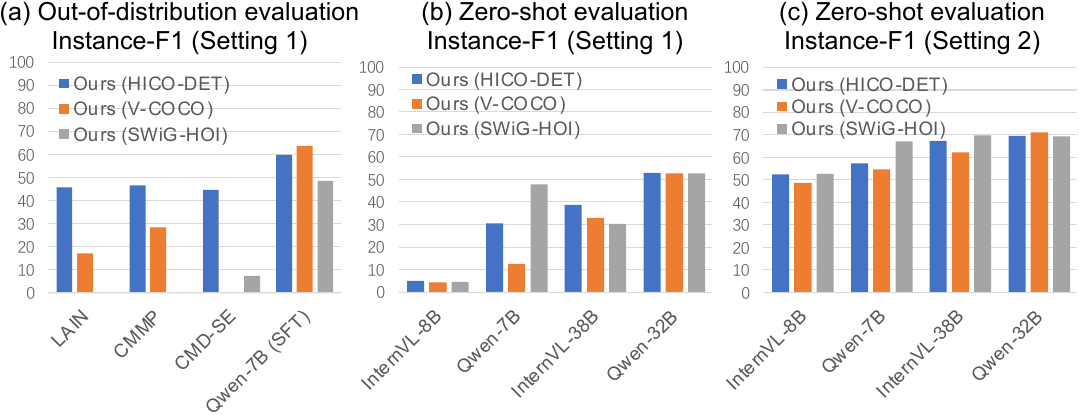}
	\caption{
        Evaluation on our HICO-DET-based, V-COCO-based and SWiG-HOI-based sub-benchmarks in \textit{Setting 1} and \textit{2}. ``InternVL'' refers to InternVL3 and ``Qwen'' refers to ``Qwen2.5-VL''.  
        %
        }
	\label{fig:crossdataset}
\end{figure}

\vspace{0.3cm}
\noindent \textbf{Multi-Dataset Evaluation}
Fig.~\ref{fig:crossdataset} compares the performance on our HICO-DET-, V-COCO-, and SWiG-HOI-based benchmarks under \textit{Setting~1} and \textit{Setting~2}. 
For HOI-specific methods such as CMMP and CMD-SE, performance drops noticeably on V-COCO- and even more on SWiG-HOI-based sub-benchmarks, shown in Fig.~\ref{fig:crossdataset}(a), indicating limited out-of-distribution generalization. 
The decline is particularly noticeable on SWiG-HOI, where most interaction categories differ from those on HICO-DET. 
By contrast, the drop on V-COCO is milder, as a subset of interactions are shared across the two datasets. 
Overall, HOI-specific models remain highly sensitive to domain and label distribution shifts. 
In contrast, the fine-tuned VLM (e.g., Qwen2.5-VL-7B~(SFT on the HICO-DET-based train set)) maintains stable performance across benchmarks. 
Qwen2.5-VL-32B in Fig.~\ref{fig:crossdataset}(b) and (c), exhibit consistent zero-shot generalization across benchmarks. Small VLMs, i.e., Qwen2.5-VL-7B, struggle in V-COCO sub-benchmark in \textit{Setting 1}, and present better generalization in \textit{Setting 2}, indicating their limited detection ability especially with multiple humans and objects.

\subsection{Discussions}

\begin{figure}[t]
	\centering
	\includegraphics[width=\linewidth]{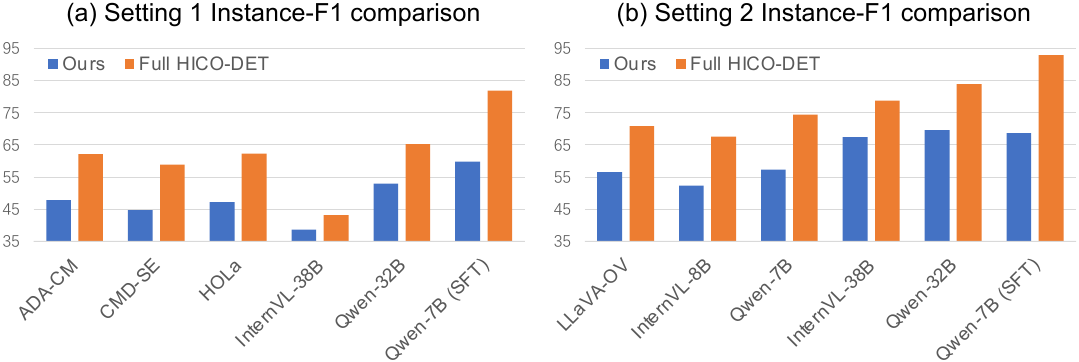}
	\caption{
        Experiment result comparison between our CrossHOI-Bench and full HICO-DET based dataset in \textit{Setting 1} and \textit{2}. ``InternVL'' refers to InternVL3 and ``Qwen'' refers to ``Qwen2.5-VL''. 
        %
        }
	\label{fig:all_hico_cmp}
\end{figure}

\noindent \textbf{Challenging Benchmarking Dataset}
To demonstrate our benchmark is more challenging than simply using the full test set of HICO-DET, we compare the performance of HOI-specific methods and VLMs on our main benchmark and the full HICO-DET based dataset as shown in Fig.~\ref{fig:all_hico_cmp}.
Despite containing fewer images than the full HICO-DET test set, both VLMs and HOI-specific models show significant performance drops, confirming that our benchmark indeed focuses on challenging HOI cases.

\noindent \textbf{The Use of VLMs in Coarse Screening}
Although VLMs are used in the coarse screening process, Table \ref{tab:easy_neg_agree} shows HOI-specific methods agree with the coarse-screening identified negatives even more than their VLM counterparts ($\sim$ 99\% vs. 95–97\%), demonstrating the selected candidate negatives are not biased toward VLMs. 
In practice, we select HOI-specific predictions with confidence scores below 0.2 as reliable negative cases, which the model considers unlikely to appear.
These identified negatives are simple cases for both HOI-specific methods and VLMs, while challenging ones come from the subsequent manual refinement.

\begin{table}[t]
\centering
\resizebox{0.9\columnwidth}{!}{
\begin{tabular}{@{}l c l c@{}}
\toprule
{Method} & {Agreement} & {Method} & {Agreement}\\
\midrule
ADA-CM           & 0.9893 & CMMP             & 0.9889 \\
HOLa             & 0.9880 & LAIN             & 0.9879 \\
CMD-SE           & 0.9705 & Qwen2.5-VL-32B   & 0.9568 \\
InternVL3-38B    & 0.9770 & InternVL2.5-38B  & 0.9731 \\
Qwen2.5-VL-7B    & 0.9761 & InternVL3-8B  & 0.9583 \\
\bottomrule
\end{tabular}
}
\caption{Validation of Unbiased Coarse Screening}
\label{tab:easy_neg_agree}
\end{table}



\section{Conclusion}
In this work, we revisit HOI evaluation in the era of large VLMs. Existing benchmarks such as HICO-DET suffer from incomplete annotations in two perspectives: limited visual information and sparse annotations, both of which lead to underestimated performance, particularly for VLMs. 
To address this, we introduce the CrossHOI-Bench benchmark that reformulates HOI detection as a multiple-answer, multiple-choice task with curated negatives, avoiding valid but unlabeled interactions from being penalized. We further design three complementary evaluation settings to assess different aspects of HOI understanding.
Our experiments show that large VLMs achieve competitive, sometimes superior performance compared with HOI-specific methods, while smaller models are competitive mainly when object detection is provided. At the same time, VLMs still struggle with multi-action recognition and cross-person attribution.
Overall, our benchmark provides a unified evaluation protocol that supports progress in both specialized HOI methods and general-purpose VLMs.

\section{Acknowledgement}
This document is supported by the Ministry of Education, Singapore,
under its MOE AcRF Tier 3 Grant (MOE-MOET32022-0001).

{
    \small
    \bibliographystyle{ieeenat_fullname}
    \bibliography{main}
}


\clearpage

\appendix
\section*{Appendix}
\addcontentsline{toc}{section}{Appendix}

\section{Benchmarking Dataset Discussion}

\subsection{Limitations of the HICO-DET Benchmark}

\noindent \textbf{Incomplete Annotation}
To understand why existing HOI benchmarks fail to provide reliable evaluation for both VLMs and HOI-specific models, we manually check the annotation quality of HICO-DET.
Through this inspection, we identify two major issues: \emph{information incompleteness} and \emph{annotation sparsity}.
These issues significantly impact evaluation correctness, especially for VLMs whose predictions may be valid but absent from the ground-truth annotations.

The first type, information incompleteness, appears when the visual or temporal evidence in a static image is insufficient to determine the correct interaction, leading to two issues in existing HOI benchmarks' annotation.
One issue is that the available visual evidence suggests a reasonable interaction, but does not provide enough visual information to confirm it.
Fig.~\ref{fig:hico_dataset_problem}(a) is an example, where inspecting a hair dryer is a plausible interaction, but the static image alone is not enough to verify this.
Because such interactions are plausible but visually unconfirmed, they are often considered as negatives in HICO-DET benchmark, although they are potentially valid interpretations.
The other issue is the inherently ambiguous temporal states (e.g., \textit{catch} vs. \textit{throw frisbee} in Fig.~\ref{fig:hico_dataset_problem}(c)).
These cases arise when multiple actions are visually indistinguishable in a single frame and only one of the plausible interactions is annotated (e.g., HICO-DET).
In our manual review of 200 randomly sampled images from the HICO-DET test set, 29 exhibited insufficient visual evidence and 12 exhibited ambiguous temporal states (combined accounts for 20.5\% of the sampled images). These examples illustrate that such ambiguities occur non-trivially in the existing HOI benchmark dataset, thus, affecting the evaluation when these images and their ground truths are used.

\input{figures/dataset_dist}

The second type, annotation sparsity, refers to missing or partial labels in multi-person scenes or within individual human–object pairs. In multi-person or multi-object cases, annotations are often provided for only a subset of individuals or objects, leaving other valid interactions unlabeled. As shown in Fig.~\ref{fig:hico_dataset_problem}(d), the female sitting on the couch is annotated, while the male on the adjacent couch is not. In Fig.~\ref{fig:hico_dataset_problem}(e), \textit{cut cake} is labeled but \textit{cut with knife} is missing. 
Sparsity also occurs within a single human–object pair when multiple concurrent or semantically related actions are possible but only a subset is annotated. For example, in Fig.~\ref{fig:hico_dataset_problem}(f), \textit{flip skateboard} is labeled, while other plausible actions such as \textit{jump skateboard} are not labeled. 
In our 200-image check, 58 images (29\%) contained unannotated persons or objects, and 49 images (24.5\%) contained incomplete action labels within annotated human–object pairs. These observations highlight the prevalence of sparse annotations that treat valid interactions as negatives, leading to false negatives and incorrect evaluation.

\input{figures/dataset_issue}

Overall, these two categories reveal that HICO-DET benchmark suffers from two fundamental sources of annotation noise: {information incompleteness} and {annotation sparsity}. Together, they introduce evaluation bias and hinder fair comparison across methods, highlighting the need for a  benchmarking dataset that enables more reliable HOI evaluation.

\begin{table*}[t]
\centering
\resizebox{1.55\columnwidth}{!}{
\begin{tabular}{l|c|c|c |c|c  }
\toprule
Dataset & \makecell{Single-Person\\Single Obj} \textcolor{blue}{$\downarrow$} & \makecell{Multi-Person \\ Diff. HOI }  \textcolor{blue}{$\uparrow$} &  \makecell{Applicable to \\ HOI methods} & \makecell{Applicable \\ to VLMs} & \makecell{Multi-Class \\ HOI Prediction} \\
\midrule
HICO-DET~\cite{chao2018learning} &60.2\%  & 7.5\% & {\color{checkmark}\ding{51}} & {\color{red}\ding{55}} & {\color{checkmark}\ding{51}}\\
V-COCO~\cite{lin2014microsoft}  &51.2\%  & 22.5\% & {\color{checkmark}\ding{51}} & {\color{red}\ding{55}} & {\color{checkmark}\ding{51}}\\
SWiG-HOI~\cite{wang2022learning}  & 62.2\%  & 0.0\% & {\color{checkmark}\ding{51}} & {\color{red}\ding{55}} & {\color{checkmark}\ding{51}}\\
Bongard-HOI~\cite{bongardhoi} & 100.0\%  & 0.0\% & {\color{checkmark}\ding{51}} & {\color{checkmark}\ding{51}} & {\color{red}\ding{55}} \\
\midrule
Ours (main benchmark)  & {33.1\%} & {31.2\%} & {\color{checkmark}\ding{51}}  &  {\color{checkmark}\ding{51}} &  {\color{checkmark}\ding{51}}\\
Ours (main+ sub- benchmarks)  & \textbf{11.2\%} & \textbf{67.5\%} & {\color{checkmark}\ding{51}}  &  {\color{checkmark}\ding{51}} &  {\color{checkmark}\ding{51}}\\
\bottomrule
\end{tabular}}
\caption{Comparison between existing HOI benchmarks and ours. }
\label{tab:benchmark_comparison}
\end{table*}

\noindent \textbf{Similar Train and Test Splits Distribution}
Another issue with HICO-DET is the nearly identical distribution between its training and test splits (Fig.~\ref{fig:dataset_dist}(a)). The KL divergence between the two splits is only 0.088, indicating that the test set closely mirrors the training distribution. As a result, models trained on this distribution may exhibit artificially high performance.
In contrast, the KL divergence between the training split and our redistributed test set increases to 0.629, introducing meaningful variation without altering the inherent long-tail nature of HOI classes in the real world. Our goal is not to impose an artificial or unrealistic class distribution, but to avoid a test set that is effectively a resampled copy of the training distribution, thereby enabling a more reliable evaluation.

\subsection{Comparison with Existing HOI Benchmarks}

\input{figures/remove_imgs}

In Table~\ref{tab:benchmark_comparison}, we provide a systematic comparison of our benchmark with existing HOI datasets across multiple dimensions.
In HICO-DET~\cite{chao2018learning}, the majority of cases (60.2\%) involve a single person interacting with a single object, which often results in relatively easy recognition. Our main benchmark, constructed from HICO-DET, reduces this proportion to 33.1\%, thereby shifting the focus toward more challenging multi-person scenarios. 
In addition, a key strength of our benchmark is the inclusion of multi-person images with different interactions. While only 7.5\% of HICO-DET and 22.5\% of V-COCO~\cite{lin2014microsoft} contain such cases, our dataset increases this proportion to 31.2\%, providing a richer evaluation of compositional reasoning across individuals. SWiG-HOI~\cite{wang2022learning} and Bongard-HOI~\cite{bongardhoi} contain no such cases, because they only provide one annotated person for each image, in contrast to our benchmark’s focus on multi-person HOI scenarios. 
Beyond the main benchmark, we create two sub-benchmarks focusing on multi-person and human-human interaction scenes. On the combined main and two sub-benchmarks, the proportion of single-person single-object cases drops to 11.2\%, while multi-person different-HOI cases rise to 67.5\%, which expands the diversity and difficulty of the evaluation.

Moreover, similar to existing HOI datasets, our benchmark remains compatible with HOI-specific methods, enabling evaluation under a unified protocol.
While HOI models output a confidence score for every pre-defined HOI class, practical applications always require a selection step for final prediction, either via a threshold~\cite{zhang2022efficient, lei2023efficient} or by choosing the top-ranked predictions~\cite{Wang_2020_CVPR}, a common step in HOI-specific inference pipelines.
Although this selection is not part of the standard HOI evaluation protocol (e.g., mAP), where all predictions are used directly for evaluation, such a step is routinely applied in real-world HOI applications. 
In our evaluation, we adopt a top-K selection strategy, 
which evaluates HOI-specific models in a practical way.
%
In addition, we also compare against threshold-based filtering and find that the top-5 selection outperforms other choices in Table 5 of the main paper.
Importantly, this conversion only selects the predicted classes before matching them to each question choices, and does not alter model behavior, as it simply aligns the output format for comparison.

Apart from supporting HOI-specific methods, our benchmark is explicitly designed to 
support general-purpose VLMs, unlike HICO-DET, V-COCO, and SWiG-HOI.
This is achieved by framing the task as multiple-choice question answering, naturally aligning with the input–output format of modern VLMs.
Bongard-HOI, although relevant for HOI recognition, is limited to binary classification tasks (i.e., ``is this interaction present or not?''). Our benchmark instead requires multi-class, multi-label prediction, reflecting the true complexity of HOI understanding in realistic images.
Although there are some recent vision-language approaches (e.g., DAM~\cite{Vu_2025_ICCV}, COCONut-PanCap~\cite{deng2025coconutpancap}) related to interaction understanding, they do not provide a unified evaluation protocol that can fairly compare VLMs with HOI-specific detectors.

Taken together, our benchmarking dataset uniquely combines the strengths of previous HOI datasets while addressing their shortcomings. First, our datasets reduces over-simplified single-person cases, emphasizes multi-person cases with different interactions, remains compatible with HOI-specific methods, introduces explicit support for VLM evaluation, and requires full multi-class HOI prediction. This makes it the first benchmark to emphasize challenging cases and enable comparison across both specialized HOI models and general-purpose VLMs.

\noindent \textbf{MCQA  vs. Open-Set HOI Generation}
Our CrossHOI-Bench adopts MCQA to make evaluation well-defined across VLMs and HOI detectors. VLMs often generate free-form text. 
Mapping free-form text to fixed HOI labels is noisy and requires ad hoc rules. MCQA avoids this by fixing the output space, while still testing visual grounding among plausible interaction alternatives. 
We acknowledge that MCQA is easier than fully open-set HOI generation, so we treat it as a diagnostic benchmark, not a replacement for open-set evaluation. 
In this sense, MCQA provides a conservative test of interaction understanding: while strong performance does not necessarily imply full open-set HOI capability, poor MCQA performance is strong evidence that the model struggles with the unconstrained setting.

\subsection{Dataset Construction Details}
We construct our benchmarking dataset using a two-stage approach: coarse screening and manual refinement. 
We refer readers to Sec.~3.1 of the main paper for the full process for coarse screening. The prompts used in the coarse screening stage are provided later in Sec.~\ref{app_sec: imp_detail}.
During the manual refinement stage, we first remove overly simple scenes that offer limited diagnostic value. As illustrated in Fig.~\ref{fig:remove_img}, cases such as multiple people riding bicycles unambiguously or a person washing a toothbrush in a simple background are straightforward. Therefore, these images are excluded from our benchmark.

\input{figures/hard_choice}
For the remaining questions, we then design more challenging choices. 
To create hard positives, we select the target person to perform interactions that differ from those of surrounding individuals (e.g., ``launch boat'' in Fig.~\ref{fig:hard_choice}(a)). 
In ambiguous temporal scenes, we allow multiple plausible actions to be simultaneously correct (e.g., both \emph{boarding} and \emph{exiting} boat in Fig.~\ref{fig:hard_choice}(b)). 
We also include images with people performing multiple actions that models might miss some of these interactions, such as ``load truck'' and ''sit on truck'' in Fig.~\ref{fig:hard_choice}(c).
To create hard negatives, we introduce interactions performed by nearby people that could be mistakenly attributed to the target person (e.g., \emph{hold banana} in
Fig.~\ref{fig:hard_choice}(d)), and we incorporate fine-grained distinctions between visually similar actions (e.g., \emph{repair} vs. \emph{type on} a
laptop in Fig.~\ref{fig:hard_choice}(e)).

In total, the manual refinement stage updated 1,956 out of 5,096 total choices in our main benchmark, accounting for 38.39\% of all choices and indicating a substantial level of human correction. 
Among them, 460 updates were made to positive choices (19.74\% of all 2,330 positives). 
The remaining 1,496 updates correspond to negative choices
(54.10\% of all 2,765 negatives).
Four HOI-focused annotators involve in the manual refinement and final decisions are made by majority vote. 
The annotators follow a shared guideline to independently verify coarse-screened negatives and propose hard positives and hard negatives.
Inter-annotator agreement is high, with 95.22\% mean pairwise agreement and 91.00\% unanimous agreement across all four annotators.  

For the V-COCO-based sub-benchmark, we focus specifically on multi-person scenarios, so we retain only images containing multiple people before applying the same coarse screening and manual refinement. 
During refinement, we add positive choices according to the predefined HOI classes in HICO-DET, since the
original V-COCO benchmark includes only 24 action classes, which is insufficient for creating challenging choices.
For the SWiG-HOI sub-benchmark, we target human–human interaction scenarios, therefore we keep only images annotated with interactions between pairs of humans. We then perform the same coarse screening and manual refinement.

\noindent \textbf{Mitigated Potential Artifacts or Bias}
During our dataset construction, VLMs are only used to generate and coarsely screen candidate negative options. They do not decide the final Multiple-Choice Question Answering (MCQA) options in the benchmark. Every negative option that is kept is checked by humans and must be judged invalid for the image. This addresses the artifact concern that a negative option is actually a valid interaction. 
As for the potential bias, where VLM-based coarse screening could change which types of negatives enter the pool and thus affect task difficulty or model rankings, Table 4 in the main paper provides evidence that the kept negatives are not just ``VLM opinions.'' HOI-specific methods agree with VLM coarse screened negatives at about 99\% agreement, while VLMs are at 95–97\%. This suggests the negatives are generally wrong, not just wrong according to the VLM.

\subsection{Our Dataset Examples}
\label{app_sec: dataset_example}

\begin{figure*}[t]
	\centering
	\includegraphics[width=0.98\linewidth]{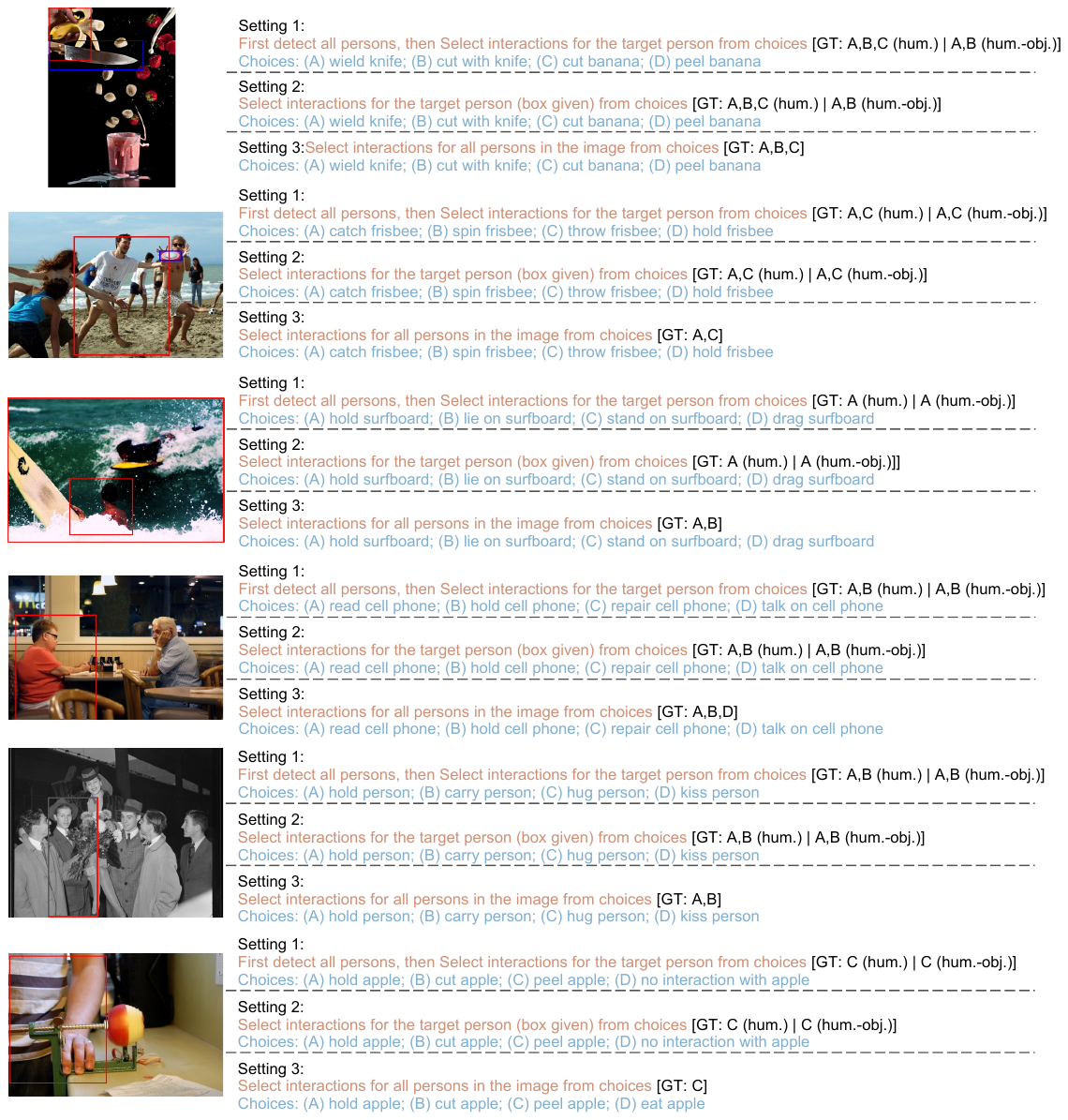}
    \caption{
        Example questions in our benchmark under the three evaluation settings. 
        }	
	\label{fig: appendix_dataset_example}
\end{figure*}

Examples in Fig.~\ref{fig: appendix_dataset_example} illustrate the main challenges our benchmark emphasizes. In multi-person scenarios (e.g., the surfboard, frisbee, and cell phone related examples), different individuals perform distinct interactions, which is potentially confusing and leads to misattributing actions across people. At the same time, certain single-person cases are difficult due to either contactless interactions (e.g., peel apple) or visually similar categories (e.g., hold person vs. hug person). As a result, our benchmark provides a challenging evaluation of HOI understanding.

\subsection{Our HOI Training Dataset}
\label{app_sec: trainset}
We primarily position our benchmark as an evaluation resource, with a central focus on zero-shot VLM performance and unified comparison against HOI-specific methods.
However, for completeness and to support future work that adapts VLMs to HOI tasks, we also provide a standardized training split constructed from the HICO-DET training data. This split includes all three question types required by our benchmark: (i) a \textit{Setting 1} question covering interactions of all annotated people, (ii) a \textit{Setting 2} question focusing on the target person, and (iii) an additional object detection question for \textit{Setting 3}.
This results in 111,459 training questions, offering a protocol for fine-tuning models.

\subsection{Dataset Licenses and Release}
\noindent \textbf{Licenses}  
We use the HICO-DET dataset~\cite{chao2018learning}, which is publicly released under a CC0: Public Domain license.  
Our sub-benchmarks also use V-COCO~\cite{lin2014microsoft} (released under a CC-BY license) and SWiG-HOI~\cite{wang2022learning} (released under its academic research license).

\input{figures/VLM_problems}

\noindent \textbf{Data Release and Ethical Considerations}  
We do not release any images or raw annotations from HICO-DET, V-COCO, or SWiG-HOI.  
Our benchmark only provides derived multiple-choice questions built on top of these datasets.  
Each question references the original image index and is derived from the HOI annotations provided in the respective dataset, supplemented with minor modifications when needed to resolve annotation noise or include additional HOI classes.

No images, bounding boxes, or HOI annotations are redistributed; users must obtain the original datasets separately under their respective licenses.  
Since our release contains only question–answer pairs and index mappings, the risk of exposing personally identifiable information or offensive content is minimal.  
All consent and licensing considerations follow those of the original datasets, and no additional consent verification is conducted beyond their official releases.

\section{Experiments}

\noindent \textbf{Baseline Details}
We evaluate two groups of baselines on our benchmark: general-purpose VLMs and HOI-specific methods.
Recent large VLMs represent the frontier of general-purpose image understanding. 
Qwen2-VL, Qwen2.5-VL (7B / 32B) and Qwen3-VL-instruct~\cite{Qwen_2_5_report} are selected as they excel in fine-grained spatial localization and visual reasoning, making them suitable for HOI tasks. 
InternVL2.5 and InternVL3 (8B / 38B)~\cite{chen2024internvl, wang2024mpo,InternVL3_2025} are included because they achieve leading performance across diverse multimodal benchmarks and emphasize high-resolution perception, which is relevant for recognizing human–object interactions. 
LLaVA-OV-7B~\cite{llavaov_tmlr_2025} is an instruction-tuned VLM designed for open-vocabulary understanding demonstrating versatility and strong performance across multiple vision–language tasks, making it a relevant baseline for HOI evaluation.

For completeness, we additionally report supervised fine-tuning results for Qwen2.5-VL-7B, providing a baseline under a finetuned setup using our HOI training dataset (see Sec.~\ref{app_sec: trainset}). We finetune the VLM using Low-Rank Adaptation (LoRA)~\cite{hu2022lora} applied to the text-decoder attention projection layers (query, key, value) and the final output projection. Training is conducted for 5 epochs with a batch size of 32.

Beyond VLM baselines, we also evaluate recent HOI detection methods. 
ADA-CM~\cite{lei2023efficient}, CMMP~\cite{lei2024exploring}, LAIN~\cite{KimJC_CVPR_2025} and HOLa~\cite{lei2025lhola} demonstrate competitive performance on the existing HICO-DET benchmark~\cite{chao2018learning}. 
In addition, CMD-SE~\cite{Lei_2024_CVPR} is a recent open-vocabulary HOI detection method emphasizing generalization ability, achieving competitive performance on SWiG-HOI~\cite{wang2022learning} and HICO-DET benchmarks. 
We use best-performing pre-trained checkpoints when available, and otherwise reproduce results with the authors’ code under the closest available configurations. 
Specifically, ADA-CM, CMMP, and HOLa are evaluated with the ViT-L vision backbone, while CMD-SE and LAIN are based on ViT-B.

\subsection{Additional Findings}
\label{app_sec: exp}

\paragraph{Additional Finding 1:  Analysis of VLM Failure Cases }
To further understand the performance of VLMs and HOI-specific models, we analyzed 200 failure cases of Qwen2.5-VL-32B~\cite{Qwen_2_5_report} in \textit{Setting 1}, where the HOI model (ADA-CM~\cite{lei2024exploring}) successfully produced the correct prediction but the VLM failed.
Here, we exclude the detection failure cases, which is one of the limitations of VLMs shown in Table 1 (main paper) and instead focus on the remaining failure cases. 
We discuss the detection limitations of VLMs in detail in paragraph \textit{Detection Limitations of VLMs}.
These non-detection failure cases are grouped into four main categories.

\textbf{1. Incomplete multi-action recognition} (54\% out of the 200 failure cases).
The most frequent failure occurs when the VLM predicts only a subset of multiple valid actions for a human–object pair.
For example, in Fig.~\ref{fig:VLM_problem}(a), in an image where a person is riding and hugging a horse, the model recognizes only riding but omits hugging.
This suggests that the VLM tends to capture a certain action while ignoring concurrent interactions.
In contrast, HOI-specific models are less affected by this issue, largely because their training data include co-occurring action annotations.
Through supervised learning on such examples, HOI models implicitly learn that multiple actions can jointly occur on the same human–object pair.

\textbf{2. Cross-person HOI misattribution} (16\% out of the 200 failure cases).
The VLM often misattributes the interaction of a surrounding person to the target person.
When multiple people appear in close proximity performing different actions, the global attention mechanism can mix the features of nearby persons and assign the wrong action to the target.
As shown in Fig.~\ref{fig:VLM_problem}(e), the model incorrectly predicts wearing a tie for the woman who is cutting the tie.
In contrast, HOI-specific models explicitly preserve spatial topology through region cropping or query anchoring, which alleviates such confusion.

\textbf{3. HOI similarity confusion} (13\% out of the 200 failure cases).
Another source of failure arises from visually similar HOI classes. In many cases, two HOIs differ only by subtle local cues that occupy a very small region of the image (e.g., just a few dozen pixels in a 400×600 image). For instance, in Fig.~\ref{fig:VLM_problem}(g), when a person drinks from the cup with their gaze directed toward the camera, the VLM predicts \textit{inspect cup} instead. These errors suggest that current VLMs struggle with fine-grained visual discrimination, which aligns with prior analyses showing VLMs have difficulty distinguishing categories that differ by subtle visual or semantic differences~\cite{xu2024benchmarking, he2025analyzing}.

\textbf{4. Hallucinated HOI inference} (8\% out of the 200 failure cases).
Some failures occur when the image itself does not provide sufficient visual evidence for the action.
For instance, in Fig.~\ref{fig:VLM_problem}(j), when the image shows a person already riding the elephant, rather than mounting it, the VLM predicts \textit{hop on elephant}. 
Despite the absence of visual cues, the VLM occasionally predicts hallucinated actions, likely driven by language priors or common object–action co-occurrence patterns.
This behavior resembles the object hallucination phenomenon reported in previous studies~\cite{datta2025evaluating, chen2024multi}, but here it reflects at the action level, where the model infers plausible yet visually ungrounded interactions.
The HOI-specific model, relying on explicit region-level visual features, is less prone to such hallucinated inferences.

\paragraph{Additional Finding 2: Problems Persist in Both VLMs and HOI-specific Models}
Although the previous failure analysis showed that HOI-specific models outperform VLMs in four categories, we find that for two of them, cross-person HOI misattribution and HOI similarity confusion, HOI-specific models only partially alleviate the problem. In the following, we discuss why HOI models improve these cases compared to VLMs, and why the problems still persist.

Both types of errors originate from how VLMs represent spatial structure and instance-level information. 
VLMs~\cite{radford2021learning, Qwen_2_5_report, InternVL3_2025} encode an image as a flattened sequence of patch embeddings, where global self-attention treats all tokens uniformly. 
Although two-dimensional positional encodings are included, instance-level associations, such as determining which hand belongs to which person, are only learned implicitly rather than structurally enforced. 
As a result, tokens from nearby persons or objects can interfere, causing cross-person misattribution. 
Meanwhile, distinguishing similar HOIs often depends on extremely sparse cues (e.g., hands or gaze direction in Fig.~\ref{fig:VLM_problem}(e)-(f)), which account for only a few patches. 
Because these cues are not explicitly emphasized, the model relies purely on data-driven learning to capture them, which can be unreliable in practice.

HOI-specific methods mitigate these issues by explicitly preserving spatial topology and emphasizing local features. 
Two-stage approaches~\cite{lei2023efficient, lei2024efficient, KimJC_CVPR_2025} extract a two-dimensional feature map after a Transformer encoder and apply RoIAlign to crop human and object regions, isolating instance-specific features and narrowing the search space for fine-grained cues. 
One-stage transformer methods~\cite{tamura2021qpic, ning2023hoiclip, kim_hotr_2021, CMD-SE_2024_CVPR} achieve a similar effect through object queries that serve as implicit region anchors focusing on distinct human–object pairs. 
However, these strategies only partially solve the problem, because cropped regions may still contain unrelated body parts, while query attention can overlap when people are close. Moreover, the critical visual cues for distinguishing actions may lie in extremely small areas, such as a few pixels around the hand or gaze, making these cues too localized to be consistently captured. 

\input{figures/detect_failurecase}

\begin{table*}[ht]
    \setlength{\tabcolsep}{3pt} 
    \renewcommand{\arraystretch}{1} 
    \footnotesize 
\centering
\resizebox{1.2\columnwidth}{!}{
	\begin{tabular}{c c c  c c | c c c c c}
		\hline
	 {Choices}&{A} &{B}& C& D  & {}&{A} &{B}&C& D  \\
		\hline
       Qwen2.5-Vl-32B&0.20 & 0.15 & 0.27 & 0.38 & GT& 0.24 & 0.25 & 0.25 & 0.26 \\ 
       Internvl3-38B&0.27 & 0.25 & 0.26 &0.22 & GT & 0.25 & 0.24 & 0.26 & 0.25\\ 
		\hline
	\end{tabular}
}
\caption{
Analysis of MCQA option bias. We report the distribution of predicted answer positions for two VLM baselines and compare it with the ground-truth distribution. The approximately uniform ground-truth distribution and the mismatch with model predictions indicate that the benchmark does not introduce positional bias.}
\label{tab: mcqa_freq}
\end{table*}

In our analysis of 200 multi-person cases where surrounding individuals perform different actions, the VLM (Qwen2.5-VL-32B~\cite{Qwen_2_5_report}) resulted in 44 misattributed failure cases (22\%), while the HOI model (CMMP~\cite{lei2024exploring}) had 30 (15\%). 
We further analyze 400 questions in the test set focusing on situations where the correct class is easily confused with a semantically related one (e.g., ``cut banana'' vs. ``peel banana'' in Fig.~\ref{fig:VLM_problem}(h)). 
Among these questions, we identify 110 HOI similarity confusions in total. Of these, 65 (59\%) are made by the VLM and 57 (52\%) by the HOI model, with overlap where both models make the same confusion. 
Both experiment results consistently indicate that HOI-specific models alleviate, but do not eliminate errors related to cross-person misattribution and fine-grained similar HOI confusion.

\begin{table*}[ht]
    \setlength{\tabcolsep}{3pt} 
    \renewcommand{\arraystretch}{1} 
    \footnotesize 
\centering
\resizebox{1.2\columnwidth}{!}{
	\begin{tabular}{c| c c | c c }
		\hline
	 {} &{Qwen2.5-VL-32B}&InternVL3-38B& ADA-CM & HOLa \\
		\hline
       Instance-F1& {52.94} $\pm$ 2.12 & 38.68 $\pm$ 2.24  & 47.76 $ \pm$ 2.32 & 47.12 $\pm$ 2.22 \\ 
       Macro-F1 & {50.71}$\pm$ 2.23 & 38.04 $\pm$ 1.91 & 43.02 $\pm$ 2.07 & 43.06$\pm$ 2.15 \\
		\hline
	\end{tabular}
}
\caption{Statistical significance analysis. We report 95\% bootstrap confidence intervals computed using 1,000 question-level resampling iterations in \textit{Setting~1}.}
\label{tab: CI}
\end{table*}

\begin{table*}[t]
\centering
\resizebox{1.9\columnwidth}{!}{
\begin{tabular}{l|c|c|c|c|c|c}
\toprule
{Method}
& Macro-F1 ($\%$)   
& Instance-F1 ($\%$)
& Micro-F1 ($\%$)
& EM ($\%$)
& Avg. Prec. ($\%$)
& Avg. Rec. ($\%$) \\
\midrule
\multicolumn{7}{l}{\textit{HOI-specific methods}} \\
\midrule
ADA-CM~\cite{lei2023efficient}  &  48.89 & \textbf{56.25} & \textbf{67.19} & \textbf{21.74} & \textbf{73.11} & \textbf{62.16} \\
CMMP~\cite{lei2024exploring}   &  \textbf{49.04} & 55.10 & 65.85 & 19.78 & 70.90 & 61.47 \\
\midrule
\multicolumn{7}{l}{\textit{VLM zero-shot evaluation}} \\
\midrule
InternVL2.5-38B~\cite{chen2024internvl} & 48.43 & 46.43 & 51.56 & 20.64 & 77.52 & 38.63 \\
InternVL3-38B~\cite{InternVL3_2025}  &  58.94 & 67.41 & 67.81 & \textbf{35.64} & \textbf{81.90} & 57.85 \\
{Qwen3-VL-30B-Instruct}~\cite{Qwen_2_5_report} & 56.43 & 64.00 & 64.59 & 32.57 & 79.89 & 54.21 \\
Qwen2.5-VL-32B~\cite{Qwen_2_5_report}  & \textbf{62.90} & \textbf{69.52} & \textbf{70.69} & 35.01 & 75.30 & \textbf{66.61} \\
\midrule
LLaVA-OV-7B~\cite{llavaov_tmlr_2025} & 47.76 & 56.53 & 54.80 & 25.12 & \textbf{77.43} & 42.40 \\
InternVL3-8B~\cite{InternVL3_2025}   & \textbf{49.88} & 52.35 & 55.54 & 23.86 & 74.41 & 44.31  \\
Qwen2-VL-7B~\cite{Qwen2-VL} & 46.90 & 53.93 & 53.61 & 23.23 & 76.84 & 41.16  \\
Qwen2.5-VL-7B~\cite{Qwen_2_5_report} & 48.93 & \textbf{57.25} & \textbf{57.53} & \textbf{25.98} & 74.49 & \textbf{46.87} \\
\bottomrule
\end{tabular}
}
\caption{\textit{Setting~2} experiment results comparison. Best performance within each group is highlighted in \textbf{bold}. ``Avg. Prec.'' and ``Avg. Rec.'' denote the precision and recall averaged across the test set, respectively.}
\label{tab:supp_new_setting2_grouped}
\end{table*}

\paragraph{Discussion: Why VLMs Outperform HOI Models in Interaction Understanding}
Although HOI-specific models are designed for HOI detection, based on our experiment results across different settings and benchmarks (main and two complementary), we observe that VLMs often exhibit stronger interaction recognition and contextual understanding abilities. We hypothesize three potential factors behind this advantage.

First, VLMs benefit from vastly broader and more diverse training data~\cite{dong2025scalable}, including internet-scale image–text pairs and instruction-tuned datasets covering a wide spectrum of visual reasoning and commonsense knowledge~\cite{xu2024vision, dai2023instructblip}.
In contrast, HOI datasets such as HICO-DET and V-COCO contain (e.g., HICO-DET train set includes 38,118 images, V-COCO train set includes 5,400 images), which are smaller than datasets VLMs trained on.
Second, VLMs possess far larger model capacities, often scaling to tens of billions of parameters, which enhance their representational power and enable stronger cross-modal generalization~\cite{shen2023scaling, chen2024scaling}. 
Third, the generative nature of VLMs may inherently enhance image understanding and reasoning ability.
Unlike HOI-specific discriminative models that map vision features to fixed categories, the generative modeling principles \edit{}{in} VLMs have been shown to drive scalability, in-context learning, and compositional generalization~\cite{alayrac2022flamingo, openai_gpt4_2023}, which likely contribute to their stronger contextual understanding~\cite{nie2025large}.
However, each factor remains an open question and requires future work to understand their individual contributions.

\paragraph{Detection Limitations of VLMs }
The performance drop from \textit{Setting 2} to \textit{Setting 1} highlights that VLMs are inferior than at detection (e.g., Qwen2.5-VL-32B drops 16.6\% Instance-F1).
When ground-truth boxes are provided in \textit{Setting 2}, VLMs can recognize interactions effectively, but their performance drops significantly in \textit{Setting 1}, where detection must be performed by the VLMs themselves (Table~1, main paper).
HOI-specific methods, in contrast, integrate detection within their pipelines and therefore cannot directly exploit ground-truth boxes, but under \textit{Setting 1} they often perform competitively, sometimes surpassing even large VLMs (e.g., ADA-CM achieves 13.6\% higher Micro-F1 than InternVL3-38B and 0.3\% higher than Qwen2.5-VL-32B).
This indicates that detection remains a major bottleneck for VLMs, while HOI-specific models strike a more balanced trade-off between detection and recognition.

We conduct error analysis using Qwen2.5-VL-32B, which achieves the highest performance in \textit{Setting 1} among VLM baselines. As shown in Fig.~\ref{fig:qwen_det_failurecase}, failures mainly occur in multi-person scenes, where the VLM struggles to correctly localize individuals in crowded settings, and in occluded scenes, where heavy occlusion of the person leads to missed or inaccurate bounding boxes. These cases show that complex layouts and occlusions remain key challenges for VLM detection.

\paragraph{MCQA Option Bias Analysis}
A potential concern with multiple-choice evaluation is that VLMs may exploit option-level shortcuts rather than genuine interaction understanding. To verify that CrossHOI-Bench does not introduce such biases, we analyze the distribution of answer positions and model predictions. As shown in Table~\ref{tab: mcqa_freq}, the correct option position is randomized and approximately uniform. Importantly, the prediction distributions of strong VLM baselines do not align with the ground-truth distribution, indicating that the models are not exploiting positional biases.
We also use a fixed template for all options and construct negatives with the same or closely related objects, so models cannot win by option wording or object-only cues.

\paragraph{Statistical Significance Analysis}
For statistical significance testing, we report 95\% bootstrap confidence intervals in Table~\ref{tab: CI}, using 1,000 question-level resampling iterations in \textit{Setting~1}. This shows our results are statistically stable despite the moderate test size.

\begin{table*}[t]
\centering
\resizebox{1.9\columnwidth}{!}{
\begin{tabular}{l|c|c|c|c|c|c}
\toprule
{Method}
& Macro-F1 ($\%$)   
& Instance-F1 ($\%$)
& Micro-F1 ($\%$)
& EM ($\%$)
& Avg. Prec. ($\%$)
& Avg. Rec. ($\%$) \\
\midrule
\multicolumn{7}{l}{\textit{HOI-specific methods}} \\
\midrule
ADA-CM  &  43.02 & \textbf{47.76} & \textbf{61.69} & 19.15 & 76.25 & 51.80 \\
CMMP   &  43.06 & 46.62 & 60.85 & 18.84 & 75.06 & 51.16 \\
LAIN  &  41.28 & 45.64 & 59.09 & 19.31 & 73.42 & 49.44  \\
HOLa   &  43.61 & 47.12 & 61.29 & 19.78 & 74.31 & \textbf{52.15} \\
CMD-SE  & \textbf{47.49} & 44.66 & 58.71 & \textbf{20.33} & \textbf{78.33} & 46.96 \\
\midrule
\multicolumn{7}{l}{\textit{VLM zero-shot evaluation (off-the-shelf object detector, DETR~\cite{carion2020end})}} \\
\midrule
InternVL2.5-38B & 48.12 & 46.69 & 52.26 & 21.04 & 78.25 & 39.23 \\
InternVL3-38B  & 59.33 & 66.13 & 67.38 & \textbf{33.20} & \textbf{82.20} & 57.08  \\
{Qwen3-VL-30B-Instruct} & 57.05 & 63.55 & 64.91 & 31.95 & 80.32 & 54.46 \\
Qwen2.5-VL-32B  & \textbf{61.59} & \textbf{67.29} & \textbf{69.48} & 31.79 & 74.65 & \textbf{64.98} \\
\midrule
LLaVA-OV-7B  & 47.03 & 55.55 & 54.68 & 24.18 & \textbf{78.43} & 41.97  \\
InternVL3-8B   & \textbf{51.81} & 53.47 & 57.05 & 24.25 & 76.77 & 45.41  \\
Qwen2-VL-7B  & 46.06 & 51.85 & 52.71 & 21.51 & {77.28} & 40.00  \\
Qwen2.5-VL-7B  & 49.56 & \textbf{56.91} & \textbf{57.98} & \textbf{25.35} & 75.22 & \textbf{47.17} \\
\bottomrule
\end{tabular}
}
\caption{\textit{Setting~1} experiment results comparison. VLM zero-shot evaluation is based on off-the-shelf detector (DETR) for object detection~\cite{carion2020end}. Results are grouped by VLM size (small vs.\ large). Best performance within each group is highlighted in \textbf{bold}. ``Avg. Prec.'' means the precision averaged across test set and ``Avg. Rec.'' means the recall averaged across test set.}
\label{tab:setting3_offdet_grouped}
\end{table*}

\subsection{Additional Results}
\paragraph{\textit{Setting 2} Discussion}
\textit{Setting 2} is a diagnostic extension of \textit{Setting 1}. It tests whether VLM underperformance in \textit{Setting 1} is mainly due to localization. We provide ground-truth boxes, so detection failures are removed. This lets us evaluate recognition in isolation. 
Most HOI methods cannot be evaluated in \textit{Setting~2} with GT boxes: one-stage (end-to-end) methods require joint detection, and among two-stage methods, those relying on detector embeddings cannot operate with GT-box input.
However, \textit{Setting~2} is not used to assess paradigm gaps, which are evaluated in \textit{Setting 1}.
For completeness, we include two HOI methods (ADA-CM and CMMP) that can accept GT boxes in Table~\ref{tab:supp_new_setting2_grouped}. They improve over their \textit{Setting 1} results, but remain below the best large VLMs.

\paragraph{VLM Zero-shot Evaluation with Off-the-shelf Object Detector}
Since VLMs often struggle with reliable person detection, we follow the two-stage HOI detection paradigm~\cite{lei2023efficient,wang2024bilateral,zhang2022efficient} and leverage a widely used off-the-shelf object detector, DETR~\cite{carion2020end} pre-trained on HICO-DET. 
Table~\ref{tab:setting3_offdet_grouped} shows that incorporating DETR helps improve performance in the \textit{Setting 1} evaluation, though remains lower than in \textit{Setting~2} due to detection errors. 
Among small VLMs, Qwen2.5-VL-7B achieves the best overall performance among most evaluation metrics, competitive to SOTA HOI-specific methods. For large models, Qwen2.5-VL-32B outperforms InternVL3-38B most of the time and other VLMs included in the comparison. 
By comparing Table~\ref{tab:setting3_offdet_grouped} and Table~1 (main paper), we observe a clear performance drop when VLMs perform detection on their own, as opposed to relying on off-the-shelf detectors (i.e., +6.6\% Macro-F1 for Qwen2.5-VL-7B, +4.34\% for Qwen2.5-VL-32B, +19.92\% for InternVL-8B and +8.35\% for InternVL-38B). This highlights that the detection capability of current VLMs still lags behind that of specialized object detectors.

\begin{table}[t]
\centering
\resizebox{\linewidth}{!}{
\begin{tabular}{l| c c c c}
\toprule
\textbf{Method} &
\textbf{Top-5} &
\textbf{Score $\ge$ 0.3} &
\textbf{Score $\ge$ 0.5} &
\textbf{Score $\ge$ 0.7} \\
\midrule
ADA-CM      & 47.76   & 33.84 & 24.31 & 14.26 \\
CMMP        & 46.62   & 32.84 & 23.33 & 13.74 \\
LAIN        & 45.64  & 30.96 & 21.10 & 10.44 \\
HOLa        & 47.12  & 34.74 & 25.47 & 13.13 \\
CMD-SE      & 44.66   & 39.95 & 35.00 & 29.05 \\
\bottomrule
\end{tabular}
}
\caption{Comparison between Top-$K$ selection and confidence-threshold filtering for HOI-specific models under Setting 1 (Instance-F1).}
\label{tab:hoi_ablation_selection}
\end{table}

\begin{table}[ht]
\centering
\resizebox{0.95\columnwidth}{!}{
	\begin{tabular}{l|c| c c  c c }
		\toprule
	 {Method}&{Top-K}&{Macro-F1} &{Instance-F1}&Micro-F1& EM   \\
		\hline
      ADA-CM&  Top3 &35.17 & 40.86 & 54.77 & 17.58 \\
       {} & Top5 &41.53 & 47.74 & 61.68 & \textbf{19.07}  \\
       {} & Top10 & \textbf{46.30} & \textbf{50.58} & \textbf{63.50} & 15.54 \\ 
		\hline
        CMD-SE & Top3 &  31.35 & 37.67 & 51.19 & 15.86 \\
        {} & Top5  & \textbf{47.49} & 44.66 & 58.71 & \textbf{20.33} \\
        {} & Top10 & 46.35 & \textbf{48.14} & \textbf{61.93} & 17.26 \\
        \bottomrule
	\end{tabular}
}
\caption{Effect of different Top-$K$ values for HOI prediction selection under Setting 1.}
\label{tab: ksweep}
\end{table}

\vspace{-0.2cm}
\begin{table}[ht]
    \setlength{\tabcolsep}{3pt} 
    \renewcommand{\arraystretch}{0.9} 
    \footnotesize 
\centering
\resizebox{1\columnwidth}{!}{
	\begin{tabular}{l| c c  c c }
		\hline
	 {Coarse Screening}&{Macro-F1} &{Instance-F1}&Micro-F1& EM  \\
		\hline
       w/o Qwen2.5-VL-32B& 49.00 & 57.51&65.42&28.00 \\ 
       w/ Qwen2.5-VL-32B & 46.77 & 56.33 & 63.87 & 29.33 \\
		\hline
	\end{tabular}
}
\caption{
Ablation study on the effect of using Qwen2.5-VL-32B in the coarse screening during dataset construction. The experiment is conducted in a 12\% random subset. ``w/o Qwen2.5-VL-32B'' means that Qwen is not used for screening; ``w/ Qwen2.5-VL-32B' means that Qwen is used for screening.
}
\label{tab: ablation_wollm}
\end{table}

\begin{table*}[t]
\centering
\resizebox{1.9\columnwidth}{!}{
\begin{tabular}{l|c|c|c|c|c|c}
\toprule
{Method}
& Macro-F1 ($\%$)   
& Instance-F1 ($\%$)
& Micro-F1 ($\%$)
& EM ($\%$)
& Avg. Prec. ($\%$)
& Avg. Rec. ($\%$) \\
\midrule
\multicolumn{7}{l}{\textit{HOI-specific methods}} \\
\midrule
ADA-CM  & 48.96 & \textbf{42.10} & 50.57 & 23.00 & 77.79 & 46.97 \\
CMMP   & 49.01 & 41.03 & 57.66 & 22.92 & 77.20 & 46.01 \\
LAIN   & 47.13 & 40.77 & 56.65 & 22.21 & 75.53 & 45.33 \\
HOLa   & 48.96 & 41.68 & \textbf{58.16} & \textbf{23.23} & 76.01 & \textbf{47.10}  \\
CMD-SE & \textbf{51.98} & 32.35 & 47.64 & 16.72 & \textbf{78.96} & 34.11 \\
\midrule
\multicolumn{7}{l}{\textit{VLM zero-shot evaluation}} \\
\midrule
InternVL2.5-38B & 7.80 & 4.48 & 6.97 & 2.35 & 77.67 & 3.65 \\
InternVL3-38B  & \textbf{21.35} & \textbf{16.50} & \textbf{24.69} & \textbf{8.79} & \textbf{78.87} & \textbf{14.64} \\
{Qwen3-VL-30B-Instruct} & - & - & -  & - & - & - \\
Qwen2.5-VL-32B  & 17.13 & 11.48 & 19.16 & 5.10 & 76.92 & 10.94 \\
\midrule
LLaVA-OV-7B & - & -& -& -& -& - \\
InternVL3-8B   & 3.07 & 0.55 & 0.45 & 0.08 & \textbf{83.33} & 0.23 \\
Qwen2-VL-7B  & - & -& -& -& -& -  \\
Qwen2.5-VL-7B & \textbf{13.49} & \textbf{10.21} & \textbf{14.88} & \textbf{4.71} & 68.54 & \textbf{8.34}  \\
\midrule
\multicolumn{7}{l}{\textit{VLM zero-shot evaluation (off-the-shelf object detector, DETR~\cite{carion2020end})}} \\
\midrule
InternVL2.5-38B & 46.64 & 34.59 & 42.52 & 16.95 & \textbf{78.34} & 29.18 \\
InternVL3-38B  & 52.72 & 44.47 & 52.77 & \textbf{22.37} & 76.70 & 40.22 \\
{Qwen3-VL-30B-Instruct} &49.17 & 42.55 & 49.95 & 20.17 & 73.80 & 37.76 \\
Qwen2.5-VL-32B  & \textbf{53.03} & \textbf{45.56} & \textbf{54.89} & 21.11 & 71.61 & \textbf{44.51} \\
\midrule
LLaVA-OV-7B & \textbf{45.77} & 38.15 & 42.30 & \textbf{18.29} & \textbf{74.40} & 29.55 \\
InternVL3-8B   & 45.50 & 36.31 & \textbf{44.55} & 15.78 & 71.17 & 32.42   \\
Qwen2-VL-7B  & 43.89 & 37.28 & 41.60 & 17.82 & 72.70 & 29.14 \\
Qwen2.5-VL-7B & 43.56 & \textbf{38.18} & 44.16 & 16.25 & 68.59 & \textbf{32.56}  \\
\bottomrule
\end{tabular}
}
 \caption{Extended \textit{Setting~1} experiment results comparison, requiring detection for both humans and objects, and then predicting HOI classes based on detected human and object boxes. Results are reported for VLMs and HOI-specific methods. Best performance within each group is highlighted in \textbf{bold}. ``Avg. Prec.'' means the precision averaged across test set and ``Avg. Rec.'' means the recall averaged across test set. }
\label{tab:setting1_extend}
\end{table*}

\begin{table*}[t]
\centering
\resizebox{1.9\columnwidth}{!}{
\begin{tabular}{l|c|c|c|c|c|c}
\toprule
{Method}
& Macro-F1 ($\%$)   
& Instance-F1 ($\%$)
& Micro-F1 ($\%$)
& EM ($\%$)
& Avg. Prec. ($\%$)
& Avg. Rec. ($\%$) \\
\midrule
\multicolumn{7}{l}{\textit{VLM zero-shot evaluation}} \\
\midrule
InternVL2.5-38B & 51.09 & 48.74 & 52.95 & 26.22 & 76.49 & 40.49 \\
InternVL3-38B  & 58.02 & 65.20 & 65.92 & \textbf{34.30} & \textbf{76.58} & 57.87 \\
Qwen3-VL-30B-Instruct & 55.00 & 62.45 & 62.66 & 30.46 & 73.79 & 54.45 \\
Qwen2.5-VL-32B  & \textbf{60.00} & \textbf{65.79} & \textbf{66.86} & 31.08 & 70.34 & \textbf{63.70} \\
\midrule
LLaVA-OV-7B & 49.82 & \textbf{56.88} & 55.38 & \textbf{28.49} & \textbf{75.35} & 43.78 \\
InternVL3-8B   &  \textbf{53.27} & 53.62 & \textbf{56.46} & 24.96 & 71.28 & 46.74   \\
Qwen2-VL-7B  & 48.24 & 54.23 & 53.39 & 26.06 & 72.33 & 42.32 \\
Qwen2.5-VL-7B &  48.86 & 55.98 & 56.32 & 26.37 & 70.43 & \textbf{46.92}  \\
\bottomrule
\end{tabular}
}
 \caption{Extended \textit{Setting~2} experiment results comparison, with both human and object boxes provided from each question. Results are reported for VLMs and HOI-specific methods. Best performance within each group is highlighted in \textbf{bold}. ``Avg. Prec.'' means the precision averaged across test set and ``Avg. Rec.'' means the recall averaged across test set.}
\label{tab:setting2_extend}
\end{table*}


\paragraph{Prediction-Selection Strategies for HOI Methods}
Table~\ref{tab:hoi_ablation_selection} compares two common prediction-selection strategies for HOI detectors: selecting the Top-$K$ predictions or applying a confidence threshold. 
We observe that Top-5 selection consistently outperforms threshold-based filtering. 
Applying confidence thresholds leads to substantial drops, as stricter filtering removes many correct predictions and severely hurts recall and F1.
Table~\ref{tab: ksweep} further studies the effect of different $K$ values. 
Increasing $K$ from 3 to 5 significantly improves all F1 metrics, indicating that HOI models often predict multiple valid interactions for a person–object pair. 
However, increasing $K$ to 10 introduces more false positives: recall-related metrics such as Instance-F1 and Micro-F1 continue to increase slightly, while EM begins to decline. 
Overall, Top-5 provides the best balance between recall and precision, and is therefore adopted in our main evaluation.

\paragraph{Potential Circular Bias Analysis}
In the original dataset construction, Qwen2.5-VL-32B was used in the Coarse Screening stage to filter obviously invalid negative options. Since the same model is also evaluated as a baseline, this raises a potential circular bias: the benchmark may favor Qwen models if the negative options were filtered using the same model family during dataset construction.
To examine this effect, we construct a 12\% random subset where Qwen2.5-VL-32B is excluded from the coarse screening step. Qwen2.5-VL-32B still achieves similar Setting-1 performance on this subset (Table~\ref{tab: ablation_wollm}). We also re-evaluate the main baselines on the same subset and observe similar ranking trends. This indicates that Qwen’s performance and the overall comparisons are not driven by a “Qwen-on-Qwen” screening loop.

\paragraph{Evaluation for Target Human-Object Pairs}
Since we require the model to recognize localized interactions for target person by default in \textit{Setting 1} and \textit{Setting 2}, here we provide the experiment results where models are required to identify interactions for target human-object pairs. We name these two settings as \textbf{Extended \textit{Setting 1}} and \textbf{Extended \textit{Setting 2}}.

In Table~\ref{tab:setting1_extend}, each model first detects all human and object instances before predicting HOI classes. The \textit{Extended Setting 1}, therefore, requires comprehensive localization of multiple objects and people in diverse scenes. The performance drop observed across all VLMs highlights their limited ability to perform reliable object detection. Even though the object categories are predefined (the 80 object classes from HICO-DET), large VLMs still struggle to consistently identify all relevant objects. In contrast, HOI-specific methods, which are trained jointly for detection and interaction recognition, achieve improved performance in this setting. These results indicate that while large VLMs exhibit promising zero-shot learning ability, their object detection remains a critical bottleneck for HOI detection.

We further leverage an off-the-shelf DETR model for object detection, which is also commonly adopted by HOI-specific methods~\cite{lei2023efficient,zhang2021spatially, zhang2022efficient}. With this detector, VLMs perform substantially better than when detecting objects by themselves, as shown in Table~\ref{tab:setting1_extend}. However, smaller VLMs still lag behind HOI-specific methods, and even large models such as InternVL3-38B and Qwen2.5-VL-32B only slightly outperform the best HOI-specific method in a few metrics (e.g., Instance-F1, Macro-F1) while falling behind in others.

Extended \textit{Setting 2} isolates the detection ability and focuses on localized human-object pair interaction recognition. However, compared with default \textit{Setting 2}, the overall performance of VLMs drops slightly when the target object is specified, as shown in Table~\ref{tab:setting2_extend}.
This is because VLMs tend to predict interactions involving non-target objects. For example, when a person is sitting on a bench while holding a book, and the target object is bench, the model often predicts both ``sit on bench'' and ``hold a book''. Quantitatively, such wrong-object predictions account for 11.60\% of all errors in InternVL3-38B and 9.51\% in Qwen2.5-VL-32B. This suggests that even with explicit object localization cues, current VLMs struggle to restrict their predictions to the intended target object.

\paragraph{Sub-benchmarks Evaluation}
The sub-benchmark experiments extend our main evaluation by evaluating model performance on the V-COCO-based and SWiG-HOI-based subsets. These subsets emphasize distinct challenges: V-COCO-based sub-benchmark focuses on multi-person scenes, whereas SWiG-HOI-based one highlights human-human interaction understanding.
Table~\ref{tab:setting1_cross_dataset}-\ref{tab:setting3_cross_dataset} display sub-benchmark evaluations across baselines.

For HOI-specific models, sub-benchmarks serve as the out-of-distribution evaluation, as we require one model to be tested across our three benchmarks (one main benchmark based on HICO-DET and two sub-benchmarks based on V-COCO and SWiG-HOI). All HOI-specific methods evaluated on sub-benchmarks are pre-trained on HICO-DET training dataset.
Tables~\ref{tab:setting1_cross_dataset}-\ref{tab:setting3_cross_dataset} show that HOI-specific methods experience clear performance drop when evaluated on the V-COCO– and SWiG-HOI–based subsets. 
The V-COCO-based sub-benchmark has a larger label space after we add additional HOI annotations based on HICO-DET pre-defined classes (Original V-COCO only include 24 action labels). 
For example, in \textit{Setting 1}, CMMP decreases from 46.62 Instance-F1 on the HICO-based main benchmark (Table 1 of main paper) to 28.39 on the V-COCO-based sub-benchmark, an absolute drop of 18.23 points. In contrast, CMD-SE decreases from 44.66 Instance-F1 (main benchmark) to 7.37 on the SWiG-HOI sub-benchmark, a much larger drop of 37.29 points. This pattern is consistent across other settings. These numbers indicate that HOI-specific models struggle with cross-dataset generalization, suffering much larger performance drops on SWiG-HOI-based sub-benchmarks, whose HOI class distributions and interaction patterns differ more from those in HICO-DET-based main benchmark.

VLMs, under zero-shot evaluation, demonstrate consistent generalization across datasets, but they nevertheless experience the largest performance drop on the V-COCO-based sub-benchmark, posing the greatest challenge among the three benchmarks.
Detection-required settings amplify this difficulty.
In \textit{Setting 1}, Qwen2.5-VL-7B drops from 30.53 Instance-F1 on the main benchmark (Table 1, main paper) to 12.66 on V-COCO (-17.87), which is substantially lower than its performance on SWiG-HOI (47.95). This pattern becomes more noticeable in Extended \textit{Setting 1}, where the requirement to detect both humans and objects causes even large VLMs drop drastically. InternVL3-38B falls from 16.50 Instance F1 on main benchmark to 10.07 on V-COCO (Table~\ref{tab:setting1_extend_cross_dataset}). These results highlight that scenarios on V-COCO with multiple people and multiple objects, significantly increase the difficulty of reliable person/object detection for VLMs.

When detection is not required, most VLMs show a clear Macro-F1 drop when moving from the main benchmark to either sub-benchmark. For example, Qwen2.5-VL-32B decreases from 62.90 Macro-F1 on the main benchmark (Table 2, main paper) to 55.61 on V-COCO and 52.74 on SWiG-HOI (Table~\ref{tab:setting2_cross_dataset}). A similar pattern appears in Extended \textit{Setting 2}, where its Macro-F1 falls from 60.00 (Table~\ref{tab:setting2_extend}) to 44.76 on V-COCO and 55.48 on SWiG-HOI (Table~\ref{tab:setting2_extend_cross_dataset}).
These consistent Macro-F1 reductions across models indicate that both sub-benchmarks exhibit a distribution shift in HOI classes, under multi-person and human-human interaction scenes.
In \textit{Setting 3}, the EM accuracy drops a lot across VLMs, when comparing main benchmark and V-COCO sub-benchmarks.
This indicates that it is challenging to recognize comprehensive interactions across the image under multiple people and multiple object scenarios.

\begin{table*}[t]
\centering
\resizebox{1.9\columnwidth}{!}{
\begin{tabular}{l|c|c|c|c|c|c|c}
\toprule
{Method}
&
{Dataset}
& Macro-F1 ($\%$)   
& Instance-F1 ($\%$)
& Micro-F1 ($\%$)
& EM ($\%$)
& Avg. Prec. ($\%$)
& Avg. Rec. ($\%$) \\
\midrule
\multicolumn{8}{l}{\textit{HOI-specific methods } \; Out-of-distribution evaluation} \\
\midrule
ADA-CM  & -&-&-&-&-&-&- \\
CMMP   & V-COCO & 25.57 & 28.39 & 37.07 & 2.81 & 80.56 & 24.07 \\
LAIN   & V-COCO & 11.62 & 17.04 & 23.99 & 3.01 & 68.26 & 14.55 \\
HOLa   & -&-&-&-&-&-&- \\
CMD-SE & SWiG-HOI & 9.09 & 7.37 & 10.81 & 0.00 & 100.0 & 5.71 \\
\midrule
\multicolumn{8}{l}{\textit{VLM } \; Zero-shot evaluation: V-COCO-based sub-benchmark} \\
\midrule
InternVL2.5-38B & V-COCO & 22.73 & 27.07 & 34.41 & 5.01 & 84.35 & 21.62 \\
InternVL3-38B  & V-COCO & 31.56 & 32.98 & 42.49 & 11.02 & \textbf{86.11} & 28.20 \\
Qwen3-VL-30B-instruct & V-COCO & - & - & - & - & - & - \\
Qwen2.5-VL-32B  & V-COCO & \textbf{43.97 }& \textbf{52.73} & \textbf{60.93} & \textbf{21.04} & 80.59 & \textbf{48.98} \\
\midrule
LLaVA-OV-7B & V-COCO& - & -& -& -& -& - \\
InternVL3-8B   & V-COCO & 5.51 & 4.39 & 6.76 & 0.80 & 77.63 & 3.53 \\
Qwen2-VL-7B  & V-COCO& - & -& -& -& -& - \\
Qwen2.5-VL-7B & V-COCO & \textbf{11.88} & \textbf{12.66} & \textbf{17.43} & \textbf{0.80} & \textbf{81.50} & \textbf{9.76} \\
\midrule
\multicolumn{8}{l}{\textit{VLM } \; Zero-shot evaluation: SWiG-HOI-based sub-benchmark} \\
\midrule
InternVL2.5-38B & SWiG-HOI & 13.69 & 14.65 & 20.91 & 8.86 & 77.93 & 12.07 \\
InternVL3-38B  & SWiG-HOI & 28.11 & 30.25 & 38.47 & 18.31 & \textbf{81.70} & 25.16 \\
Qwen3-VL-30B-instruct & SWiG-HOI &1.97 & 1.35 & 2.19 & 0.76 & 76.60 & 1.11 \\
Qwen2.5-VL-32B  & SWiG-HOI & \textbf{45.62} & \textbf{52.66} & \textbf{59.92} & \textbf{25.66} & {68.31} & \textbf{53.37} \\
\midrule
LLaVA-OV-7B & SWiG-HOI & - & -& -& -& -& - \\
InternVL3-8B  & SWiG-HOI & 5.33 & 4.55 & 7.32 & 2.38 & 74.85 & 3.85 \\
Qwen2-VL-7B & SWiG-HOI & - & -& -& -& -& - \\
Qwen2.5-VL-7B & SWiG-HOI & \textbf{39.36} & \textbf{47.95} & \textbf{54.08} & \textbf{24.53} & \textbf{70.91} & \textbf{43.70} \\
\midrule
\multicolumn{8}{l}{\textit{VLM } \; Zero-shot evaluation: Combined (main + two sub-) benchmarks } \\
\midrule
InternVL2.5-38B & Combined & 20.67 & 18.58 & 26.82 & 8.44 & 82.34 & 16.02 \\
InternVL3-38B  &Combined & 34.93 & 33.56 & 42.61 & 18.02 & \textbf{83.81} & 28.56 \\
Qwen3-VL-30B-instruct &Combined & - & - & - & - & - & -  \\
Qwen2.5-VL-32B  & Combined &  \textbf{48.65 }& \textbf{52.77} & \textbf{60.61} & \textbf{25.17} & 72.82 & \textbf{51.91} \\
\midrule
LLaVA-OV-7B &Combined & - & -& -& -& -& - \\
InternVL3-8B   & Combined & 6.42 & 4.66 & 7.56 & 2.04 & \textbf{75.85} & 3.99 \\
Qwen2-VL-7B &Combined & - & -& -& -& -& - \\
Qwen2.5-VL-7B &  Combined & \textbf{29.67} & \textbf{36.02} & \textbf{42.34} & \textbf{17.66} & 72.92 & \textbf{29.83}  \\
\bottomrule
\end{tabular}
}
 \caption{\textit{Setting~1} experiment results comparison under two sub-benchmarks. Results are reported for VLMs and HOI-specific methods. Best performance within each group is highlighted in \textbf{bold}. ``Avg. Prec.'' means the precision averaged across test set and ``Avg. Rec.'' means the recall averaged across test set.
 }
\label{tab:setting1_cross_dataset}
\end{table*}

\begin{table*}[t]
\centering
\resizebox{1.9\columnwidth}{!}{
\begin{tabular}{l|c|c|c|c|c|c|c}
\toprule
{Method}
&
{Dataset}
& Macro-F1 ($\%$)   
& Instance-F1 ($\%$)
& Micro-F1 ($\%$)
& EM ($\%$)
& Avg. Prec. ($\%$)
& Avg. Rec. ($\%$) \\
\midrule
\multicolumn{8}{l}{\textit{HOI-specific methods } \; Out-of-distribution evaluation} \\
\midrule
ADA-CM  & -&-&-&-&-&-&- \\
CMMP   & V-COCO & 24.93 & 22.61 & 34.84 & 4.01 & 80.76 & 22.21 \\
LAIN   & V-COCO & 12.42 & 14.28 & 23.61 & 3.21 & 66.54 & 14.35 \\
HOLa   & -&-&-&-&-&-&- \\
CMD-SE & SWiG-HOI & 16..67 & 5.26 & 5.56 & 0.05 & 100.0 & 2.86 \\
\midrule
\multicolumn{8}{l}{\textit{VLM } \; Zero-shot evaluation: V-COCO-based sub-benchmark} \\
\midrule
InternVL2.5-38B & V-COCO & 5.83 & 5.88 & 6.58 & 2.00 & 71.67 & 3.45 \\
InternVL3-38B  & V-COCO & \textbf{10.87} & \textbf{10.07} & \textbf{14.57} & \textbf{4.01} & \textbf{72.66} & \textbf{8.10} \\
Qwen3-VL-30B-instruct & V-COCO & - & - & - & - & - & - \\
Qwen2.5-VL-32B  & V-COCO & 1.30 & 3.93 & 3.68 & 0.80 & 41.38 & 1.92 \\
\midrule
LLaVA-OV-7B & V-COCO & - & -& -& -& -& - \\
InternVL3-8B  & V-COCO & - & -& -& -& -& -  \\
Qwen2-VL-7B  & V-COCO& - & -& -& -& -& - \\
Qwen2.5-VL-7B & V-COCO & \textbf{3.05} & \textbf{2.87} & \textbf{2.19} & \textbf{0.40} & \textbf{48.28} & \textbf{1.12} \\
\midrule
\multicolumn{8}{l}{\textit{VLM } \; Zero-shot evaluation: SWiG-HOI-based sub-benchmark} \\
\midrule
InternVL2.5-38B &SWiG-HOI & 10.39 & 7.82 & 12.10 & 4.75 & 78.02 & 6.56 \\
InternVL3-38B  &  SWiG-HOI & 20.12 & 22.75 & 31.69 & 13.61 & 78.28 & 19.86 \\
Qwen3-VL-30B-instruct &SWiG-HOI & - & - & - & - & - & - \\
Qwen2.5-VL-32B  &SWiG-HOI & \textbf{38.48} & \textbf{42.72} & \textbf{52.81} & \textbf{21.02} & \textbf{66.92} & \textbf{43.61} \\
\midrule
LLaVA-OV-7B &SWiG-HOI & - & -& -& -& -& - \\
InternVL3-8B   & SWiG-HOI & 2.43 & 2.43 & 4.07 & 1.24 & \textbf{73.12} & 2.09 \\
Qwen2-VL-7B  &SWiG-HOI & - & -& -& -& -& - \\
Qwen2.5-VL-7B & SWiG-HOI & \textbf{32.43} & \textbf{36.61} & \textbf{45.62} & \textbf{18.04} & 70.63 & \textbf{33.69} \\
\midrule
\multicolumn{8}{l}{\textit{VLM } \; Zero-shot evaluation: Combined (main + two sub-) benchmarks } \\
\midrule
InternVL2.5-38B & Combined & 6.14 & 6.36 & 9.43 & 3.53 & 77.06 & 5.02 \\
InternVL3-38B  & Combined & 18.70 & 18.46 & 26.49 & 10.60 & \textbf{77.88} & 15.96 \\
Qwen3-VL-30B-instruct & Combined & - & - & - & - & - & - \\
Qwen2.5-VL-32B  & Combined & \textbf{19.78} & \textbf{25.52} & \textbf{36.63} & \textbf{12.64} & 67.58 & \textbf{25.12 } \\
\midrule
LLaVA-OV-7B &  Combined & - & - & - & - & - & - \\
InternVL3-8B   & Combined & - & - & - & - & - & -   \\
Qwen2-VL-7B  &  Combined & - & - & - & - & - & -  \\
Qwen2.5-VL-7B &  Combined &  \textbf{16.17} & \textbf{21.91} & \textbf{30.26} & \textbf{10.93 }& \textbf{69.97} & \textbf{19.31} \\
\bottomrule
\end{tabular}
}
 \caption{Extended \textit{Setting~1} experiment results comparison, derived from the V-COCO and SWiG-HOI images with 2499 questions in total. Results are reported for VLMs and HOI-specific methods. Best performance within each group is highlighted in \textbf{bold}. ``Avg. Prec.'' means the precision averaged across test set and ``Avg. Rec.'' means the recall averaged across test set.}
\label{tab:setting1_extend_cross_dataset}
\end{table*}

\begin{table*}[t]
\centering
\resizebox{1.9\columnwidth}{!}{
\begin{tabular}{l|c|c|c|c|c|c|c}
\toprule
{Method}
&
{Dataset}
& Macro-F1 ($\%$)   
& Instance-F1 ($\%$)
& Micro-F1 ($\%$)
& EM ($\%$)
& Avg. Prec. ($\%$)
& Avg. Rec. ($\%$) \\
\midrule
\multicolumn{8}{l}{\textit{VLM } \; Zero-shot evaluation: V-COCO-based sub-benchmark} \\
\midrule
InternVL2.5-38B & V-COCO & 38.46 & 48.43 & 52.39 & 8.42 & 83.91 & 38.08 \\
InternVL3-38B  & V-COCO & 48.48 & 62.19 & 64.03 & 17.23 & \textbf{85.10 }& 51.32 \\
Qwen3-VL-30B-instruct & V-COCO& 41.36 & 55.99 & 57.48 & 8.42 & 84.36 & 43.59 \\
Qwen2.5-VL-32B  & V-COCO & \textbf{55.61} & \textbf{71.18} & \textbf{72.85} & \textbf{28.86} & 82.34 & \textbf{65.33} \\
\midrule
LLaVA-OV-7B & V-COCO & 30.82 & 35.75 & 37.79 & 0.60 & 82.00 & 24.55 \\
InternVL3-8B   & V-COCO & 37.78 & 48.73 & 51.12 & 5.41 & 83.24 & 36.89 \\
Qwen2-VL-7B  & V-COCO & 34.49 & 48.76 & 48.07 & 1.80 & \textbf{85.21} & 33.47 \\
Qwen2.5-VL-7B & V-COCO & \textbf{42.65} & \textbf{54.76} & \textbf{55.56} & \textbf{6.21} & 82.35 & \textbf{41.92} \\
\midrule
\multicolumn{8}{l}{\textit{VLM } \; Zero-shot evaluation:SWiG-HOI-based sub-benchmark} \\
\midrule
InternVL2.5-38B &SWiG-HOI & 42.64 & 51.45 & 55.39 & 31.06 & 81.51 & 41.95 \\
InternVL3-38B & SWiG-HOI & 51.78 & {69.73} & 67.96 & {43.11} & \textbf{82.30} & 57.87 \\
Qwen3-VL-30B-instruct & SWiG-HOI & \textbf{57.16} & \textbf{74.87} & \textbf{73.62} & \textbf{45.60} & 82.22 & 66.65 \\
Qwen2.5-VL-32B  & SWiG-HOI & {52.74} & 69.30 & {69.69} & 32.52 & 68.90 & \textbf{70.50} \\
\midrule
LLaVA-OV-7B & SWiG-HOI & 43.23 & 62.26 & 58.61 & 37.93 & 80.71 & 46.01 \\
InternVL3-8B  & SWiG-HOI & 44.83 & 52.76 & 55.96 & 31.98 & 78.06 & 43.61 \\
Qwen2-VL-7B & SWiG-HOI & 46.34 & 62.80 & 60.84 & \textbf{38.36} & \textbf{84.63} & 47.49 \\
Qwen2.5-VL-7B & SWiG-HOI & \textbf{50.38} & \textbf{67.08} & \textbf{66.36} & 34.52 & 72.13 & \textbf{61.44} \\
\midrule
\multicolumn{8}{l}{\textit{VLM } \; Zero-shot evaluation: Combined (main + two sub-) benchmarks } \\
\midrule
InternVL2.5-38B & Combined & 42.52 & 49.24 & 53.48 & 24.28 & 80.72 & 39.99 \\
InternVL3-38B  & Combined & 55.53 & 67.66 & 67.04 & \textbf{36.92} & \textbf{82.74} & 56.35 \\
Qwen3-VL-30B-instruct & Combined & 54.22 & 67.84 & 67.34 & {35.46 }& 81.79 & 57.22 \\
Qwen2.5-VL-32B  & Combined & \textbf{59.10} & \textbf{69.70} & \textbf{70.68} & 32.89 & 73.52 & \textbf{68.06} \\
\midrule
LLaVA-OV-7B &  Combined & 44.76 & 55.78 & 53.19 & \textbf{28.28} & 79.74 & 39.91 \\
InternVL3-8B   & Combined & 45.83 & 52.73 & 55.43 & 26.05 & 78.33 & 42.89 \\
Qwen2-VL-7B  &  Combined & 44.56 & 57.40 & 55.77 & 28.01 & \textbf{82.13 }& 42.22 \\
Qwen2.5-VL-7B &  Combined &\textbf{ 48.27} &\textbf{ 61.65} & \textbf{61.43} & 27.62 & 74.52 & \textbf{52.26 } \\
\bottomrule
\end{tabular}
}
 \caption{\textit{Setting~2} experiment results comparison, derived from the V-COCO and SWiG-HOI images with 2499 questions in total. Results are reported for VLMs and HOI-specific methods. Best performance within each group is highlighted in \textbf{bold}. ``Avg. Prec.'' means the precision averaged across test set and ``Avg. Rec.'' means the recall averaged across test set.}
\label{tab:setting2_cross_dataset}
\end{table*}

\begin{table*}[t]
\centering
\resizebox{1.9\columnwidth}{!}{
\begin{tabular}{l|c|c|c|c|c|c|c}
\toprule
{Method}
&
{Dataset}
& Macro-F1 ($\%$)   
& Instance-F1 ($\%$)
& Micro-F1 ($\%$)
& EM ($\%$)
& Avg. Prec. ($\%$)
& Avg. Rec. ($\%$) \\
\midrule
\multicolumn{8}{l}{\textit{VLM } \; Zero-shot evaluation: V-COCO-based sub-benchmark} \\
\midrule
InternVL2.5-38B & V-COCO & 40.72 & 45.36 & 48.77 & 20.84 & \textbf{73.51} & 36.49 \\
InternVL3-38B  & V-COCO & \textbf{47.22} & 60.55 & 62.45 & \textbf{29.66} & 72.21 & 55.01 \\
Qwen3-VL-30B-instruct &  V-COCO &39.72 & 53.33 & 55.65 & 23.25 & 67.39 & 47.39 \\
Qwen2.5-VL-32B  & V-COCO & 44.76 & \textbf{63.14} & \textbf{64.80} & 28.26 & 67.78 & \textbf{62.07} \\
\midrule
LLaVA-OV-7B & V-COCO & 26.10 & 34.17 & 37.64 & 12.02 & 65.67 & 26.38 \\
InternVL3-8B   & V-COCO & \textbf{39.20} & \textbf{49.51} & \textbf{52.55} & \textbf{24.05} & 66.63 & \textbf{43.38} \\
Qwen2-VL-7B  & V-COCO & 31.32 & 45.55 & 46.60 & 16.03 & \textbf{68.00} & 35.45 \\
Qwen2.5-VL-7B & V-COCO & {38.00} & 48.97 & 50.61 & 18.04 & 64.75 & 41.54\\
\midrule
\multicolumn{8}{l}{\textit{VLM } \; Zero-shot evaluation: SWiG-HOI-sub-benchmark} \\
\midrule
InternVL2.5-38B & SWiG-HOI & 42.84 & 53.97 & 57.19 & 31.98 & 79.42 & 44.69 \\
InternVL3-38B & SWiG-HOI & 48.94 & 68.83 & 67.80 & {40.90} & \textbf{80.46} & 58.58 \\
Qwen3-VL-30B-instruct & SWiG-HOI & \textbf{58.51} & \textbf{75.70} & \textbf{74.53} & \textbf{46.03} & 81.42 & 68.71 \\
Qwen2.5-VL-32B & SWiG-HOI & {55.48} & {69.37} & {70.14} & {32.47} & 68.49 & \textbf{71.88} \\
\midrule
LLaVA-OV-7B & SWiG-HOI & 44.22 & 62.77 & 59.03 & 38.36 & 81.34 & 46.32 \\
InternVL3-8B & SWiG-HOI & 45.22 & 52.56 & 56.78 & 30.96 & 78.52 & 44.47 \\
Qwen2-VL-7B & SWiG-HOI & 47.93 & \textbf{67.50} & 63.29 & \textbf{42.57} & \textbf{84.99} & 50.42 \\
Qwen2.5-VL-7B & SWiG-HOI & \textbf{50.44} & 66.67 & \textbf{65.75} & 32.52 & 71.12 & \textbf{61.13} \\
\midrule
\multicolumn{8}{l}{\textit{VLM } \; Zero-shot evaluation: Combined (main + two sub-) benchmarks } \\
\midrule
InternVL2.5-38B & Combined & 42.60 & 50.73 & 54.28 & 28.42 & 77.46 & 41.78 \\
InternVL3-38B  & Combined & 53.23 & 66.19 & 66.17 & 37.03 & \textbf{77.59 }& 57.68 \\
Qwen3-VL-30B-instruct & Combined & 54.21 & \textbf{67.14} & 67.27 & \textbf{37.53} & 76.86 & 59.80 \\
Qwen2.5-VL-32B  & Combined & \textbf{56.75} & 67.09 & \textbf{68.14} & 31.40 & 68.93 & \textbf{67.37 }\\
\midrule
LLaVA-OV-7B &  Combined & 45.18 & 55.87 & 54.18 & 31.26 & 77.07 & 41.77 \\
InternVL3-8B   & Combined & 46.38 & 52.39 & 55.86 & 27.90 & 73.61 & 45.01 \\
Qwen2-VL-7B  &  Combined & 45.49 & 59.26 & 57.03 & \textbf{33.11} & \textbf{77.92} & 44.97 \\
Qwen2.5-VL-7B &  Combined &  \textbf{47.36} & \textbf{60.03} & \textbf{60.18} & 28.37 & 69.91 & \textbf{52.82} \\
\bottomrule
\end{tabular}
}
 \caption{Extended \textit{Setting~2} experiment results comparison, derived from the V-COCO and SWiG-HOI images with 2499 questions in total. Results are reported for VLMs and HOI-specific methods. Best performance within each group is highlighted in \textbf{bold}. ``Avg. Prec.'' means the precision averaged across test set and ``Avg. Rec.'' means the recall averaged across test set.}
\label{tab:setting2_extend_cross_dataset}
\end{table*}

\begin{table*}[t]
\centering
\resizebox{1.9\columnwidth}{!}{
\begin{tabular}{l|c|c|c|c|c|c|c}
\toprule
{Method}
&
{Dataset}
& Macro-F1 ($\%$)   
& Instance-F1 ($\%$)
& Micro-F1 ($\%$)
& EM ($\%$)
& Avg. Prec. ($\%$)
& Avg. Rec. ($\%$) \\
\midrule
\multicolumn{8}{l}{\textit{HOI-specific methods } \; Out-of-distribution evaluation} \\
\midrule
ADA-CM  & -&-&-&-&-&-&- \\
\hline
CMMP   & V-COCO & 29.68 & 46.50 & 52.58 & 2.81 & 77.46 & 39.80 \\
\hline
LAIN   & V-COCO & 15.35 & 28.49 & 35.11 & 3.81 & 65.49 & 23.99 \\
\hline
HOLa   & -&-&-&-&-&-&- \\
\hline
CMD-SE & SWiG-HOI & 6.67 & 7.37 & 10.81 & 0.00 & 100.0 & 5.71 \\
\midrule
\multicolumn{8}{l}{\textit{VLM } \; Zero-shot evaluation: V-COCO-based sub-benchmark} \\
\midrule
InternVL2.5-38B & V-COCO & 38.58 & 50.76 & 55.05 & 8.02 & {78.31} & 42.45 \\
InternVL3-38B  & V-COCO & 45.43 & 59.07 & 61.49 & 11.62 & 78.03 & 50.73 \\
Qwen3-VL-30B-instruct & V-COCO & 45.31 & 60.57 & 61.66 & 11.22 & \textbf{81.75} & 49.50 \\
Qwen2.5-VL-32B  & V-COCO & \textbf{47.17} & \textbf{66.83} & \textbf{68.81} & \textbf{16.83} & 75.02 & \textbf{63.55}\\
\midrule
LLaVA-OV-7B  & V-COCO & 29.67 & 36.55 & 38.26 & 0.60 & \textbf{84.20} & 24.75 \\
InternVL3-8B   & V-COCO & \textbf{45.43} & \textbf{59.07} & \textbf{61.49} & \textbf{11.62} & 78.03 & \textbf{50.73} \\
Qwen2-VL-7B  & V-COCO & 38.27 & 51.66 & 52.11 & 5.01 & 84.14 & 37.74 \\
Qwen2.5-VL-7B  & V-COCO & 43.34 & 55.59 & 57.38 & 8.62 & 79.79 & 44.80 \\
\midrule
\multicolumn{8}{l}{\textit{VLM } \; Zero-shot evaluation: SWiG-HOI-based sub-benchmark} \\
\midrule
InternVL2.5-38B & SWiG-HOI & 45.82 & 59.27 & 61.26 & 34.41 & 79.82 & 49.70 \\
InternVL3-38B & SWiG-HOI & 49.15 & 69.22 & 67.49 & \textbf{41.11} & \textbf{82.50} & 57.11 \\
Qwen3-VL-30B-instruct & SWiG-HOI & \textbf{56.94} & \textbf{75.08} & \textbf{73.91} & {40.73} & 78.05 & 70.19 \\
Qwen2.5-VL-32B & SWiG-HOI & {53.05} &{ 69.60 }& {70.42} & 32.36 & 68.93 & \textbf{71.99}\\
\midrule
LLaVA-OV-7B & SWiG-HOI & 43.11 & 64.12 & 59.66 & 38.68 & \textbf{83.61} & 46.37 \\
InternVL3-8B & SWiG-HOI & \textbf{49.15} & \textbf{69.22} & \textbf{67.49} &\textbf{41.11} & 82.50 & \textbf{57.11} \\
Qwen2-VL-7B  & SWiG-HOI & 38.40 & 44.80 & 52.93 & 22.42 & 80.21 & 39.50 \\
Qwen2.5-VL-7B  & SWiG-HOI & 48.97 & 67.34 & 65.94 & 34.47 & 72.90 & 60.20 \\
\midrule
\multicolumn{8}{l}{\textit{VLM } \; Zero-shot evaluation: Combined (main + two sub-) benchmarks } \\
\midrule
InternVL2.5-38B & Combined & 47.41 & 55.29 & 57.86 & 24.75 & 80.99 & 45.01 \\
InternVL3-38B  & Combined & 54.77 & 65.44 & 64.72 & 30.57 & \textbf{83.01 }& 53.04 \\
Qwen3-VL-30B-instruct & Combined & 57.01 & \textbf{69.87} & 69.18 & \textbf{30.71} & 81.29 & 60.21 \\
Qwen2.5-VL-32B  & Combined & \textbf{59.15} & 68.31 & \textbf{69.54} & 26.99 & 72.97 & \textbf{66.41} \\
\midrule
LLaVA-OV-7B &  Combined & 43.79 & 55.99 & 52.88 & 24.39 & \textbf{84.28 }& 38.53\\
InternVL3-8B   & Combined & \textbf{52.17} & \textbf{63.37} & \textbf{62.78} & \textbf{28.42} & 81.77 & \textbf{50.95 }\\
Qwen2-VL-7B  &  Combined & 34.94 & 43.31 & 48.89 & 14.65 & 81.83 & 34.86\\
Qwen2.5-VL-7B &  Combined &  50.16 & 61.50 & 60.87 & 24.01 & 76.29 & 50.63 \\
\bottomrule
\end{tabular}
}
 \caption{\textit{Setting~3} experiment results comparison, derived from the V-COCO and SWiG-HOI images with 2499 questions in total. Results are reported for VLMs and HOI-specific methods.  Best performance within each group is highlighted in \textbf{bold}. ``Avg. Prec.'' means the precision averaged across test set and ``Avg. Rec.'' means the recall averaged across test set.}
\label{tab:setting3_cross_dataset}
\end{table*}

\begin{table*}[t]
\centering
\resizebox{1.98\columnwidth}{!}{
\begin{tabular}{l|c|c|c|c|c|c|c}
\toprule
{Method}
& Setting
& Macro-F1 ($\%$)   
& Instance-F1 ($\%$)
& Micro-F1 ($\%$)
& EM ($\%$)
& Avg. Prec. ($\%$)
& Avg. Rec. ($\%$) \\
\midrule
\midrule
\multicolumn{7}{l}{\textit{Main Benchmark}} \\
\midrule
 SFT Baseline  & 1   & 61.12 & 61.70 & 69.63 & 26.22 & 67.69 & 71.67 \\
  SFT Baseline  & Extended 1  & 41.86 & 34.90 & 53.17 & 14.21 & 66.04 & 44.51 \\
 SFT Baseline  & 2  & 63.64 & 68.93 & 72.57 & 28.89 & 66.88 & 79.31 \\
 SFT Baseline  & Extended 2  & 62.90 & 67.85 & 71.28 & 27.71 & 62.75 & 82.49 \\
  SFT Baseline & 3 &  70.14 & 75.39 & 78.62 & 32.10 & 77.08 & 80.72 \\
\midrule
\multicolumn{7}{l}{\textit{Sub Benchmarks: V-COCO-based}} \\
\midrule
 SFT Baseline  & 1   & 55.44 & 71.31 & 77.60 & 34.27 & 74.46 & 81.02 \\
  SFT Baseline  & Extended 1  & 24.91 & 24.92 & 44.10 & 8.42 & 62.72 & 34.00 \\
 SFT Baseline  & 2  & 58.87 & 80.29 & 81.60 & 36.87 & 73.82 & 91.20 \\
 SFT Baseline  & Extended 2  & 46.72 & 67.06 & 70.60 & 25.05 & 56.89 & 93.02 \\
  SFT Baseline & 3 &  57.18 & 80.98 & 82.17 & 36.47 & 74.00 & 92.36 \\
 \midrule
\multicolumn{7}{l}{\textit{Sub Benchmarks: SWiG-HOI-based}} \\
\midrule
 SFT Baseline  & 1   & 49.63 & 55.78 & 64.81 & 21.23 & 57.55 & 74.16 \\
  SFT Baseline  & Extended 1  & 40.55 & 44.67 & 57.71 & 15.88 & 56.84 & 58.61 \\
 SFT Baseline  & 2  & 57.48 & 70.32 & 72.39 & 27.28 & 59.24 & 93.04 \\
 SFT Baseline  & Extended 2  & 57.16 & 70.37 & 72.44 & 27.82 & 59.32 & 93.01 \\
  SFT Baseline & 3 & 59.72 & 70.27 & 72.52 & 26.04 & 59.28 & 93.37  \\
\bottomrule
\end{tabular}
}
\caption{Experiment results comparison when Qwen2.5-VL-7B is fine-tuned on HOI datasets~\cite{carion2020end} using Supervised Fine-Tuning (SFT). ``Avg. Prec.'' means the precision averaged across test set and ``Avg. Rec.'' means the recall averaged across test set. }
\label{app_tab:SFT_cmp}
\end{table*}

\paragraph{Fine-tuned VLM Evaluation}
Table~\ref{app_tab:SFT_cmp} provides results of the baseline under a finetuned setup, Qwen2.5-VL-7B fine-tuned on our training dataset.
Before fine-tuning, VLMs typically exhibit much higher average precision than recall, reflecting a conservative prediction style that favors fewer outputs, shown in Table~1,2,3 (main paper). After training on our dataset, however, the model learns to adjust its strategy: recall surpasses precision across settings. This shift indicates that the model adapts to predict multiple interactions for a given question.
Predicting more for each question increases the chance of covering all correct answers, although this comes at the cost of lower precision.

\input{figures/result_dist_main}

Moreover, in Fig.~\ref{fig:results_analysis_main}(b), recall (red bins) decreases steadily from head to tail classes, revealing a clear head-class bias after fine-tuning.  
In contrast, the pre-trained VLM without fine-tuning shows no obvious head-class bias, as recall remains relatively flat across classes (Fig.~\ref{fig:results_analysis_main}(a)).  
Overall, while class imbalance has long been recognized as a challenge for HOI-specific methods, our benchmark demonstrates that fine-tuned VLMs are not immune to this issue.

\paragraph{Evaluation among all HICO-DET images}
Tables~\ref{tab:setting1_allhico},~\ref{tab:setting2_allhico},~\ref{tab:setting3_allhico} present results evaluated on the 9,546 questions derived from the entire HICO-DET test set, corresponding to the same three settings as in Tables~1,2,3 (main paper) of our benchmark. Across all metrics in Tables~\ref{tab:setting1_allhico},~\ref{tab:setting2_allhico},~\ref{tab:setting3_allhico} both HOI-specific methods and VLMs achieve consistently much higher scores, typically by around 8 – 12 points in Instance-F1 and Micro-F1, compared with their performance on our main benchmark. This substantial margin confirms that our benchmark is significantly more challenging than the original HICO-DET benchmark. Unlike HICO-DET, which contains many simple, single-person single-object scenes, our benchmark emphasizes multi-person and confusing interactions.
The noticeable performance gap validates the rationale of constructing a smaller but more challenging benchmark based on HICO-DET.

\begin{table*}[t]
\centering
\resizebox{1.9\columnwidth}{!}{
\begin{tabular}{l|c|c|c|c|c|c}
\toprule
{Method}
& Macro-F1 ($\%$)   
& Instance-F1 ($\%$)
& Micro-F1 ($\%$)
& EM ($\%$)
& Avg. Prec. ($\%$)
& Avg. Rec. ($\%$) \\
\midrule
\multicolumn{7}{l}{\textit{HOI-specific methods}} \\
\midrule
ADA-CM  & 58.54 & 68.49 & \textbf{79.95} & 26.15 & 90.80 & 71.41 \\
CMMP   & 58.49 & 68.00 & 79.51 & 25.84 & 89.66 & 71.43 \\
LAIN   & 55.51 & 66.32 & 77.84 & 25.79 & 87.50 & 70.11 \\
HOLa   & \textbf{58.83} & 68.41 & 79.53 & 26.57 & 89.51 & \textbf{71.56}  \\
CMD-SE & 52.36 & \textbf{69.73} & 79.80 & \textbf{32.83} & \textbf{90.81} & 71.18  \\
\midrule
\multicolumn{7}{l}{\textit{VLM zero-shot evaluation}} \\
\midrule
InternVL2.5-38B & 65.41 & 68.08 & 68.58 & 41.55 & 93.30 & 54.22 \\
InternVL3-38B  & 71.32 & 76.05 & 72.41 & 46.55 & \textbf{93.45} & 59.10 \\
Qwen2.5-VL-32B  & \textbf{74.89} & \textbf{82.15} & \textbf{80.05} & \textbf{53.15} & 89.25 & \textbf{72.57} \\
\midrule
LLaVA-OV-7B & 62.48 & 69.04 & 63.87 & 36.93 & 92.57 & 48.76 \\
InternVL3-8B   &  \textbf{71.92} & \textbf{78.21} & \textbf{75.49} & \textbf{50.74} & \textbf{93.86} & \textbf{63.13}  \\
Qwen2-VL-7B  & 44.77 & 42.84 & 49.05 & 19.24 & 93.57 & 33.24 \\
Qwen2.5-VL-7B & {68.49} & 74.10 & 69.53 & 42.55 & 93.11 & 55.48  \\
\bottomrule
\end{tabular}
}
 \caption{\textit{Setting~1} experiment results comparison, derived from the whole HICO-DET images with 9546 questions in total. Results are reported for VLMs and HOI-specific methods. Best performance within each group is highlighted in \textbf{bold}. ``Avg. Prec.'' means the precision averaged across test set and ``Avg. Rec.'' means the recall averaged across test set.}
\label{tab:setting1_allhico}
\end{table*}

\begin{table*}[t]
\centering
\resizebox{1.9\columnwidth}{!}{
\begin{tabular}{l|c|c|c|c|c|c}
\toprule
{Method}
& Macro-F1 ($\%$)   
& Instance-F1 ($\%$)
& Micro-F1 ($\%$)
& EM ($\%$)
& Avg. Prec. ($\%$)
& Avg. Rec. ($\%$) \\
\midrule
\multicolumn{7}{l}{\textit{VLM zero-shot evaluation}} \\
\midrule
InternVL2.5-38B & 59.91 & 61.82 & 63.97 & 38.07 & 92.18 & 48.98 \\
InternVL3-38B  & 71.41 & 78.72 & 76.99 & 51.99 & \textbf{93.58} & 65.39 \\
Qwen2.5-VL-32B  & \textbf{75.76} & \textbf{83.85} & \textbf{83.01} & \textbf{58.27} & 91.02 & \textbf{76.29} \\
\midrule
LLaVA-OV-7B & 64.28 & 70.75 & 65.35 & 40.60 & 92.42 & 50.54 \\
InternVL3-8B   &  64.40 & 67.55 & 67.63 & 42.34 & 90.26 & 54.07  \\
Qwen2-VL-7B  & 61.29 & 67.06 & 63.94 & 37.94 & 92.23 & 48.94 \\
Qwen2.5-VL-7B &  \textbf{67.23} & \textbf{74.33} & \textbf{70.59} & \textbf{45.06} & \textbf{92.69} & \textbf{57.00}  \\
\bottomrule
\end{tabular}
}
 \caption{\textit{Setting~2} experiment results comparison, derived from the whole HICO-DET images with 9546 questions in total. Results are reported for VLMs and HOI-specific methods. Best performance within each group is highlighted in \textbf{bold}. ``Avg. Prec.'' means the precision averaged across test set and ``Avg. Rec.'' means the recall averaged across test set.}
\label{tab:setting2_allhico}
\end{table*}

\begin{table*}[t]
\centering
\resizebox{1.9\columnwidth}{!}{
\begin{tabular}{l|c|c|c|c|c|c}
\toprule
{Method}
& Macro-F1 ($\%$)   
& Instance-F1 ($\%$)
& Micro-F1 ($\%$)
& EM ($\%$)
& Avg. Prec. ($\%$)
& Avg. Rec. ($\%$) \\
\midrule
\multicolumn{7}{l}{\textit{HOI-specific methods}} \\
\midrule
ADA-CM  & 56.03 & 62.08 & \textbf{76.66} & 38.68 & 90.32 & 66.59  \\
CMMP   & 55.87 & 61.47 & 76.11 & 38.35 & 89.03 & 66.46 \\
LAIN   & 53.33 & 60.42 & 74.98 & 37.80 & 87.24 & 65.73 \\
HOLa   & \textbf{56.69} & \textbf{62.32} & 76.54 & \textbf{39.69} & 88.87 & \textbf{67.22}  \\
CMD-SE & 51.82 & 58.81 & 73.72 & 38.60 & \textbf{90.99} & 61.97  \\
\midrule
\multicolumn{7}{l}{\textit{VLM zero-shot evaluation}} \\
\midrule
InternVL2.5-38B & 34.43 & 28.29 & 36.54 & 17.31 & 93.09 & 22.73 \\
InternVL3-38B  & 49.39 & 43.18 & 52.87 & 28.82 & \textbf{94.27} & 36.73  \\
Qwen2.5-VL-32B  & \textbf{64.24} & \textbf{65.18} & \textbf{73.00} & \textbf{45.09} & 91.39 & \textbf{60.77} \\
\midrule
LLaVA-OV-7B & - & -& -& -& -& - \\
InternVL3-8B   &  12.98 & 8.68 & 13.36 & 5.48 & 93.32 & 7.20  \\
Qwen2-VL-7B  & - & -& -& -& -& - \\
Qwen2.5-VL-7B &  \textbf{43.37} & \textbf{38.94} & \textbf{45.81} & \textbf{23.30} & \textbf{93.32} & \textbf{30.36}  \\
\bottomrule
\end{tabular}
}
 \caption{\textit{Setting~3} experiment results comparison, derived from the whole HICO-DET images with 9546 questions in total. Results are reported for VLMs and HOI-specific methods. Best performance within each group is highlighted in \textbf{bold}. ``Avg. Prec.'' means the precision averaged across test set and ``Avg. Rec.'' means the recall averaged across test set.}
\label{tab:setting3_allhico}
\end{table*}

\begin{table*}[t]
\centering
\resizebox{1.9\columnwidth}{!}{
\begin{tabular}{l|c|c|c|c|c|c|c}
\toprule
{Train set}
& Setting
& Macro-F1 ($\%$)   
& Instance-F1 ($\%$)
& Micro-F1 ($\%$)
& EM ($\%$)
& Avg. Prec. ($\%$)
& Avg. Rec. ($\%$) \\
\midrule
\midrule
  Ours & 1 & 87.75 & 94.60 & 94.64 & 85.02 & 94.58 & 94.69 \\
 V-COCO-based  & 1 & 75.12 & 83.41 & 81.85 & 53.45 & 87.35 & 77.00 \\
\midrule
 Ours & 2 & 85.88 & 92.87 & 93.57 & 83.56 & 92.48 & 94.70 \\
 V-COCO-based  & 2 & 72.94 & 81.54 & 81.57 & 52.91 & 85.40 & 78.08 \\
\midrule 
 Ours  & 3 & 81.72 & 81.78 & 89.51 & 73.64 & 93.12 & 86.18 \\
 V-COCO-based  & 3 & 68.32 & 69.83 & 76.81 & 44.91 & 86.26 & 69.22  \\
\bottomrule
\end{tabular}
}
\caption{Experiment results comparison, where the test set is derived from the whole HICO-DET images with 9546 questions in total. Qwen2.5-VL-7B is fine-tuned on HOI datasets~\cite{carion2020end} using Supervised Fine-Tuning (SFT). ``Avg. Prec.'' means the precision averaged across test set and ``Avg. Rec.'' means the recall averaged across test set. }
\label{app_tab:SFT_cmp_allhico}
\end{table*}

\subsection{Implementation Details}
\label{app_sec: imp_detail}

\noindent \textbf{Prompts for Benchmarking VLMs}
For general-purpose VLMs, we provide the question prompt together with explicit answer-format instructions.

In \textit{Setting 1} and \textit{Setting 2}, 
we obtain choice selection results with the following prompt for target-human or target human-object pair accordingly:

\begin{quote}
\small
\texttt{
Context: You are given an image $\langle$image$\rangle$ and a target person with a bounding box $\langle$ human box $\rangle$.
Question: Which of the following describes the interactions between the target person and any object in the image? Choices: (A) ...,(B) ...,(C) ...,(D)...
IMPORTANT: Reply with the letter(s) ONLY, separated by commas if multiple (e.g. A,B).
For example, if correct answers are (A) and (B), your output must be: A,B
Do NOT include any brackets or other symbols.
}
\end{quote}

\begin{quote}
\small
\texttt{
Context: You are given an image $\langle$image$\rangle$ and and two bounding boxes: - Person bbox: $\langle$ human box $\rangle$; - Object bbox: $\langle$ object box $\rangle$.
Question: Which of the following describes the interactions between the target person and the object? Choices: (A) ...,(B) ...,(C) ...,(D)...
IMPORTANT: Reply with the letter(s) ONLY, separated by commas if multiple (e.g. A,B).
For example, if correct answers are (A) and (B), your output must be: A,B
Do NOT include any brackets or other symbols.
}
\end{quote}

In \textit{Setting 3}, the prompt used is 

\begin{quote}
\small
\texttt{
Question: Which of the following properly describes the interactions in the image  $\langle$image$\rangle$? Choices: (A) ...,(B) ...,(C) ...,(D)...
IMPORTANT: Reply with the letter(s) ONLY, separated by commas if multiple (e.g. A,B).
For example, if correct answers are (A) and (B), your output must be: A,B
Do NOT include any brackets or other symbols.
}
\end{quote}

In \textit{Setting 1}, a VLM is required to predict the target person or human-object bounding boxes before the choice selection. 
We use prompt to localize humans or objects before HOI prediction. Specifically, we provide an image to the model together with a text prompt asking it to output bounding boxes in JSON format. 
For person detection only, the following prompt is used:

\begin{quote}
\small
\texttt{
Provide the bounding box coordinates for every single person in the input image.\\
The box coordinates represent as [x1, y1, x2, y2], where x is the horizontal pixel coordinate from the left edge, and y is the vertical pixel coordinate from the top edge.\\
Return the detection results in JSON format strictly.\\
For example:\\
\{ "boxes": [[32, 109, 644, 418], [517, 0, 644, 23], [100, 50, 160, 200]] \}
}
\end{quote}

For detecting both persons and objects, we modify the prompt to:

\begin{quote}
\small
\texttt{
Provide the bounding box coordinates for all visible objects and humans in the input image based on the following object list: \{OBJ\_IDX\_TO\_OBJ\_NAME\}.\\
The box coordinates represent as [x1, y1, x2, y2], where x is the horizontal pixel coordinate from the left edge, and y is the vertical pixel coordinate from the top edge.\\
Return the detection results in JSON format strictly.\\
For example:\\
\{ "boxes": [[32, 109, 644, 418], [517, 0, 644, 23], [100, 50, 160, 200]],\\
\ \ "labels": ["person", "bench", "cup"] \}\\
Only include objects from the given list. Ensure the output is a valid JSON dictionary without additional comments, and that the lengths of the boxes and labels arrays are equal.
}
\end{quote}

Here, \texttt{OBJ\_IDX\_TO\_OBJ\_NAME} denotes the 80 predefined object classes in HICO-DET. After obtaining the detection results, we compute the Intersection-over-Union (IoU) between the predicted boxes and the ground-truth boxes. Predictions with IoU greater than 0.5 are considered correct and passed to the HOI choice selection step, where the detected boxes are used to localize the corresponding human or human–object pair.

\noindent \textbf{Bounding Box Process for VLMs}
In \textit{Setting 1} and \textit{Setting 2}, the input requires bounding boxes of the target person. Since different VLMs preprocess images in different ways, we adapt the bounding boxes accordingly to ensure consistency with the model input. Specifically,
Qwen2/2.5-VL resizes input images such that both height and width are multiples of 14, while Qwen3-VL adjusts images to multiples of 16. We therefore resize the bounding boxes proportionally to the resized image coordinates.
InternVL2.5/3 does not fix the image size but internally normalizes it. To align with its view of the image, we first query the model with a prompt asking for the perceived input resolution: 
\begin{quote}
\small
\texttt{
Please provide the coordinates for the bottom-right point of the input image. 
Assume the coordinate system origin is at the top-left of the image, 
with x increasing to the right and y increasing downward. 
Return the coordinates as [width, height] in JSON format strictly. 
For example:
[638, 415].
}
\end{quote}

Based on its returned size, we then rescale the bounding boxes into that coordinate system.
LLaVA-OV takes the original image size directly as input. In this case, we use the original bounding boxes without additional processing.
This preprocessing ensures that the bounding boxes we provide are always aligned with how each model internally processes the input image.

\noindent \textbf{Prompt for Coarse Screening}
We provide the prompt template used for the coarse screening stage. This stage serves as an initial screening to identify negative candidates before applying the fine-grained manual refinement. 

Below is the general template we used for GPT-4.1 to separate semantically consistent or inconsistent candidates:

\begin{quote}
\small
\texttt{
You are given an image and a human bounding box: $\langle$human box$\rangle$. \\
A list of candidate interactions is provided: $\langle$HOI candidates$\rangle$. \\
The ground-truth interaction is: $\langle$annotated ground-truth HOI classes$\rangle$.
Find interactions that are clearly different and unrelated to the ground-truth interaction in the image.  \\
Any visually or semantically similar interactions (e.g., synonyms, paraphrases, same action with different wording) must NOT be selected. \\
Return ONLY the NEGATIVE group as a JSON list of action+object phrases. Do not include explanations. Limit to under 50 words. \\
Examples: \\
- GT: "ride a/an bike" \\
- Negative: ["inspect a/an bike", "repair a/an bike", "wash a/an bike"] \\
Now, based on the image and the ground-truth interaction, return the negative group from the candidate list.  
}
\end{quote}

Then, we construct prompts for Qwen2.5-VL-32B and GPT-4o to re-evaluate the inconsistent candidates returned by GPT-4.1, ensuring that these cases are indeed true negatives.

\begin{quote}
\small
\texttt{
You are given an image and a human bounding box: $\langle$human box$\rangle$. \\
Definitions: \\
- Positive: Interactions that appear in the image or are semantically/visually related or they may occur simultaneously. \\
- Negative: Interactions that are clearly different or unrelated. \\
Question: \\
Is the interaction $\langle$an HOI negative candidate$\rangle$ positive or negative with respect to the image? \\
Please answer in the following format: \\
Answer: Positive \\
or \\
Answer: Negative
}
\end{quote}

This careful two-stage verification ensures that our coarse screening does not introduce bias to VLMs. As shown in Table 4 of the main paper, HOI-specific models even more agree with the negatives identified through this process than other VLM baselines, confirming that only clear negative HOIs, validated independently by three VLMs, are selected.

\section{Evaluation Metrics Details}

Let $Q$ denote the set of all evaluation questions. For each question $q \in Q$, let $P_q$ be the set of predicted interaction labels and $G_q$ the ground-truth set of positive choices. 
Macro-F1 evaluates performance in a class-balanced manner. 
Let $\mathcal{C}$ denote the set of HOI classes. 
For each class $c \in \mathcal{C}$, we compute the F1-score over all questions involving $c$, denoted as $\text{F1}_c$. 
Macro-F1 is then obtained by averaging $\text{F1}_c$ across all classes.
%
\begin{equation}
\begin{aligned}
\small
  &  \text{Macro-F1} = \frac{1}{|\mathcal{C}|} \sum_{c \in \mathcal{C}} 
\frac{2 \sum_{q} \mathbf{1}[c \in P_q \cap G_q]}
{\sum_{q} \mathbf{1}[c \in P_q] + \sum_{q} \mathbf{1}[c \in G_q]},
\end{aligned}
\end{equation}
where $\mathbf{1}[\cdot]$ is the indicator function, which equals 1 if the condition is true and 0 otherwise.

Instance-F1 measures performance at the question level. 
For each $q$, we compute the F1-score between $P_q$ and $G_q$, and then average over all questions to obtain the overall score:
\begin{equation}
\text{Instance-F1} = \frac{1}{|Q|} \sum_{q \in Q} \text{F1}(q)
= \frac{1}{|Q|} \sum_{q \in Q} \frac{2|P_q \cap G_q|}{|P_q| + |G_q|}.
\end{equation}
Here $| \cdot |$ denotes set cardinality, and $P_q \cap G_q$ is the set of correctly predicted labels for question $q$.

Micro-F1 measures overall performance by aggregating predictions across all questions and computing a single F1-score from the total number of predicted and ground-truth labels:
\begin{equation}
\text{Micro-F1} = \frac{2 \sum_{q} |P_q \cap G_q|}{\sum_{q} |P_q| + \sum_{q} |G_q|}.
\end{equation}

Finally, we adopt Exact Match Accuracy (EM), which checks whether the predicted interaction set for a question exactly matches the ground-truth set.  
Unlike the exact-match mAP metric in traditional HOI benchmarks, which is often affected by incomplete annotations and penalizes unlabeled interactions, our multiple-choice design mitigates this issue through curated negatives.  
Thus, EM provides a complementary measure of strict correctness: it reports how often the model’s predictions are entirely correct.  
\begin{equation}
\text{EM} = \frac{1}{|Q|} \sum_{q \in Q} \mathbf{1}[P_q = G_q].
\end{equation}

\end{document}

%% file: figures/VLM_problems_main.tex
\begin{figure}[t]
    \centering

    \begin{minipage}[b]{0.23\linewidth}
        \centering
        \subcaption*{(a)}%
        \includegraphics[width=\linewidth]{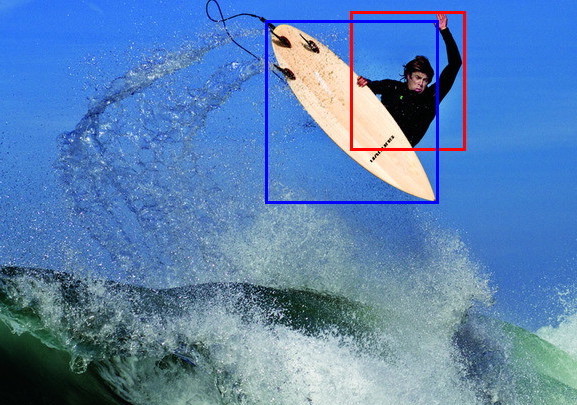}
    \end{minipage}%
    \hfill
    \begin{minipage}[b]{0.25\linewidth}
        \footnotesize
        \textbf{GT: } \\
        jump surfboard \\
        \textcolor{red}{hold surfboard}

        \textbf{VLM preds: } \\
        jump surfboard 
    \end{minipage}
    \begin{minipage}[b]{0.23\linewidth}
    \subcaption*{ (b) } %
        \includegraphics[width=\linewidth]{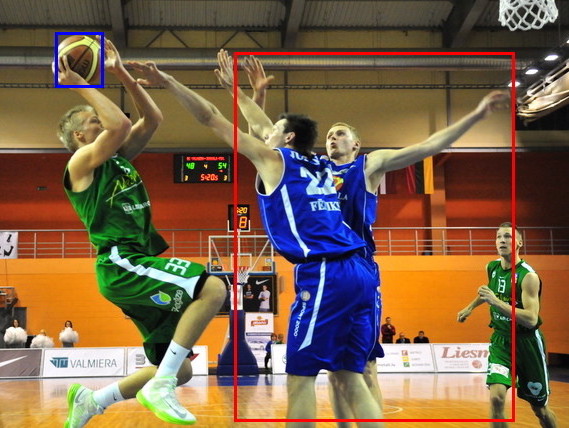}
    \end{minipage}%
    \hfill
    \begin{minipage}[b]{0.25\linewidth}
        \footnotesize
        \textbf{GT: } \\
        block sportsball

        \textbf{VLM preds: } \\
        block sportball \\
        \textcolor{red}{throw sportball} 
    \end{minipage}
    \vspace{0.1em} 
    \begin{minipage}[b]{0.23\linewidth}
        \centering
        \subcaption*{(c)}%
        \includegraphics[width=0.6\linewidth]{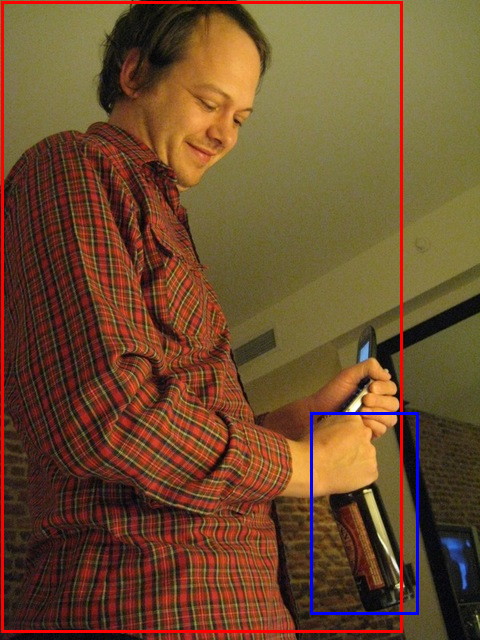}
    \end{minipage}%
    \hfill
    \begin{minipage}[b]{0.25\linewidth}
        \footnotesize
        \textbf{GT: } \\
        hold cellphone

        \textbf{VLM preds: } \\
        hold cellphone \\
        \textcolor{red}{text on cellphone}  
    \end{minipage}
    \begin{minipage}[b]{0.23\linewidth}
    \subcaption*{ (d) } %
        \includegraphics[width=\linewidth]{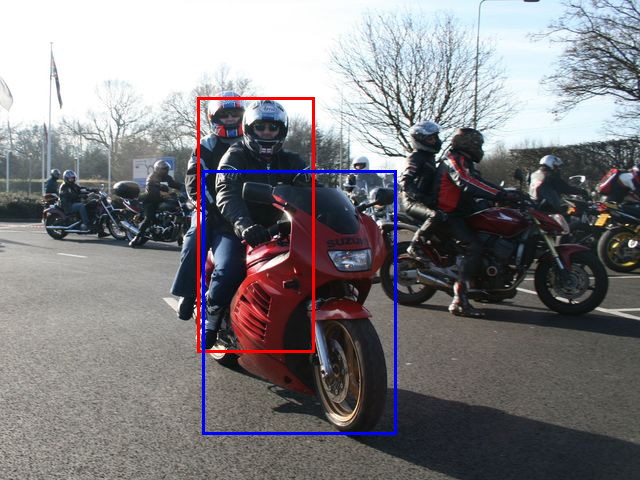}
    \end{minipage}%
    \hfill
    \begin{minipage}[b]{0.25\linewidth}
        \footnotesize
        \textbf{GT: } \\
        ride motorcycle

        \textbf{VLM preds: } \\
        ride motorcycle \\
        \textcolor{red}{hop on motorcycle} 
    \end{minipage}
    
    \caption{Illustration of VLM (Qwen2.5-VL-32B) failure cases in \textit{Setting 1}, and \textcolor{red}{red HOI classes} refer to missing ground-truth interactions or incorrect predictions.
    }
    \label{fig:VLM_problem_main}
\end{figure}

%% file: figures/dataset_dist.tex
\begin{figure}[t]
    \centering
    \begin{minipage}[b]{0.98\linewidth}
		\includegraphics[width=\linewidth]{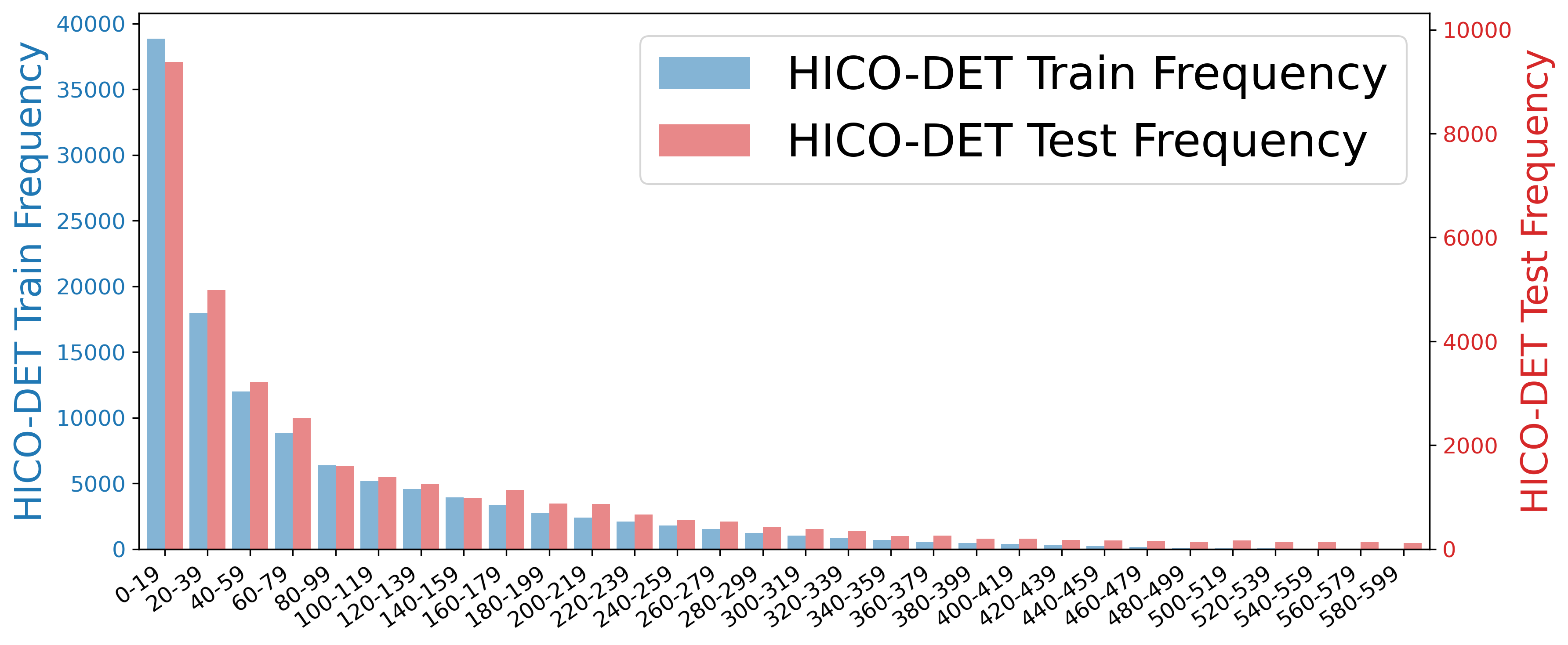}
	\end{minipage}   
    
    \begin{minipage}[b]{0.98\linewidth}
		\includegraphics[width=\linewidth]{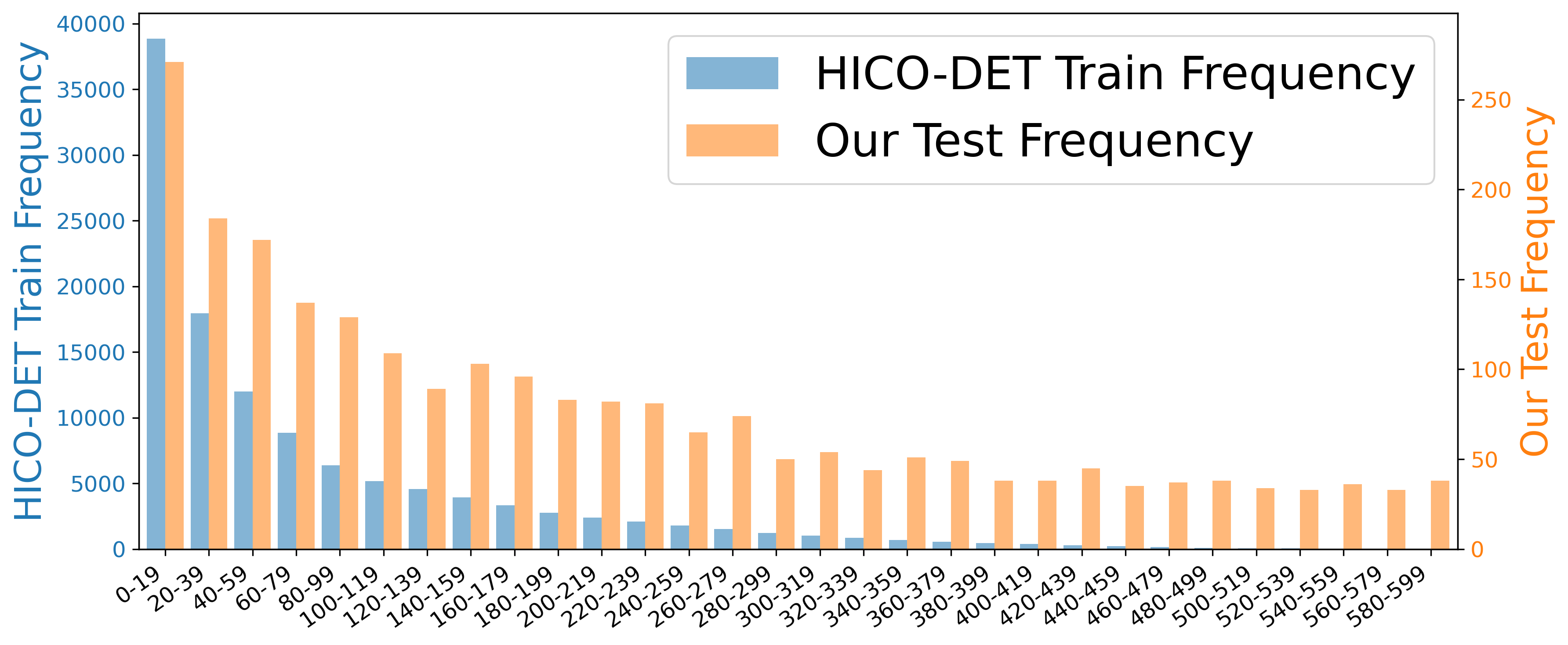}
	\end{minipage}

\caption{Comparison of HOI class frequency distributions between training and testing sets.} 
\label{fig:dataset_dist}
\end{figure}

%% file: figures/dataset_issue.tex
\begin{figure*}[t]
    \centering

    \begin{minipage}[b]{0.16\linewidth}
     \subcaption*{ (a) } %
        \includegraphics[width=0.95\linewidth]{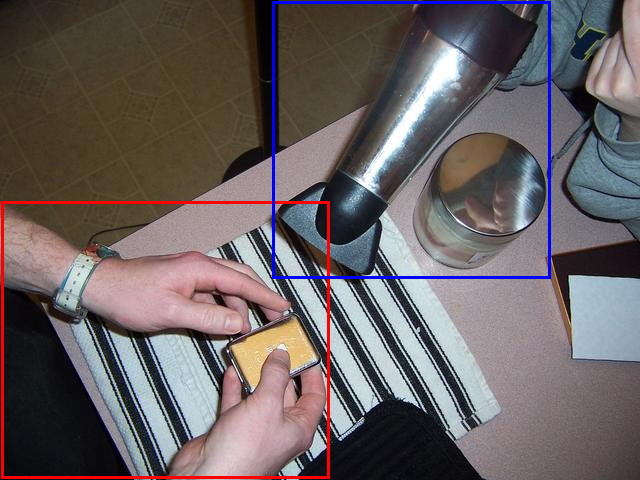}
    \end{minipage}%
    \hfill
    \begin{minipage}[b]{0.17\linewidth}
        \small
        \textbf{Annotated HOI: } \\
        no interaction with hair drier

        \textbf{Ambiguous negative: } \\
        inspect hair drier 
    \end{minipage}
    \begin{minipage}[b]{0.16\linewidth}
      \subcaption*{ (b) } %
        \includegraphics[width=0.95\linewidth]{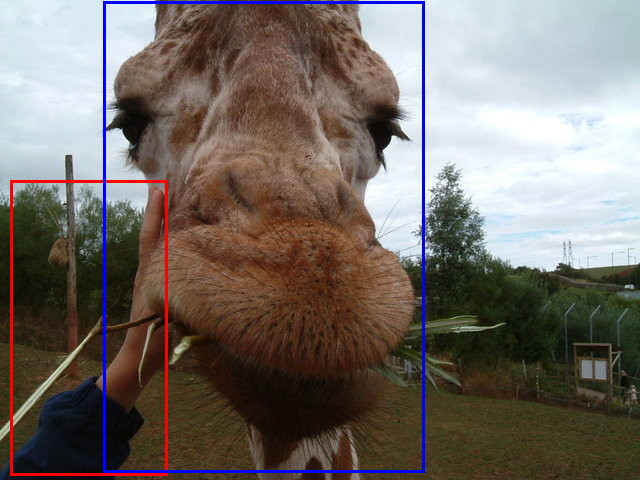}
    \end{minipage}%
    \hfill
    \begin{minipage}[b]{0.17\linewidth}
        \small
        \textbf{Annotated HOI: } \\
        pet giraffe

        \textbf{Ambiguous negative: } \\
        inspect giraffe \\

    \end{minipage}
    \begin{minipage}[b]{0.16\linewidth}
      \subcaption*{ (c) } %
        \includegraphics[width=0.95\linewidth]{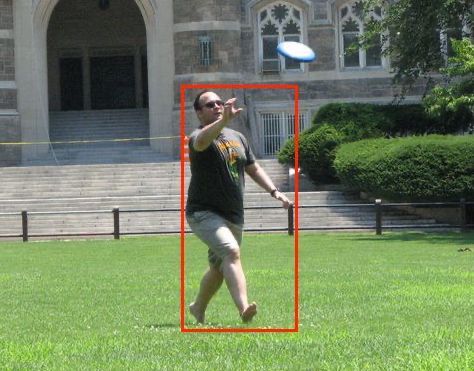}
    \end{minipage}%
    \hfill
    \begin{minipage}[b]{0.17\linewidth}
        \small
        \textbf{Annotated HOI: } \\
        catch frisbee

        \textbf{Ambiguous negative: } \\
        throw frisbee \\

    \end{minipage}

    \vspace{1em} 
    \begin{minipage}[b]{0.16\linewidth}
     \subcaption*{ (d) } %
        \includegraphics[width=0.95\linewidth]{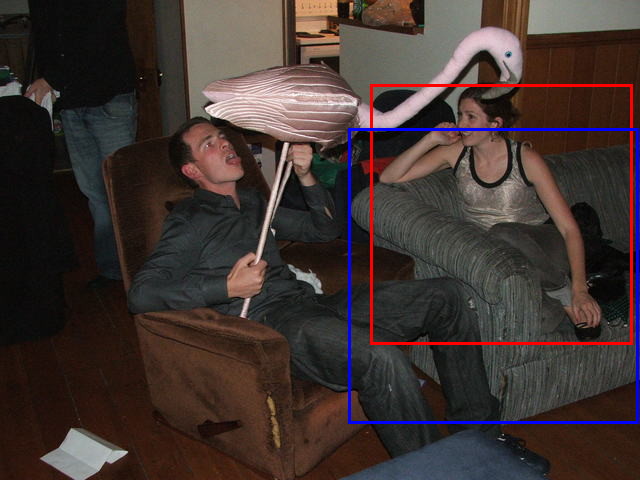}
    \end{minipage}%
    \hfill
    \begin{minipage}[b]{0.17\linewidth}
        \small
        \textbf{Annotated HOI: } \\
        sit on couch (female) \\
        \textbf{Missing HOI: } \\
        sit on couch (male) \\

    \end{minipage}
    \begin{minipage}[b]{0.16\linewidth}
        \subcaption*{ (e) } %
        \includegraphics[width=0.95\linewidth]{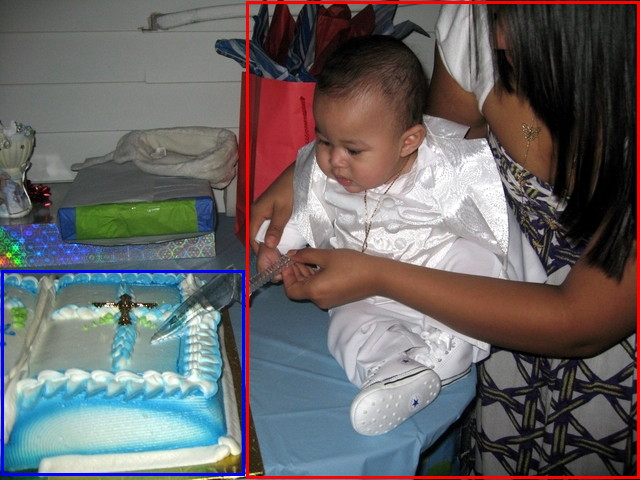}
    \end{minipage}%
    \hfill
    \begin{minipage}[b]{0.17\linewidth}
        \small
        \textbf{Annotated HOI: } \\
        cut cake

        \textbf{Missing HOI: } \\
        cut with knife \\

    \end{minipage}
    \begin{minipage}[b]{0.16\linewidth}
     \subcaption*{ (f) } %
        \includegraphics[width=0.95\linewidth]{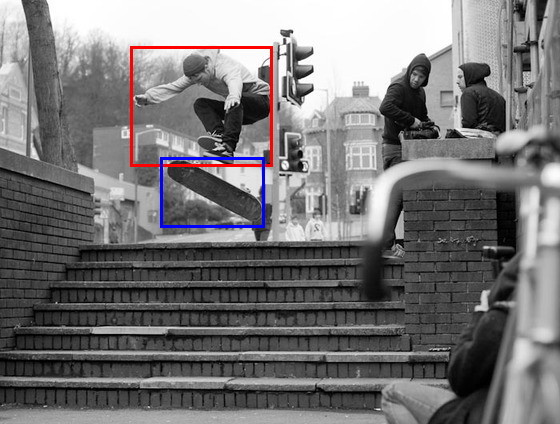}
    \end{minipage}%
    \hfill
        \begin{minipage}[b]{0.17\linewidth}
        \small
        \textbf{Annotated HOI: } \\
        flip skateboard

        \textbf{Missing HOI: } \\
        jump skateboard \\

    \end{minipage}

    \caption{Illustration of HOI dataset annotation problems. 
    }
    \label{fig:hico_dataset_problem}
\end{figure*}

%% file: figures/remove_imgs.tex
\begin{figure*}[t]
    \centering
    \begin{minipage}[b]{0.12\linewidth}
    \centering
       \subcaption*{\small ride elephant} %
		\includegraphics[width=\linewidth]{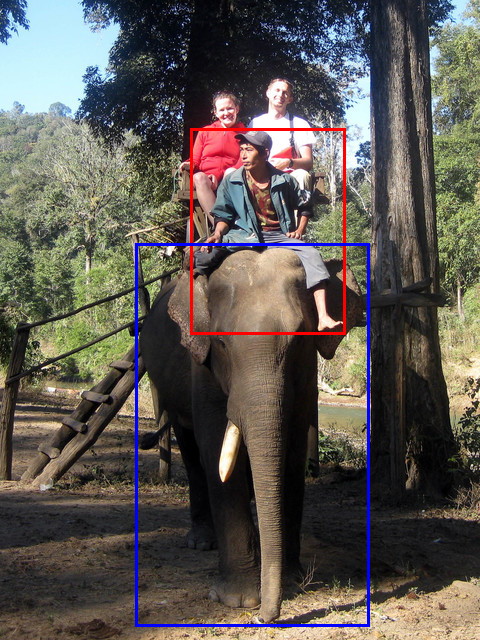}
	\end{minipage}
    \begin{minipage}[b]{0.12\linewidth}
    \centering
        \subcaption*{\small ride bicycle} %
		\includegraphics[width=\linewidth]{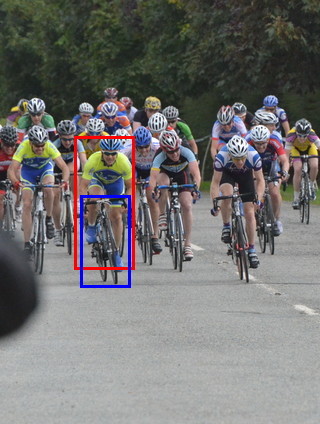}
	\end{minipage}
    \begin{minipage}[b]{0.12\linewidth}
    \centering
        \subcaption*{\small fly kite} %
		\includegraphics[width=0.89\linewidth]{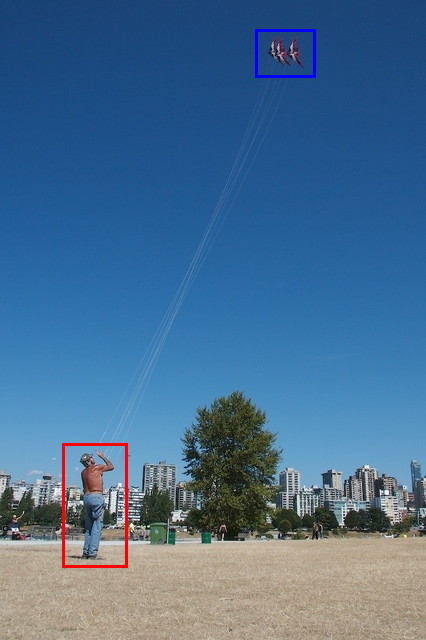}
	\end{minipage}	
    \begin{minipage}[b]{0.185\linewidth}
        \subcaption*{\small jump horse} %
		\includegraphics[width=\linewidth]{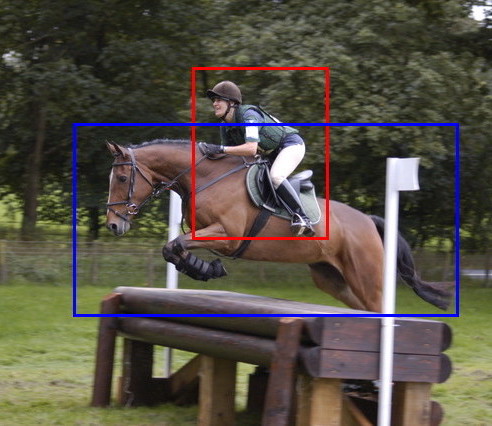}
	\end{minipage}	
    \begin{minipage}[b]{0.13\linewidth}
    \centering
        \subcaption*{\small wash toothbrush} %
		\includegraphics[width=\linewidth]{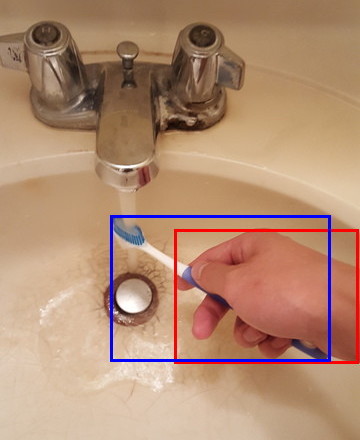}
	\end{minipage}	
    \begin{minipage}[b]{0.213\linewidth}
    \centering
        \subcaption*{\small drink with bottle} %
		\includegraphics[width=\linewidth]{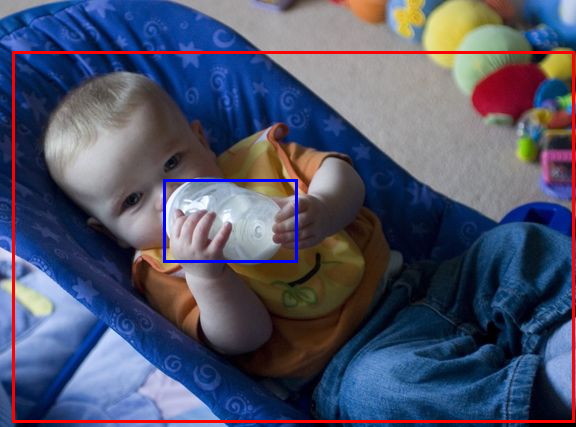}
	\end{minipage}

    \begin{minipage}[b]{0.12\linewidth}
         \centering
		\includegraphics[width=\linewidth]{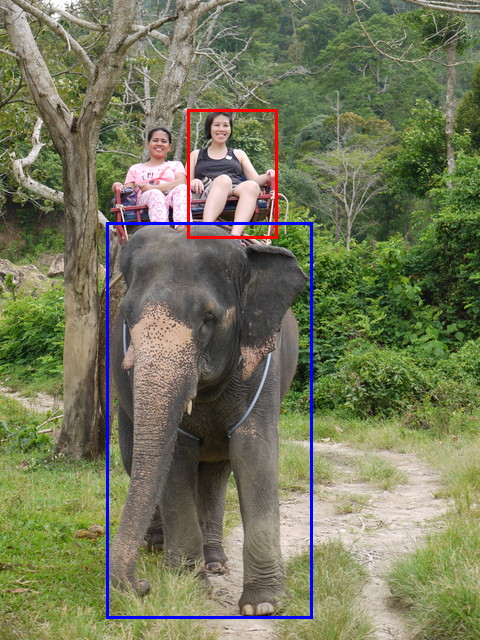}
	\end{minipage}
    \begin{minipage}[b]{0.12\linewidth}
    \centering
		\includegraphics[width=0.9\linewidth]{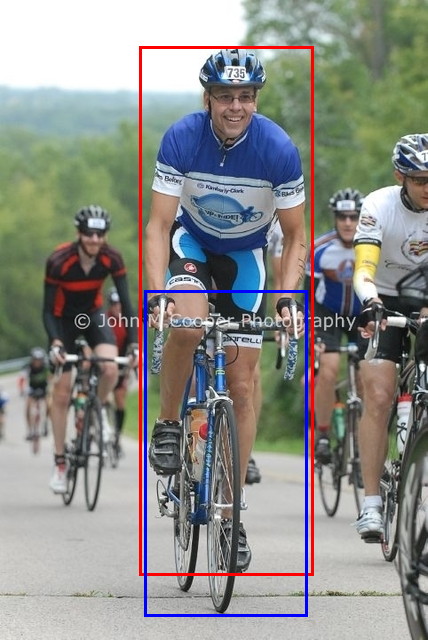}
	\end{minipage}
    \begin{minipage}[b]{0.12\linewidth}
        \centering
		\includegraphics[width=\linewidth]{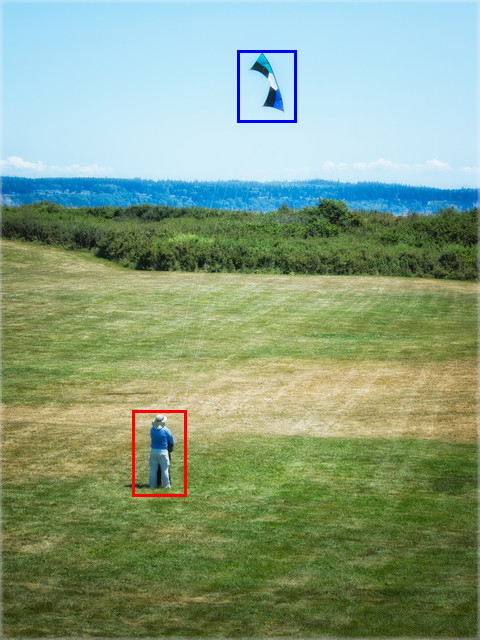}
	\end{minipage}	
    \begin{minipage}[b]{0.185\linewidth}
		\includegraphics[width=\linewidth]{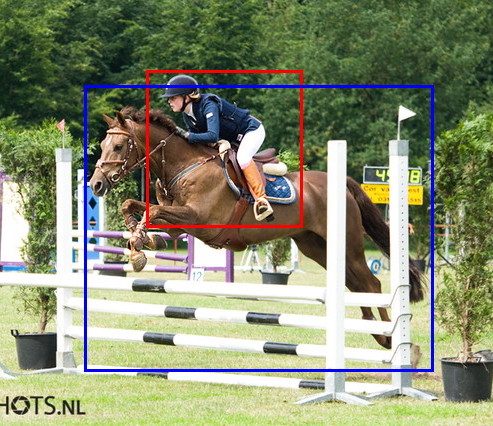}
	\end{minipage}	
    \begin{minipage}[b]{0.13\linewidth}
    \centering
		\includegraphics[width=\linewidth]{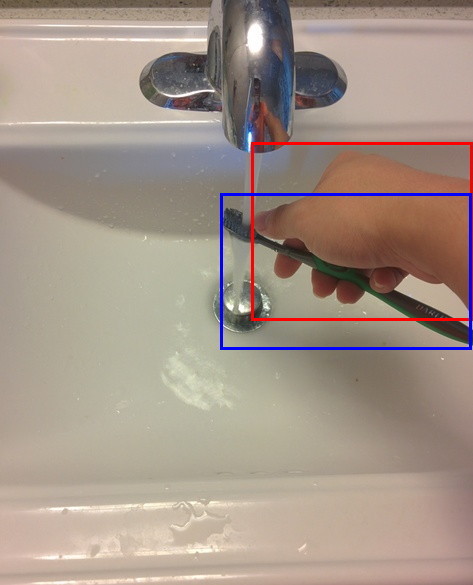}
	\end{minipage}	
    \begin{minipage}[b]{0.213\linewidth}
    \centering
		\includegraphics[width=\linewidth]{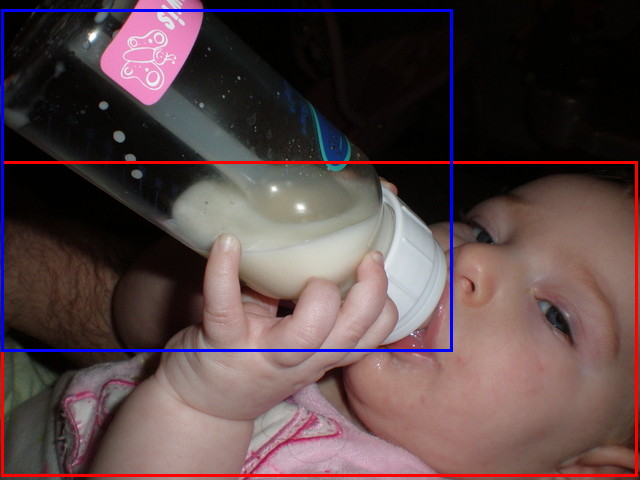}
	\end{minipage}	
    
\caption{ Removed simple scenes from HICO-DET test set during manual refinement stage.
} 
\label{fig:remove_img}
\end{figure*}

%% file: figures/hard_choice.tex
\begin{figure*}[t]
    \centering

    \begin{minipage}[b]{0.16\linewidth}
    \centering
      \subcaption*{ (a) } %
        \includegraphics[width=0.7\linewidth]{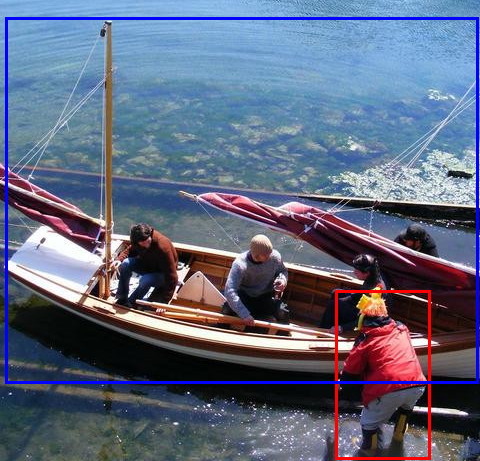}
    \end{minipage}%
    \hfill
    \begin{minipage}[b]{0.17\linewidth}
        \small
        \textbf{Hard positive: } \\
        launch boat 

        \textbf{Hard negative: } \\
        sit on boat; \\
        ride boat
    \end{minipage}
    \begin{minipage}[b]{0.16\linewidth}
    \centering
      \subcaption*{ (b) } %
        \includegraphics[width=0.9\linewidth]{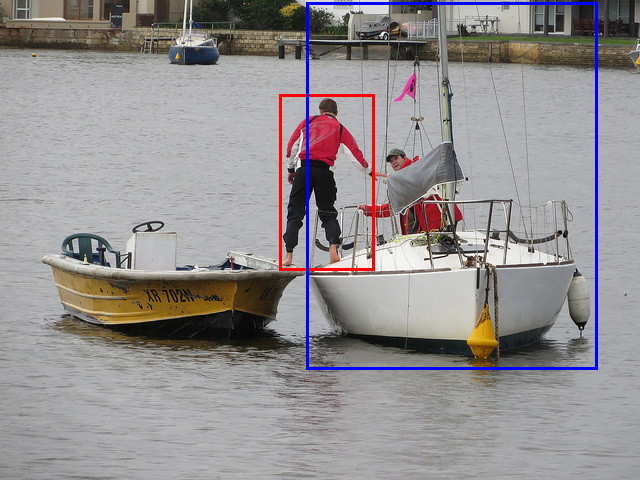}
    \end{minipage}%
    \hfill
    \begin{minipage}[b]{0.17\linewidth}
        \small
        \textbf{Annotated HOI: } \\
        board boat 

        \textbf{Hard positive: } \\
        exit boat \\
    \end{minipage}
    \begin{minipage}[b]{0.16\linewidth}
     \subcaption*{ (c) } %
        \includegraphics[width=0.95\linewidth]{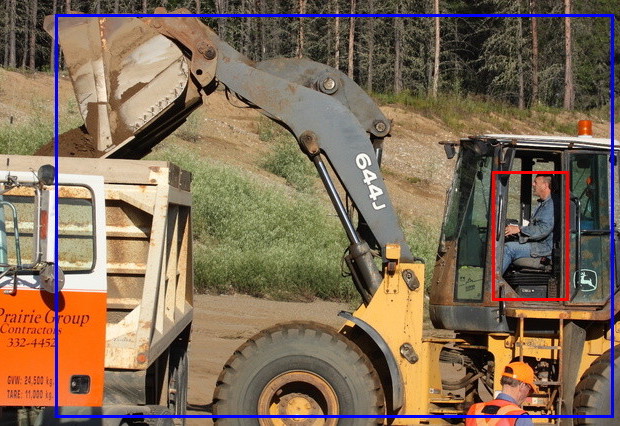}
    \end{minipage}%
    \hfill
    \begin{minipage}[b]{0.17\linewidth}
        \small
        \textbf{Hard positive: } \\
        load truck; \\
        sit on truck \\ 
        \\
        
    \end{minipage}

    \vspace{1em} 
    \begin{minipage}[b]{0.16\linewidth}
     \subcaption*{ (d) } %
        \includegraphics[width=0.95\linewidth]{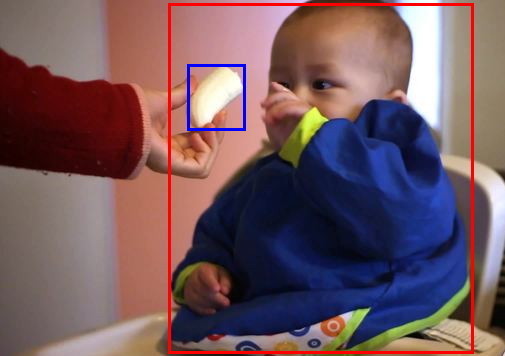}
    \end{minipage}%
    \hfill
        \begin{minipage}[b]{0.17\linewidth}
        \small
        \textbf{Annotated HOI: } \\
        inspect banana

        \textbf{Hard negative: } \\
        hold banana \\

    \end{minipage}
    \begin{minipage}[b]{0.16\linewidth}
     \subcaption*{ (e) } %
        \includegraphics[width=0.95\linewidth]{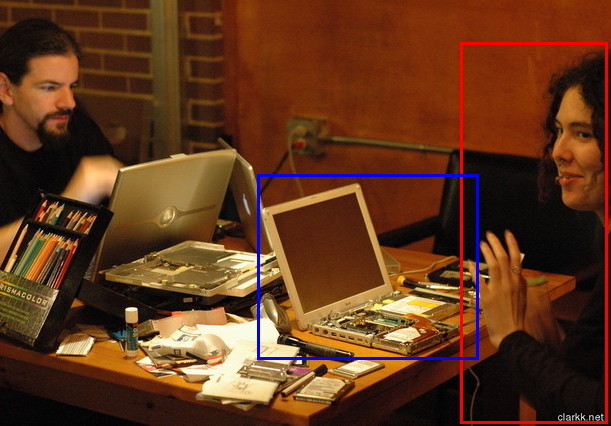}
    \end{minipage}%
    \hfill
    \begin{minipage}[b]{0.17\linewidth}
        \small
        \textbf{Annotated HOI: } \\
        repair laptop \\
        \textbf{Hard negative: } \\
        read laptop; \\
        type on laptop 

    \end{minipage}
    \begin{minipage}[b]{0.16\linewidth}
    \centering
        \subcaption*{ (f) } %
        \includegraphics[width=0.8\linewidth]{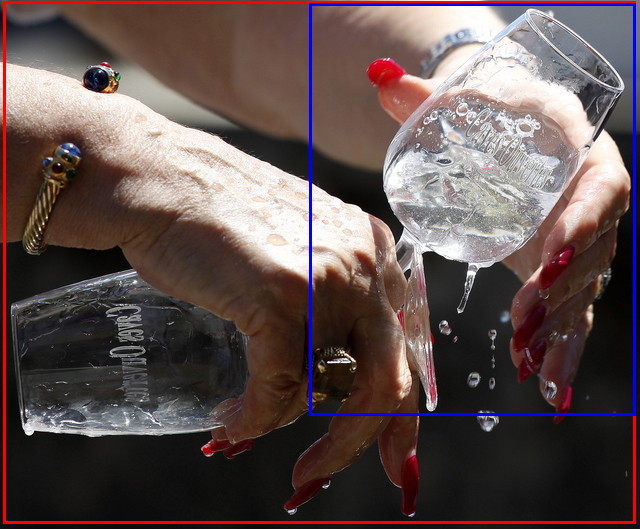}
    \end{minipage}%
    \hfill
    \begin{minipage}[b]{0.17\linewidth}
        \small
        \textbf{Annotated HOI: } \\
        wash wine glass

        \textbf{Hard negative: } \\
        fill wine glass \\

    \end{minipage}

    \caption{Hard choices modified during manual refinement. 
    }
    \label{fig:hard_choice}
\end{figure*}

%% file: figures/VLM_problems.tex
\begin{figure*}[t]
    \centering

    \begin{minipage}[b]{0.16\linewidth}
    \centering
    \subcaption*{ (a) } %
        \includegraphics[width=0.55\linewidth]{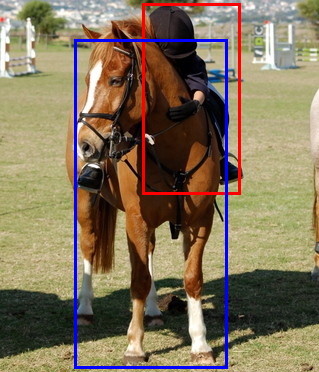}
    \end{minipage}%
    \hfill
    \begin{minipage}[b]{0.17\linewidth}
    \centering
        \small
        \textbf{Target person HOI: } \\
        ride horse; \\
        \textcolor{red}{hug horse} \\
        \textbf{VLM predictions: } \\
        hug horse 

    \end{minipage}
    \begin{minipage}[b]{0.16\linewidth}
    \centering
    \subcaption*{ (b) } %
        \includegraphics[width=0.95\linewidth]{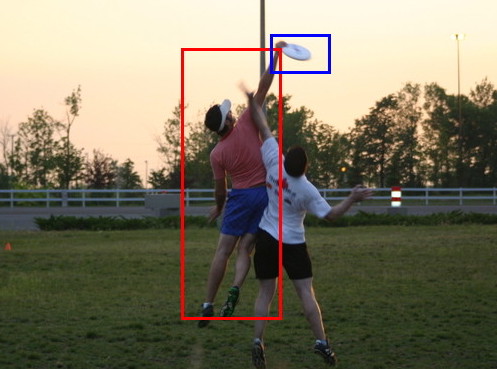}
    \end{minipage}%
    \hfill
    \begin{minipage}[b]{0.17\linewidth}
    \centering
        \small
        \textbf{Target person HOI: } \\
        catch frisbee; \\
        \textcolor{red}{hold frisbee}

        \textbf{VLM predictions: } \\
        catch frisbee 
    \end{minipage}
      \begin{minipage}[b]{0.16\linewidth}
    \centering
    \subcaption*{ (c) } %
        \includegraphics[width=0.48\linewidth]{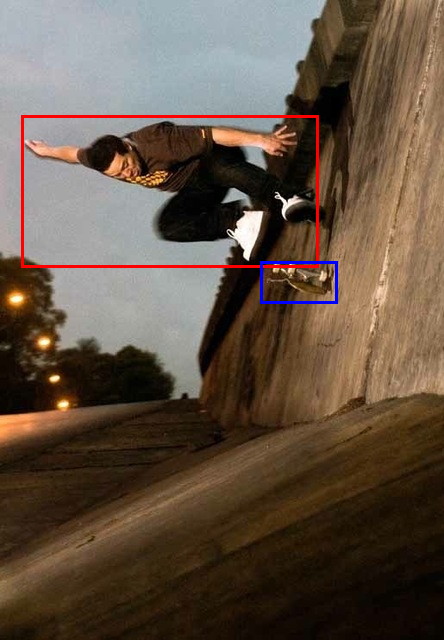}
    \end{minipage}%
    \hfill
    \begin{minipage}[b]{0.17\linewidth}
    \centering
        \small
        \textbf{Target person HOI: } \\
        jump skateboard; \\
        \textcolor{red}{flip skateboard}

        \textbf{VLM predictions: } \\
        jump skateboard
    \end{minipage}  

    \begin{minipage}[b]{0.16\linewidth}
    \centering
    \subcaption*{ (d) } %
        \includegraphics[width=0.95\linewidth]{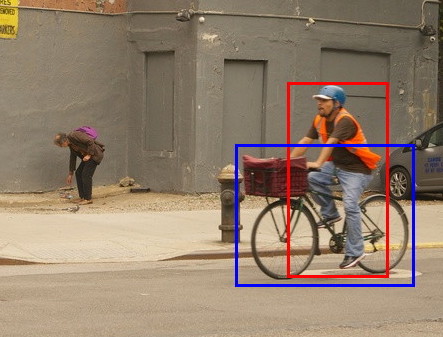}
    \end{minipage}%
    \hfill
    \begin{minipage}[b]{0.17\linewidth}
    \centering
        \small
        \textbf{Target person HOI: } \\
        ride bicycle

        \textbf{VLM predictions: } \\
        ride bicycle; \\
        \textcolor{red}{wear backpack($\times$)} 
    \end{minipage}
    \begin{minipage}[b]{0.16\linewidth}
    \centering
    \subcaption*{ (e) } %
        \includegraphics[width=0.95\linewidth]{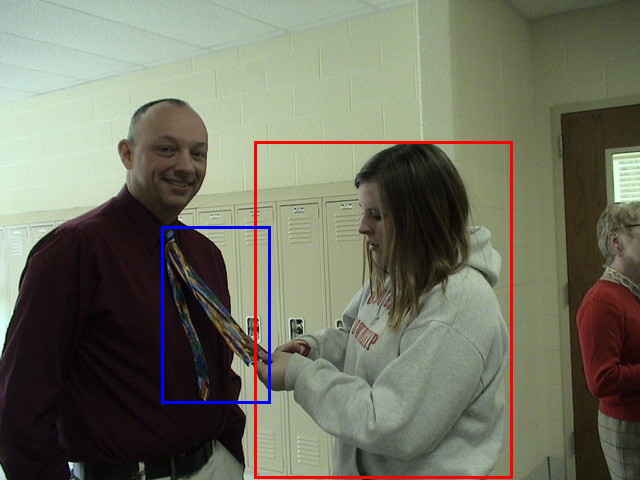}
    \end{minipage}%
    \hfill
    \begin{minipage}[b]{0.17\linewidth}
    \centering
        \small
        \textbf{Target person HOI: } \\
        cut tie

        \textbf{VLM predictions: } \\
        cut tie; \\
        \textcolor{red}{wear tie ($\times$)} 
    \end{minipage}
    \begin{minipage}[b]{0.16\linewidth}
    \centering
    \subcaption*{ (f) } %
        \includegraphics[width=0.95\linewidth]{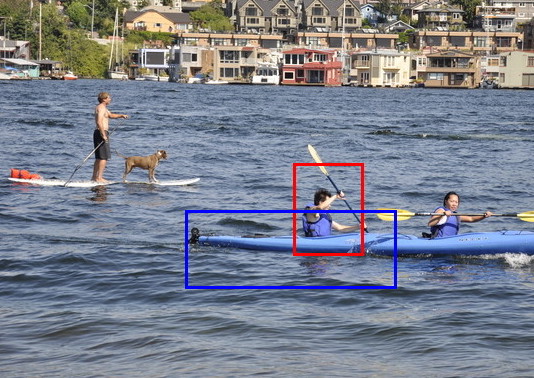}
    \end{minipage}%
    \hfill
    \begin{minipage}[b]{0.17\linewidth}
    \centering
        \small
        \textbf{Target person HOI: } \\
        row boat

        \textbf{VLM predictions: } \\
        row boat; \\
        \textcolor{red}{stand on boat ($\times$)} 
    \end{minipage}
    
    \begin{minipage}[b]{0.16\linewidth}
    \centering
    \subcaption*{ (g) } %
        \includegraphics[width=0.75\linewidth]{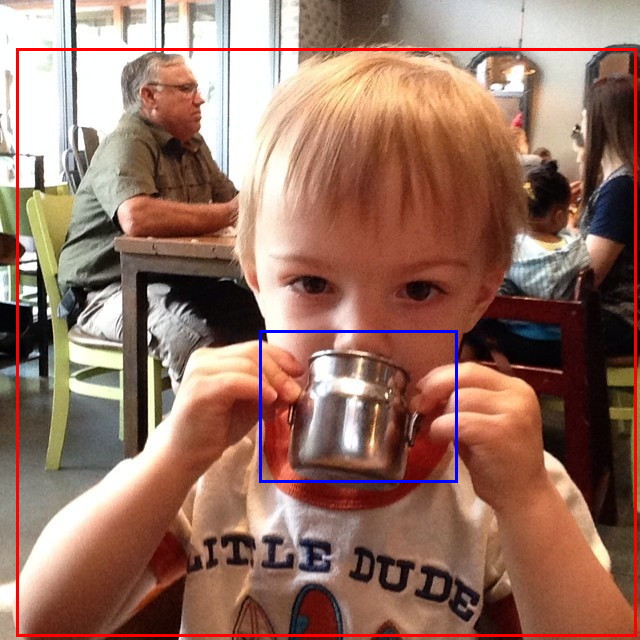}
    \end{minipage}%
    \hfill
        \begin{minipage}[b]{0.17\linewidth}
        \centering
        \small
        \textbf{Target person HOI: } \\
        drink with cup

        \textbf{VLM predictions: } \\
        drink with cup; \\
        \textcolor{red}{inspect cup ($\times$)} 
    \end{minipage}
    \begin{minipage}[b]{0.16\linewidth}
    \centering
    \subcaption*{ (h) } %
        \includegraphics[width=0.52\linewidth]{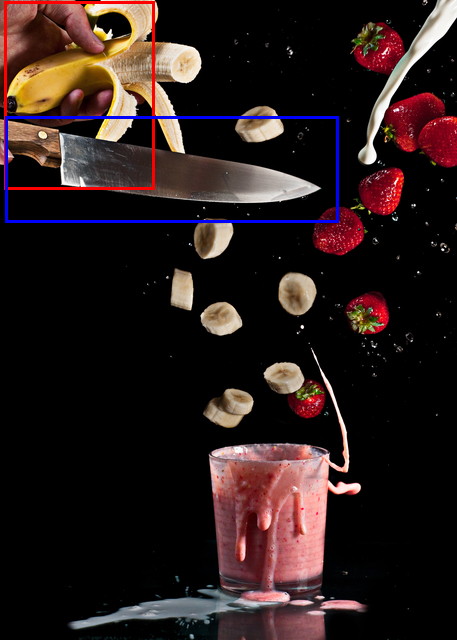}
    \end{minipage}%
    \hfill
        \begin{minipage}[b]{0.17\linewidth}
        \centering
        \small
        \textbf{Target person HOI: } \\
        cut banana

        \textbf{VLM predictions: } \\
        cut banana; \\
        \textcolor{red}{peel banana ($\times$)} 
    \end{minipage}
    \begin{minipage}[b]{0.16\linewidth}
    \centering
    \subcaption*{ (i) } %
        \includegraphics[width=0.95\linewidth]{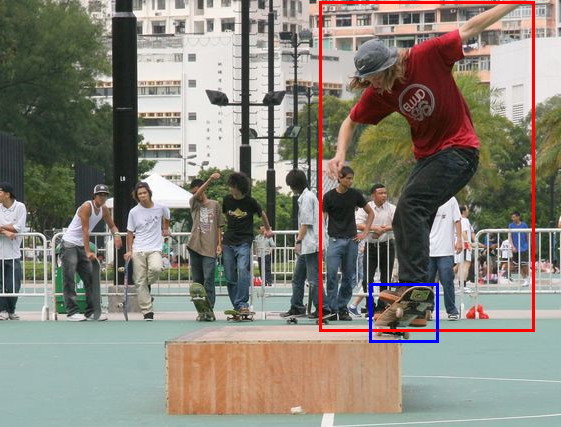}
    \end{minipage}%
    \hfill
        \begin{minipage}[b]{0.17\linewidth}
        \centering
        \small
        \textbf{Target person HOI: } \\
        grind skateboard

        \textbf{VLM predictions: } \\
        \textcolor{red}{flip skateboard ($\times$)} \\
    \end{minipage}
    
    \begin{minipage}[b]{0.16\linewidth}
    \centering
    \subcaption*{ (j) } %
        \includegraphics[width=0.95\linewidth]{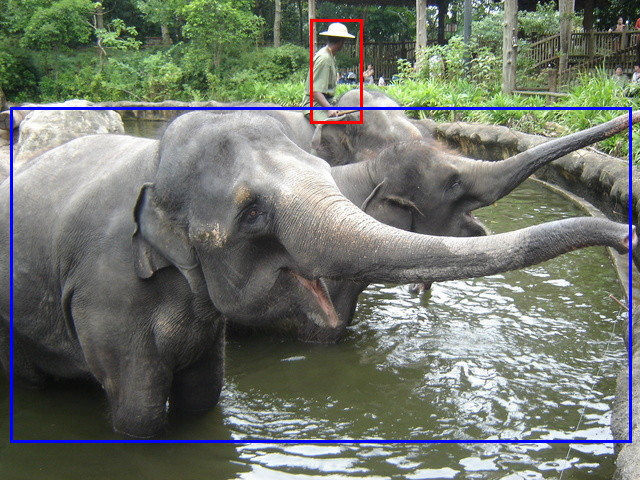}
    \end{minipage}%
    \hfill
    \begin{minipage}[b]{0.17\linewidth}
    \centering
        \small
        \textbf{Target person HOI: } \\
        ride elephant

        \textbf{VLM predictions: } \\
        ride elephant; \\
        \textcolor{red}{hop on elephant ($\times$)} 
    \end{minipage}
    \begin{minipage}[b]{0.16\linewidth}
    \centering
    \subcaption*{ (k) } %
        \includegraphics[width=0.95\linewidth]{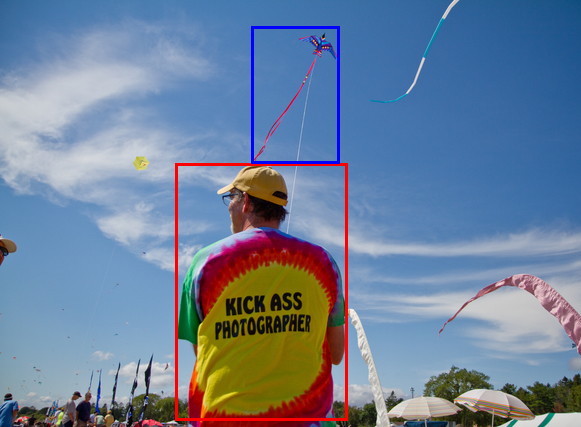}
    \end{minipage}%
    \hfill
        \begin{minipage}[b]{0.17\linewidth}
        \centering
        \small
        \textbf{Target person HOI: } \\
        fly kite

        \textbf{VLM predictions: } \\
        fly kite; \\
        \textcolor{red}{launch kite ($\times$)} 
    \end{minipage}
    \begin{minipage}[b]{0.16\linewidth}
    \centering
    \subcaption*{ (l) } %
        \includegraphics[width=0.5\linewidth]{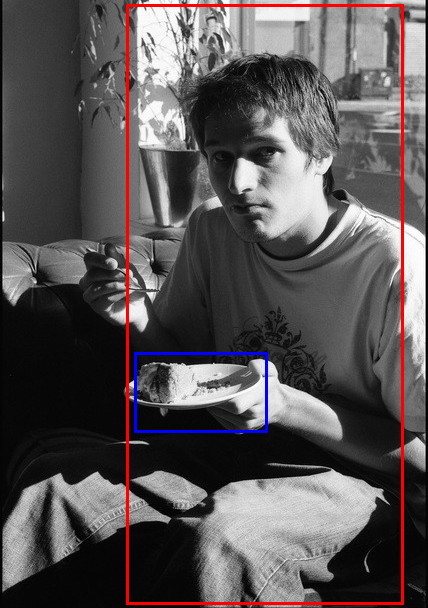}
    \end{minipage}%
    \hfill
        \begin{minipage}[b]{0.17\linewidth}
        \centering
        \small
        \textbf{Target person HOI: } \\
        hold cake

        \textbf{VLM predictions: } \\
        hold cake; \\
        \textcolor{red}{pick up cake($\times$)} 
    \end{minipage}
    
    \caption{Illustration of VLM (Qwen2.5-VL-32B) failure cases in \textit{Setting 1}, where HOI-specific model (ADA-CM) predicts correctly. 
    }
    \label{fig:VLM_problem}
\end{figure*}

%% file: figures/detect_failurecase.tex
\begin{figure*}[t]
    \centering
    \begin{minipage}[b]{0.23\linewidth}
       \subcaption*{\scriptsize Multi-person Scenario} %
		\includegraphics[width=\linewidth]{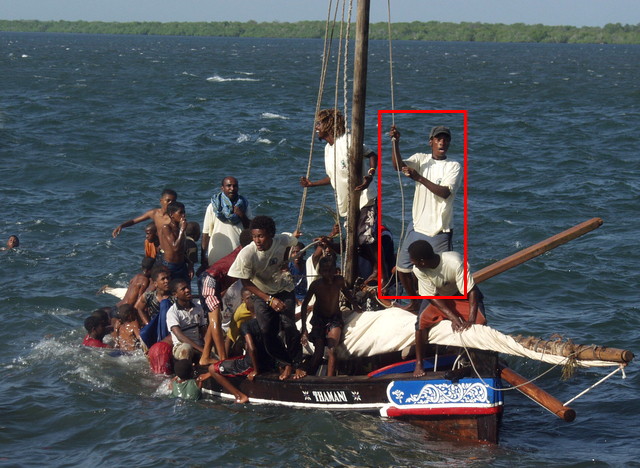}
	\end{minipage}
    \begin{minipage}[b]{0.23\linewidth}
		\includegraphics[width=\linewidth]{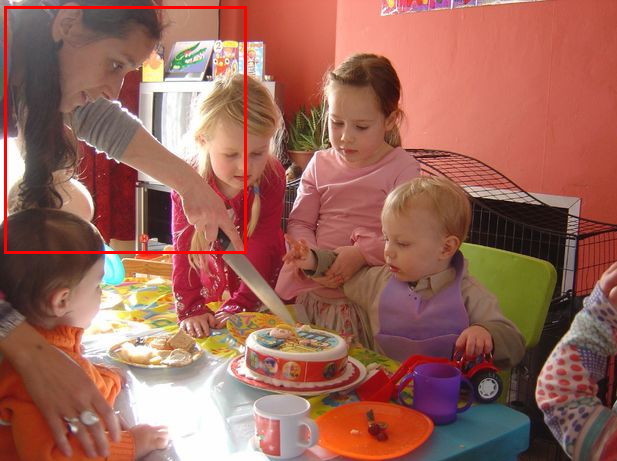}
	\end{minipage}
    \begin{minipage}[b]{0.23\linewidth}
		\includegraphics[width=\linewidth]{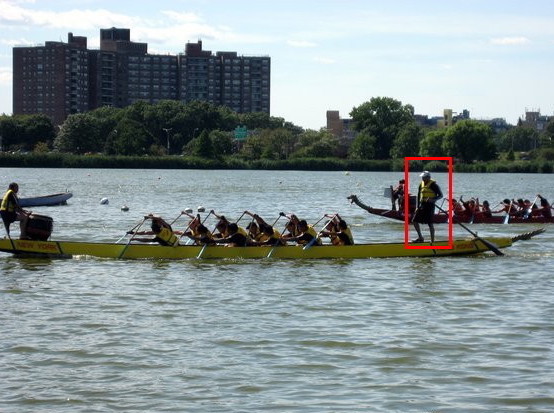}
	\end{minipage}	
    \begin{minipage}[b]{0.23\linewidth}
		\includegraphics[width=\linewidth]{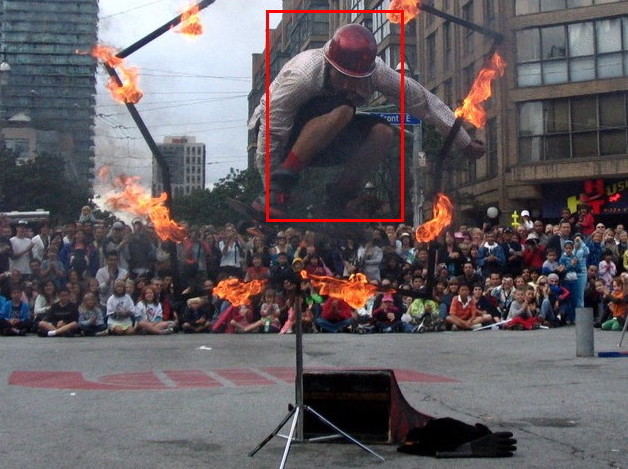}
	\end{minipage}

    \begin{minipage}[b]{0.23\linewidth}
       \subcaption*{\scriptsize Occluded Scenario} %
		\includegraphics[width=\linewidth]{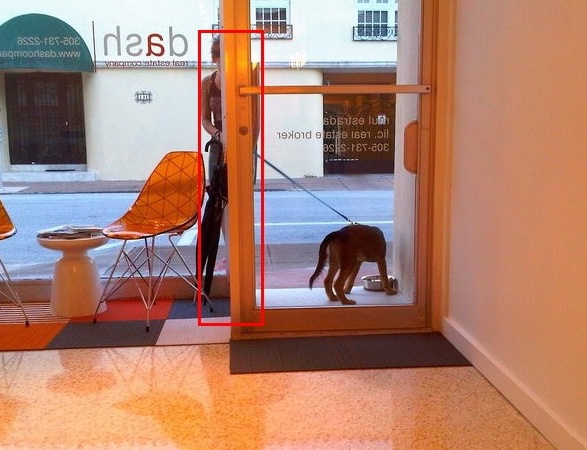}
	\end{minipage}
    \begin{minipage}[b]{0.23\linewidth}
		\includegraphics[width=\linewidth]{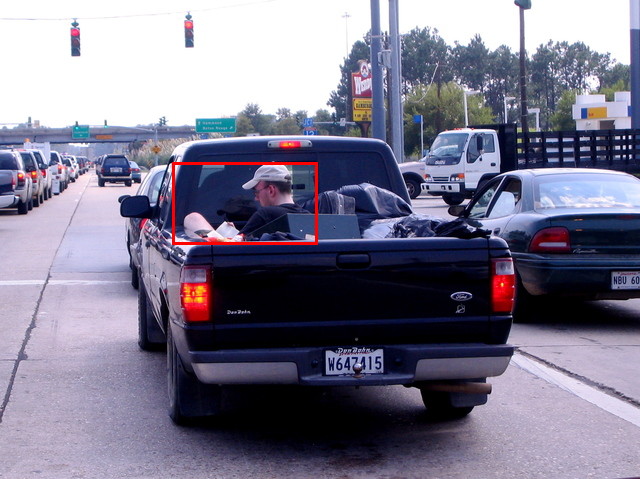}
	\end{minipage}
    \begin{minipage}[b]{0.23\linewidth}
		\includegraphics[width=\linewidth]{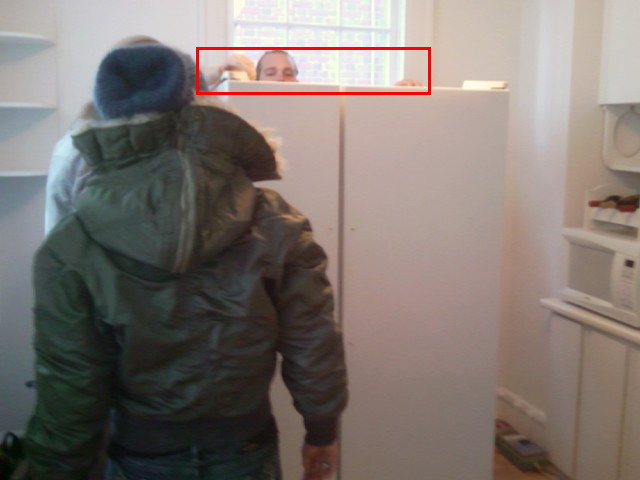}
	\end{minipage}	
    \begin{minipage}[b]{0.23\linewidth}
		\includegraphics[width=\linewidth]{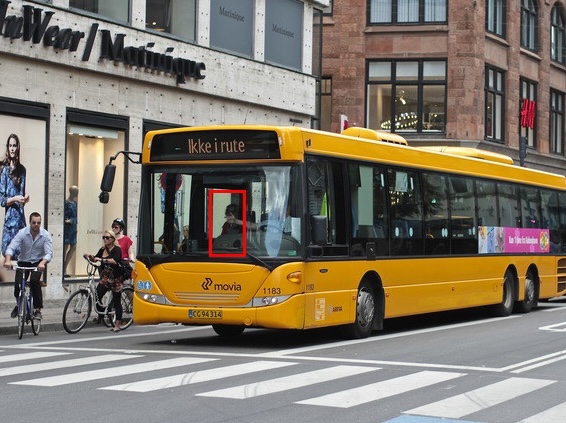}
	\end{minipage}

\caption{Failure detection cases of the Qwen2.5-VL-32B model in \textit{Setting 3}. The red box marks the target person specified in the question. The first row shows failures in multi-person scenarios, while the second row shows failures under occlusion.
} 
\label{fig:qwen_det_failurecase}
\end{figure*}

%% file: figures/result_dist_main.tex
\begin{figure}[t]
    \centering
    \begin{minipage}[b]{0.48\linewidth}
        \subcaption*{\scriptsize (a) Pre-trained} %
		\includegraphics[width=\linewidth]{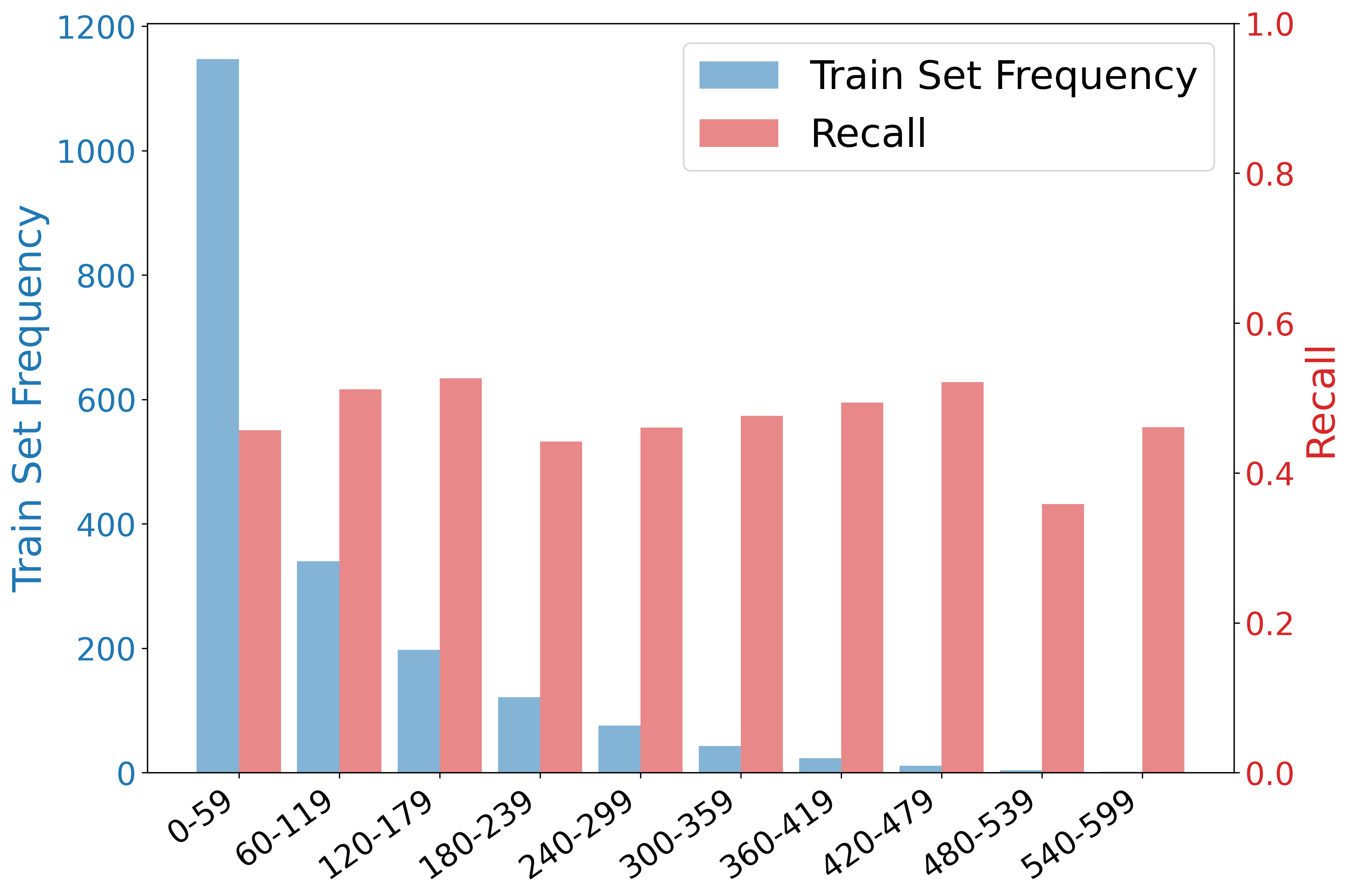}
	\end{minipage}
    \begin{minipage}[b]{0.48\linewidth}
       \subcaption*{\scriptsize (b) SFT (Our train set)} %
		\includegraphics[width=\linewidth]{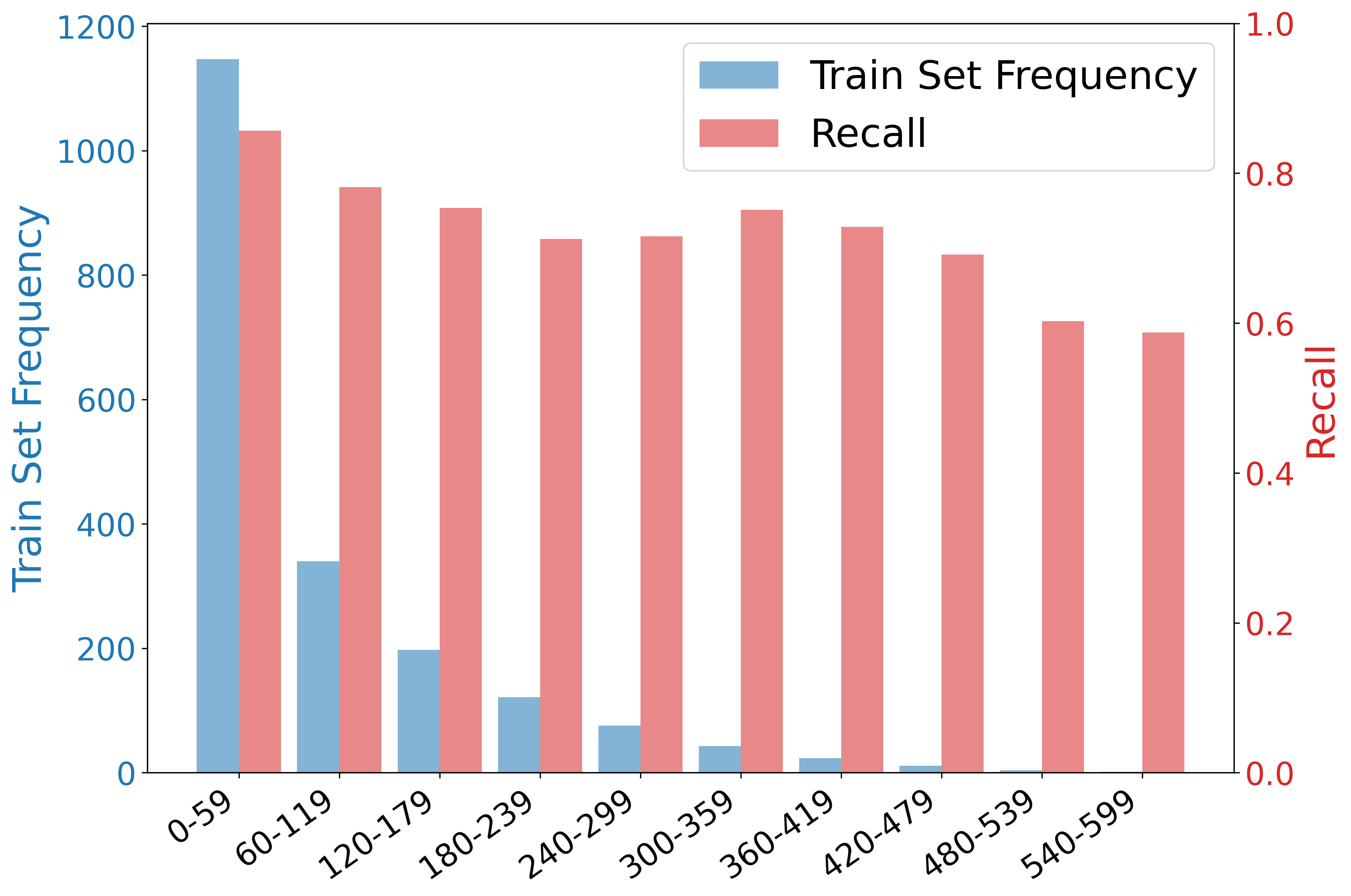}
	\end{minipage}


\caption{Comparison among pre-trained Qwen2.5-VL-7B~\cite{Qwen_2_5_report} and SFT finetuned Qwen2.5-VL-7B on our train set. The blue histograms indicate the binned class frequency in our training dataset, while the red histograms present the recall rate. Head and tail classes are defined by HOI class frequency in our training set, and all HOI classes are ordered accordingly. }
\label{fig:results_analysis_main}
\end{figure}